%% file: main.tex
\documentclass[10pt,letterpaper, twocolumn]{article}
\input{config/macro}

\input{config/mlpreamble}

\usepackage[pagenumbers]{config/cvpr} 

\usepackage{graphicx}
\usepackage{amsmath, mathtools, empheq, esint}
\usepackage{amssymb}
\usepackage{booktabs}
\usepackage{wrapfig}
\usepackage{lipsum}
\usepackage{xspace}
\usepackage{amsthm}
\usepackage{arydshln}
\usepackage{siunitx}
\usepackage{xcolor,colortbl}
\usepackage{multirow}
\usepackage{tabularx}
\usepackage{url}
\usepackage{etoc}
\usepackage{titletoc}
\usepackage{paralist}
\usepackage[toc,page]{appendix}
\graphicspath{{images/}}

\usepackage[pagebackref,breaklinks,colorlinks]{hyperref}

\usepackage[capitalize]{cleveref}
\crefname{section}{Sec.}{Secs.}
\Crefname{section}{Section}{Sections}
\Crefname{table}{Table}{Tables}
\crefname{table}{Tab.}{Tabs.}

\def\cvprPaperID{0357} 
\def\confName{CVPR}
\def\confYear{2023}

\begin{document}

\title{Multi-View Azimuth Stereo via Tangent Space Consistency}

\input{config/authors}

\input{sections/figures_tables/teaser.tex}

\input{sections/00_abstract.tex}

\input{sections/01_introduction.tex}

\input{sections/02_related_work.tex}

\input{sections/03_method.tex}

\input{sections/04_experiments}

\input{sections/06_conclusions}

\section*{Acknowledgement}
We thank Wenqi Yang, Akshat Dave, and Berk Kaya for code/data, and Boxin Shi, Min Li, and Heng Guo for discussions.
This work was supported by JSPS KAKENHI Grant Number JP19H01123.

{\small
\bibliographystyle{config/ieee_fullname}
\bibliography{config/egbib}
}

\clearpage
\input{supp/supp.tex}
\end{document}

%% file: config/macro.tex
\renewcommand{\nu}{\ensuremath{\mathbf{n}(\mathbf{u})}\xspace}  

\newcommand{\R}{\ensuremath{\mathbb{R}}\xspace}

\newcommand{\halfpi}{\ensuremath{\pm {\pi \over 2}}\xspace}

\newcommand{\zenith}{zenith\xspace}
\newcommand{\surface}{\ensuremath{\mathcal{M}}\xspace}
\newcommand{\visibility}{\ensuremath{\Phi_{i}}\xspace}
\newcommand{\point}{\ensuremath{\V{x}}\xspace}
\newcommand{\normal}{\ensuremath{\V{n}}\xspace}
\newcommand{\tangent}{\ensuremath{\V{t}}\xspace}
\newcommand{\cameraNum}{\ensuremath{C}\xspace}

\newcommand{\batchsize}{\ensuremath{P}\xspace}
\newcommand{\mask}{\ensuremath{O}\xspace}
\newcommand{\projectedTangentVector}{projected tangent vector\xspace}
\newcommand{\projectedTangentVectors}{projected tangent vectors\xspace}
\newcommand{\stackedTangentVectors}{\ensuremath{\V{T}(\point)}\xspace}
\newcommand{\diligentmv}{\mbox{DiLiGenT-MV}\xspace}

\newcommand{\loss}{\mathcal{L}\xspace}
\newcommand{\opticalAxis}{\ensuremath{\V{e}_{z}\xspace}}

\newcommand{\pandora}{\mbox{PANDORA}\xspace}
\newcommand{\psnerf}{\mbox{PS-NeRF}\xspace}
\newcommand{\sdps}{\mbox{SDPS}\xspace}
\newcommand{\uanet}{\mbox{UA-MVPS}\xspace}
\newcommand{\rmvps}{\mbox{R-MVPS}\xspace}
\newcommand{\bmvps}{\mbox{B-MVPS}\xspace}
\newcommand{\volsdf}{\mbox{VolSDF}\xspace}

\newcommand{\mvas}{MVAS\xspace}

\newcommand{\tsc}{\mbox{TSC}\xspace}

\newcommand{\pointOne}{\ensuremath{\point_1}\xspace}
\newcommand{\pointTwo}{\ensuremath{\point_2}\xspace}
\newcommand{\pointsetOne}{\ensuremath{\chi_{1}}\xspace}
\newcommand{\pointsetTwo}{\ensuremath{\chi_{2}}\xspace}
\newcommand{\fscoreThreshold}{\ensuremath{\tau}\xspace}
\newcommand{\chamferDist}{\ensuremath{d(\pointsetOne, \pointsetTwo)}\xspace}
\newcommand{\precision}{\ensuremath{\mathcal{P}}\xspace}
\newcommand{\recall}{\ensuremath{\mathcal{R}}\xspace}
\newcommand{\fscore}{\ensuremath{\mathcal{F}}\xspace}

\newcommand{\phaseangle}{\ensuremath{\hat{\phi}}\xspace}
\newcommand{\azimuthangle}{\ensuremath{\phi}\xspace}

\newcommand{\colorbar}[3]{
\begin{tabular}[t]{@{}l@{}l@{}}
	\includegraphics[height=#1\linewidth,width=0.5em]{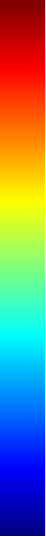} & 
	\begin{tabular}[b]{@{}l}
		#2 \\ [#3pt]
		$0$
	\end{tabular}
\end{tabular}
}

%% file: config/mlpreamble.tex
\usepackage{xcolor}
\newcounter{todos}
\AtEndDocument{\ifnum\value{todos}>0 \PackageWarning{TODOS}{There are \arabic{todos} todos left in this paper! Fix them before submitting the paper!} \fi}

\newcommand{\argmin}{\mathop{\mathrm{argmin}}\limits}

\newcommand{\norm}[1]{\lVert#1\rVert}

\newcommand{\V}[1]{\ensuremath{\mathbf{#1}}}

\usepackage{bm}
\newcommand{\Vg}[1]{\ensuremath{\boldsymbol{#1}}}

\newcommand{\abs}[1]{\lvert#1\rvert}


%% file: config/authors.tex
\author{
	Xu Cao \quad 
	Hiroaki Santo \quad
	Fumio Okura \quad
	Yasuyuki Matsushita
	\\
	Osaka University
	\\
	{\tt \small \{cao.xu, santo.hiroaki, okura, yasumat\}@ist.osaka-u.ac.jp}
	\\
	{\tt \small Source code: \url{https://github.com/xucao-42/mvas}}
}

%% file: sections/figures_tables/teaser.tex
\twocolumn[{
    \renewcommand\twocolumn[1][]{#1}
    \maketitle
    \centering
    \vspace{-0.5em}
    \begin{minipage}{\textwidth}
        \centering
        \includegraphics[trim=000mm 000mm 000mm 000mm, clip=False, width=\linewidth]{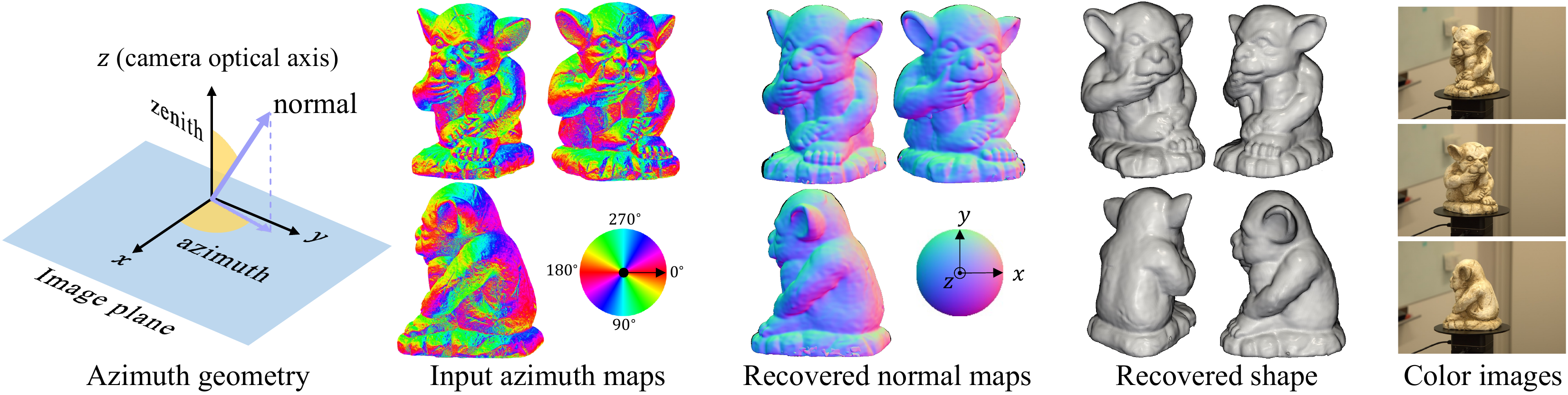}
    \end{minipage}
    \vspace{-0.5 em}
    \captionof{figure}{
3D reconstruction from calibrated multi-view azimuth maps ($3$ out of $31$ are shown).
An azimuth angle indicates the surface normal's orientation in the image plane, and an azimuth map records the azimuth angles across the entire surface.
We show that azimuth maps can be effectively used for shape and normal recovery.
Color images are for reference only and are not used in shape optimization. 
  }
    \label{fig.teaser}
    \vspace{1.2em}
}]

%% file: sections/00_abstract.tex
\begin{abstract}
We present a method for 3D reconstruction only using calibrated multi-view surface azimuth maps.
Our method, multi-view azimuth stereo, is effective for textureless or specular surfaces, which are difficult for conventional multi-view stereo methods.
We introduce the concept of tangent space consistency: Multi-view azimuth observations of a surface point should be lifted to the same tangent space. 
Leveraging this consistency, we recover the shape by optimizing a neural implicit surface representation.
Our method harnesses the robust azimuth estimation capabilities of photometric stereo methods or polarization imaging while bypassing potentially complex zenith angle estimation.
Experiments using azimuth maps from various sources validate the accurate shape recovery with our method, even without zenith angles.
\end{abstract}

%% file: sections/01_introduction.tex
\section{Introduction}
\label{sec:intro}
Recovering 3D shapes of real-world scenes is a fundamental problem in computer vision, and multi-view stereo~(MVS) has emerged as a mature geometric method for reconstructing dense scene points. Using 2D images taken from different viewpoints, MVS finds dense correspondences between images based on the photo-consistency assumption, that a scene point's brightness should appear similar across different viewpoints~\cite{multiview2007vogiatzis,multiview2007goesele,vu2011high, multiview2005vogiatzis}. However, MVS struggles with textureless or specular surfaces, as the lack of texture leads to ambiguities in establishing correspondences, and the presence of specular reflections violates the photo-consistency assumption~\cite{furukawa2015multi}.

Photometric stereo (PS) offers an alternative approach for dealing with textureless and specular surfaces~\cite{shi2019}. 
By estimating single-view surface normals using varying lighting conditions~\cite{woodham1980photometric}, PS enables high-fidelity 2.5D surface reconstruction~\cite{nehab2005efficiently}. However, extending PS to a multi-view setup, known as multi-view photometric stereo~(MVPS)~\cite{hernandez2008multiview}, significantly increases image acquisition costs, as it requires multi-view and multi-light images under highly controlled lighting conditions~\cite{li2020multi}. 

To mitigate image acquisition costs, simpler lighting setups such as circularly or symmetrically placed lights have been explored~\cite{alldrin2007toward,Zhou2010,chandraker2012differential,minami2022symmetric}. 
With these lighting setups, estimating the surface normal's azimuth (the angle in the image plane) becomes considerably easier than estimating the zenith (the angle from the camera optical axis)~\cite{alldrin2007toward,chandraker2012differential,minami2022symmetric}. 
The ease of azimuth estimation also appears in polarization imaging~\cite{rahmann2001reconstruction}.
While azimuth can be determined up to a $\pi$-ambiguity using only polarization data, zenith estimation requires more complex steps~\cite{smith2016linear,miyazaki2003polarizationtwoview,stolz2012shape}.

In this paper, we introduce Multi-View Azimuth Stereo~(MVAS), a method that effectively uses calibrated multi-view azimuth maps for shape recovery (\cref{fig.teaser}). 
MVAS is particularly advantageous when working with accurate azimuth acquisition techniques. 
With circular-light photometric stereo~\cite{chandraker2012differential}, MVAS has the potential to be applied to surfaces with arbitrary isotropic materials.
With polarization imaging~\cite{dave2022pandora}, MVAS allows a passive image acquisition as simple as MVS while being more effective for textureless or specular surfaces.

\begin{figure}
	\centering
	\includegraphics[width=\linewidth]{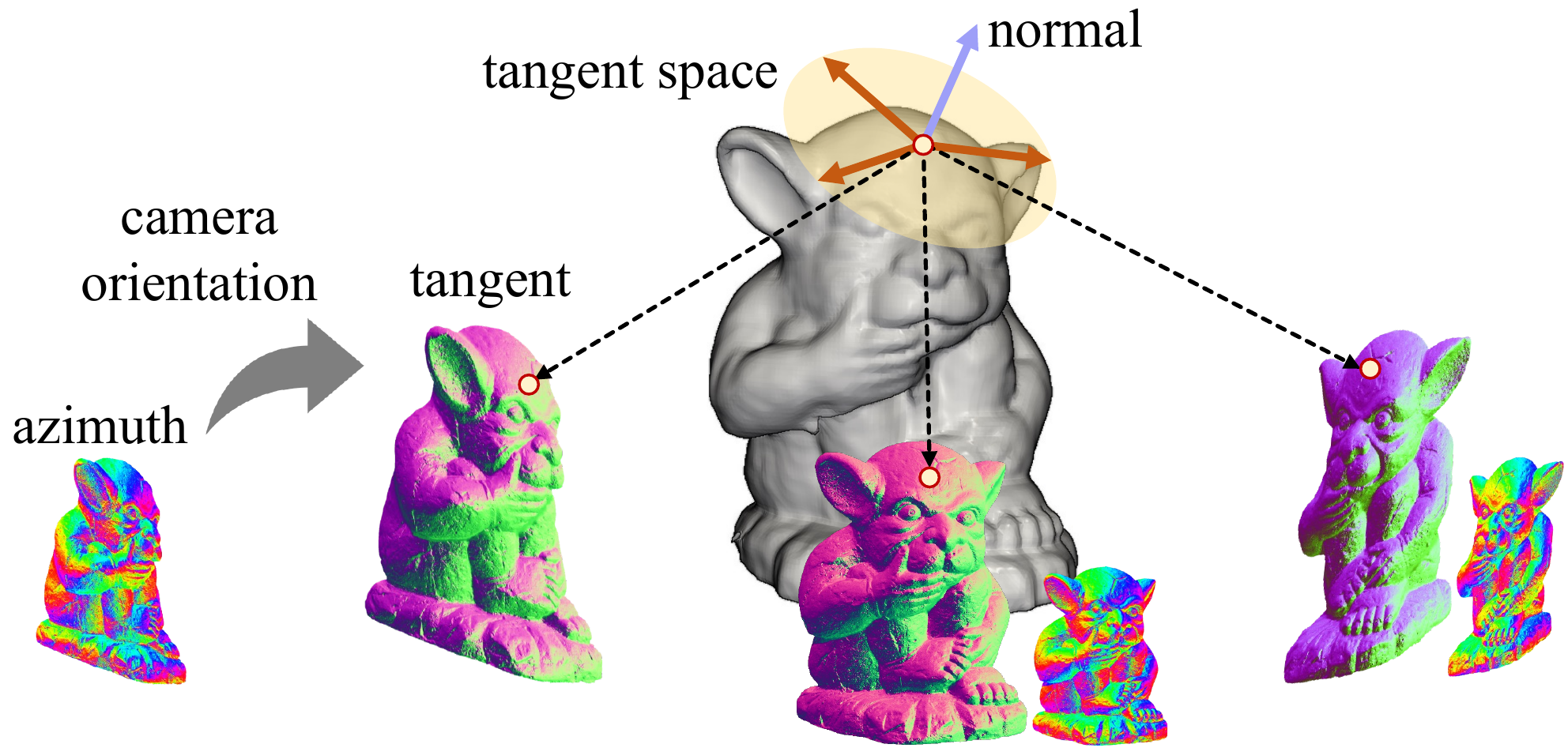}
	\caption{Tangent space consistency. The azimuth can be converted to a tangent by camera orientation. The tangents in different views, but projected from the same surface point, should lie in the same tangent space and can directly determine the surface normal. }
	\label{fig.tsc}
\end{figure}

The key insight enabling MVAS is the concept of Tangent Space Consistency (TSC) for multi-view azimuth angles. 
We find that the azimuth can be transformed into a tangent using camera orientation.
Therefore, multi-view azimuth observations of the same surface point should be lifted to the same tangent space (\cref{fig.tsc}). 
TSC helps determine if a 3D point lies on the surface, similar to photo-consistency for finding image correspondences. 
Moreover, TSC can directly determine the surface normal as the vector orthogonal to the tangent space, enabling high-fidelity reconstruction comparable to MVPS methods. 
Notably, TSC is invariant to the $\pi$-ambiguity of the azimuth angle, making MVAS well-suited for polarization imaging.

With TSC, we reconstruct the surface implicitly represented as a neural signed distance function (SDF), by constraining the surface normals (\ie, the gradients of the SDF). Experimental results show that MVAS achieves comparable reconstruction performance to MVPS methods~\cite{kaya2022uncertainty, yang2022psnerf, park2016robust}, even in the absence of zenith information. Further, MVAS outperforms MVS methods~\cite{schoenberger2016mvs} in textureless or specular surfaces using azimuth maps from symmetric-light photometric stereo~\cite{minami2022symmetric} or a snapshot polarization camera~\cite{dave2022pandora}.

In summary, this paper's key contributions are:
\begin{itemize}
	\setlength{\itemsep}{0.2em}
	\setlength{\parskip}{0.2em}
	\item Multi-View Azimuth Stereo (MVAS), which enables accurate shape reconstruction even for textureless and specular surfaces;
	\item Tangent Space Consistency (TSC), which establishes the correspondence between multi-view azimuth observations, thereby facilitating the effective use of azimuth data in 3D reconstruction; and
	\item A comprehensive analysis of TSC, including its necessary conditions, degenerate scenarios, and the application to optimizing neural implicit representations.
\end{itemize}

%% file: sections/02_related_work.tex
\section{Related Tasks and Concept}
\label{sec.related_work}

This section discusses the relation of MVAS to multi-view photometric stereo (MVPS) and shape-from-polarization (SfP), and compares TSC to photo-consistency.
\vspace{-1em}
\paragraph{MVPS versus MVAS}
MVPS aims for high-fidelity shape and reflectance recovery using images from different angles and under different lighting conditions~\cite{hernandez2008multiview, logothetis2019differential}.
These ``multi-light'' images can be used for estimating and fusing multi-view normal maps~\cite{chang2007multiview,kaya2022uncertainty}, for refining coarse meshes initialized by MVS~\cite{park2016robust}, or for jointly estimating the shape and materials in an inverse-rendering manner~\cite{yang2022psnerf}.

Compared to MVPS, MVAS has the potential to be applied to (1) surfaces of a broader range of materials and/or (2) in uncontrolled scenarios, benefiting from azimuth inputs.
First, azimuth estimation is valid for arbitrary isotropic materials using an uncalibrated circular moving light~\cite{chandraker2012differential}, while MVPS methods require specific surface reflectance modeling (\eg, Lambertian~\cite{chang2007multiview} or the microfacet model~\cite{yang2022psnerf}) or prior learning~\cite{kaya2022uncertainty}.
Second, MVAS allows passive image capture with polarization imaging, while MVPS has to actively illuminate the scene, limiting MVPS's application in highly controlled environments. 

\vspace{-1em}
\paragraph{SfP versus MVAS}
SfP recovers surfaces using polarization imaging~\cite{sonyPolar}. 
For dielectric surfaces, the measured angle of polarization (AoP) aligns with the surface normal's azimuth component, up to a $\pi$ ambiguity.
SfP studies determine surface normals by resolving this $\pi$-ambiguity and estimating the zenith component~\cite{smith2016linear, smith2018height,drbohlav2001unambiguous,fukao2021polarimetric,ding2021polarimetric,kadambi2015polarized,kadambi2017depth,rahmann2001reconstruction,zhu2019depth}.
Some studies use polarization data to refine coarse shapes initialized by multi-view reconstruction methods~\cite{Cui_2017_CVPR,zhao2020polarimetric}, but the geometric relation between multi-view azimuth angles are not considered.

With TSC and MVAS, both the $\pi$-ambiguity and zenith estimation can be bypassed.
Our method relies on TSC, not requiring MVS methods to initialize shapes.

\vspace{-1em}
\paragraph{Photo-consistency versus tangent space consistency}
Photo-consistency is a key assumption in MVS for establishing correspondence between multi-view images. This assumption states that a scene point appears similar across different views and struggles with specular surfaces~\cite{furukawa2009accurate}.

In contrast, TSC is derived from geometric principles and strictly holds for multi-view azimuth angles. Further, TSC can determine the surface normal, providing more information than photo-consistency. However, TSC requires at least three cameras with non-parallel optical axes and can degrade to photo-consistency under certain camera configurations. 
Similar to photo-consistency's challenges with textureless surfaces, TSC might struggle to establish correspondences for planar surfaces.
Details are in \cref{sec.functional}.

%% file: sections/03_method.tex
\section{Proposed Method}
\label{sec:proposed_method}

We aim to recover the shape from calibrated and masked azimuth maps. Let $\Omega_i$ represent the $i$-th image pixel domain. For each view $i \in \{1, 2, ..., \cameraNum\}$, we assume the following are available:
\begin{itemize}
	\setlength{\itemsep}{0.2em}
	\setlength{\parskip}{0.2em}
	\item a surface azimuth map $\phi_i: \Omega_i \rightarrow [0, 2\pi]$,
	\item a binary mask indicating whether a pixel is inside the shape silhouette $\mask_i: \Omega_i \rightarrow \{0, 1\}$, and
	\item the projection from the world coordinates to the image pixel coordinates $\Pi_i: \R^3 \rightarrow \Omega_i$, consisting of the extrinsic rigid-body transformation $\V{P}_i = [\V{R}_i \mid \V{t}_i] \in SE(3)$ and intrinsic perspective camera projection $\V{K}_i$.
\end{itemize}

We describe the proposed method in three sections. First, we detail the transformation from an azimuth angle to a projected tangent vector (\cref{sec.azimuth2tangent}). Next, we discuss multi-view tangent space consistency for surface points, including its four degenerate scenarios and $\pi$-invariance (\cref{sec.functional}). Lastly, we present the surface reconstruction by optimizing a neural implicit representation based on the tangent space consistency loss (\cref{sec.sdf_optimization}).

\subsection{The projected tangent vector}
\label{sec.azimuth2tangent}
This section will show how to convert an azimuth angle to a tangent vector of the surface point, given the world-to-camera rotation. We will only consider single-view observations and ignore the view index in this section.

\begin{figure}[t]
	\centering
	\includegraphics[width=\linewidth]{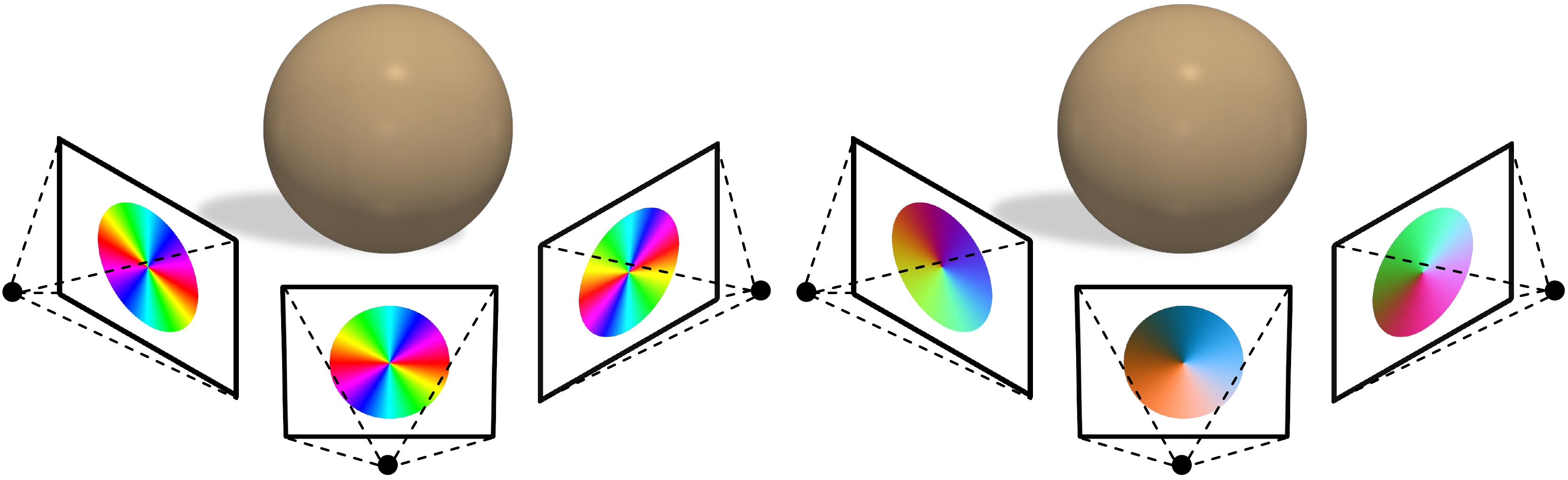}
	\caption{The azimuth angle observations \textbf{(left)} are lifted to tangent vectors \textbf{(right)} by world-to-camera rotations. 
	The tangent vectors are coded by 8-bit RGB colors using $255(\tangent+\V{1})/2$.}
	\label{fig.azimuth2tangent}
\end{figure}

In the world coordinates, consider a unit normal vector $\V{n}(\V{x})\in \mathcal{S}^2 \subset \R^3$ of a surface point $\point \in \R^3$.
Suppose a rigid-body transformation $[\V{R} \mid \V{t}]$ transforms the surface from the world coordinates to the camera coordinates.
The direction of the normal vector in the camera coordinates  $\V{n}^c$ is rotated accordingly as
\begin{equation}
	\V{R}\V{n} = \V{n}^c.
	\label{eq.world_normal_to_camera_normal}
\end{equation}
In the camera coordinates, we can parameterize the unit normal vector by its azimuth angle $\phi \in [0, 2\pi]$ and \zenith angle~$\theta \in \left[0, {\pi \over2}\right)$ as
\begin{equation}
	\V{n}^c = \left[ 
	\begin{matrix}
		n_x^c \\
		n_y ^c\\
		n_z^c
	\end{matrix}
	\right] =
	\left[ \begin{matrix}
		\sin\theta \cos \phi \\
		\sin\theta \sin \phi \\
		\cos\theta \\
	\end{matrix}  \right].
	\label{eq.normal_parameterization}
\end{equation}
From~\cref{eq.normal_parameterization}, we can derive the relation between $n_x^c$ and $n_y^c$ in terms of only the azimuth angle as
\begin{equation}
	n_x^c \sin \phi = n_y^c \cos \phi.
	\label{eq.azimuth_ratio}
\end{equation}
Denoting the rotation matrix as 
\begin{equation}
	\V{R} = \left[\begin{matrix}
	-	\V{r}_1^\top - \\
	-	\V{r}_2^\top - \\
	-	\V{r}_3^\top -
	\end{matrix}\right] \in SO(3),
	\label{eq.rotation_matrix}
\end{equation}
and putting \cref{eq.world_normal_to_camera_normal,eq.normal_parameterization,eq.azimuth_ratio,eq.rotation_matrix} together, we obtain
\begin{equation}
	\V{r}_1^\top \V{n} \sin \phi = \V{r}_2^\top \V{n} \cos \phi.
	\label{eq:world_normal_azimuth}
\end{equation}
Rearranging \cref{eq:world_normal_azimuth} yields
\begin{equation}
	\V{n}^\top \underbrace{(\V{r}_1 \sin \phi - \V{r}_2 \cos \phi)}_{\V{t}(\phi)} = 0.
	\label{eq.normal_tangent}
\end{equation}
We call $\V{t}(\phi)$ the \emph{\projectedTangentVector}, as it is computed from the projected azimuth angle and perpendicular to the surface normal.
As shown in \Cref{fig.azimuth2tangent}, the transformation from azimuth maps to tangent maps reveals that projected tangent vectors encode camera orientation information, providing useful hints for multi-view reconstruction.

\vspace{-1em}
\paragraph{Properties}
The \projectedTangentVector is the unit vector parallel to the intersection of the tangent and image spaces.
Based on \cref{eq.normal_tangent}, 
\begin{equation}
	\begin{aligned}
		\V{t}^\top\V{t} &= \V{r}_1 ^\top \V{r}_1  \sin^2\phi+\V{r}_2^\top \V{r}_2\cos^2\phi -2\V{r}_1^\top\V{r}_2\cos\phi \sin \phi\\
		&=  \sin^2\phi +  \cos^2\phi = 1,
	\end{aligned}
	\label{eq.tangent_unit_length}
\end{equation}
since $\V{r}_1$ and $\V{r}_2$ are orthonormal vectors.

\begin{wrapfigure}{r}[1em]{0.4\linewidth}
	\hspace{-3em}
	\vspace{-1em}
	\centering
	\includegraphics[width=\linewidth]{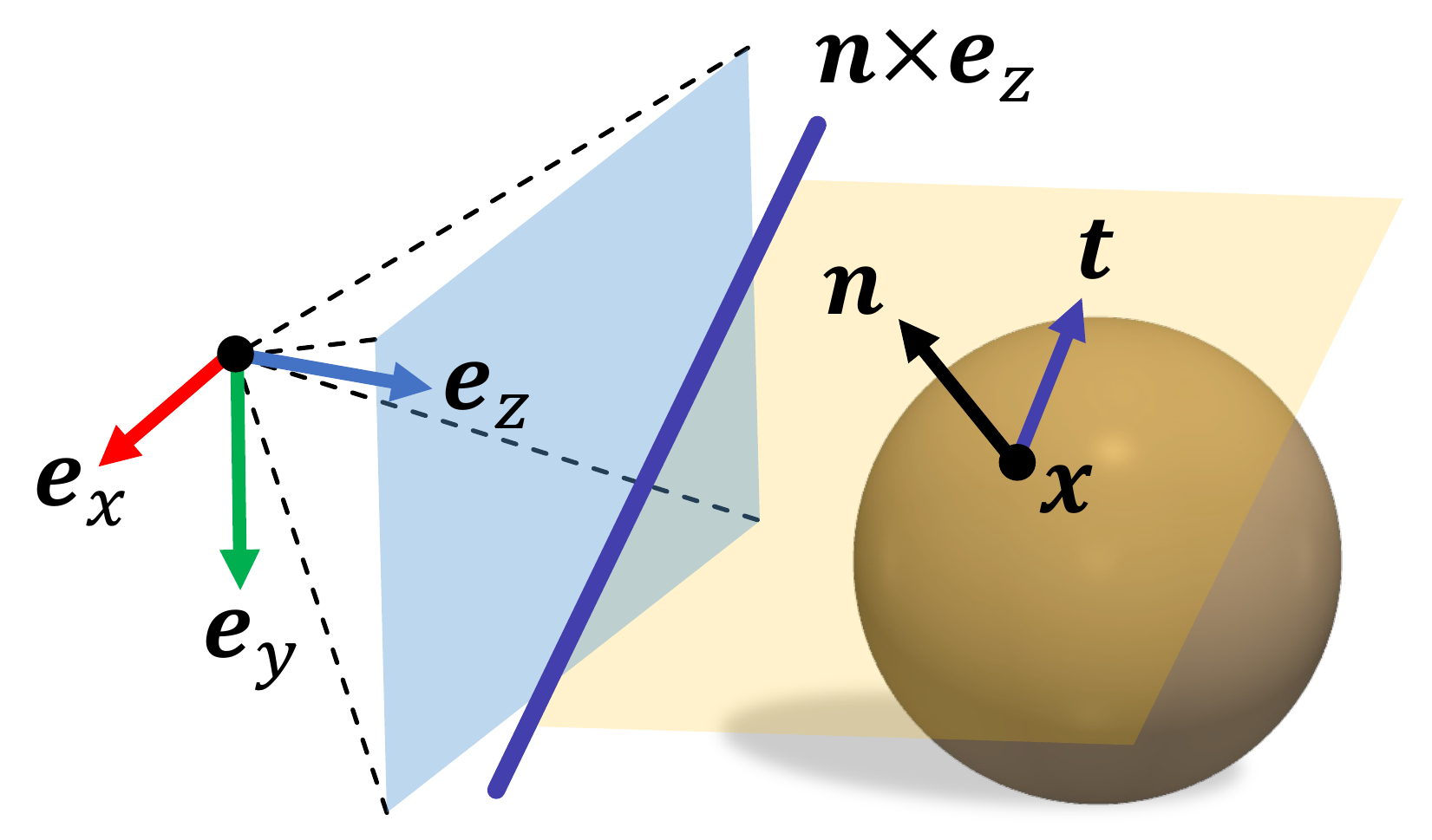}
\end{wrapfigure}
The inset illustrates the second property.
Let $\V{e}_x$, $\V{e}_y$, and $\V{e}_z$ be the unit direction vector of the $x$-, $y$-, and $z$-axis of the camera coordinates in the world coordinates.
Then $\V{r}_1 = \V{e}_x$ and $\V{r}_2 = \V{e}_y$, which follows that $\V{t}(\phi)$ is a linear combination of camera's $x$- and $y$-axes and thus parallel to the image plane.
We can compute the intersection direction of two planes by taking the cross-product of their normals, namely, the surface normal and the principle axis. Hence,
\begin{equation}
	\tangent \parallel \normal \times \opticalAxis .
	\label{eq.tangent_constraint}
\end{equation}
The two properties are helpful in analyzing the tangent space consistency, as described next.

\subsection{Multi-view tangent space consistency}
\label{sec.functional}
This section discusses the consistency between multi-view azimuth observations in the tangent space of a surface point.
In addition, four degenerate scenarios and $\pi$-invariance will be discussed.
We assume the surface point under consideration is visible to all cameras in this section.

Denote the \projectedTangentVector of a surface point in $i$-th view as $\tangent_{i}(\point)=\tangent(\phi_i(\Pi_i(\point)))$.
By \cref{eq.normal_tangent}, a surface point $\point$, its normal direction $\normal$, and its multi-view \projectedTangentVectors $\tangent_i$ should satisfy:
\begin{align}
	\normal(\point)^\top \tangent_i(\point) = 0 \quad \forall i.
	\label{eq.tsc}
\end{align}
Let
$
\stackedTangentVectors=[\V{t}_1(\point), \V{t}_2(\point), ..., \V{t}_\cameraNum(\point)]^\top \in \R^{\cameraNum \times 3}
$
be the matrix formed by stacking \projectedTangentVectors of all \cameraNum views. 
Then \cref{eq.tsc} reads
\begin{align}
	\stackedTangentVectors \normal(\point) = \V{0}.
	\label{eq.tsc_mat}
\end{align}
\Cref{eq.tsc_mat} can only be satisfied if the rank of \stackedTangentVectors is either 1 or 2. The rank cannot be $0$ as \projectedTangentVectors are unit length. The case rank$(\stackedTangentVectors)=3$ cannot satisfy \cref{eq.tsc_mat} as surface normals are non-zero vectors.

We refer to the case where the rank of \stackedTangentVectors is $2$ as \emph{tangent space consistency} (TSC). In this case, multi-view \projectedTangentVectors from a surface point span its tangent space, and the surface normal is determined up to a sign ambiguity. On the other hand, when rank$(\stackedTangentVectors)=1$, the \projectedTangentVectors can only span a tangent line and constrain the surface normal on the plane orthogonal to the tangent line. This can occur when camera optical axes are parallel, as explained later.

\begin{figure}[t]
	\centering
	\includegraphics[width=\linewidth]{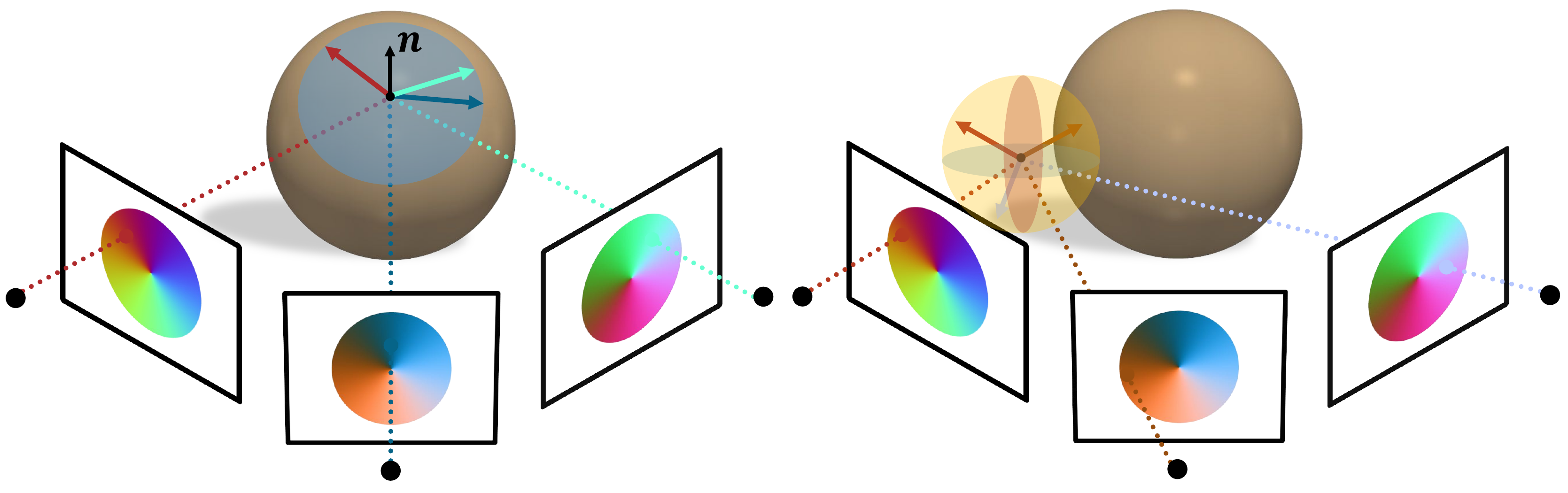}
	\caption{\textbf{(Left)} Multi-view \projectedTangentVectors from a surface point span its tangent space and determine the surface normal. \textbf{(Right) }Conversely, multi-view \projectedTangentVectors from a non-surface point (\ie, wrong correspondence) are expected to span the 3D space.}
	\label{fig.tangent_space_consistency}
\end{figure}

TSC can help distinguish non-surface points (wrong correspondences) from surface points (possibly correct correspondences) and determine the surface normals, as shown in \cref{fig.tangent_space_consistency}. For wrong correspondences, their \projectedTangentVectors are expected to have a rank of $3$ and span the entire 3D space. On the other hand, for surface points, their \projectedTangentVectors span the tangent space, i.e., rank$(\stackedTangentVectors)=2$. In addition, TSC requires the surface normal to be in the null space of \stackedTangentVectors, i.e., perpendicular to the tangent space spanned by \projectedTangentVectors. 
This makes TSC more informative than photo-consistency since photo-consistency cannot directly determine the surface normal.

\begin{figure}[t]
	\centering
		\includegraphics[width=\linewidth]{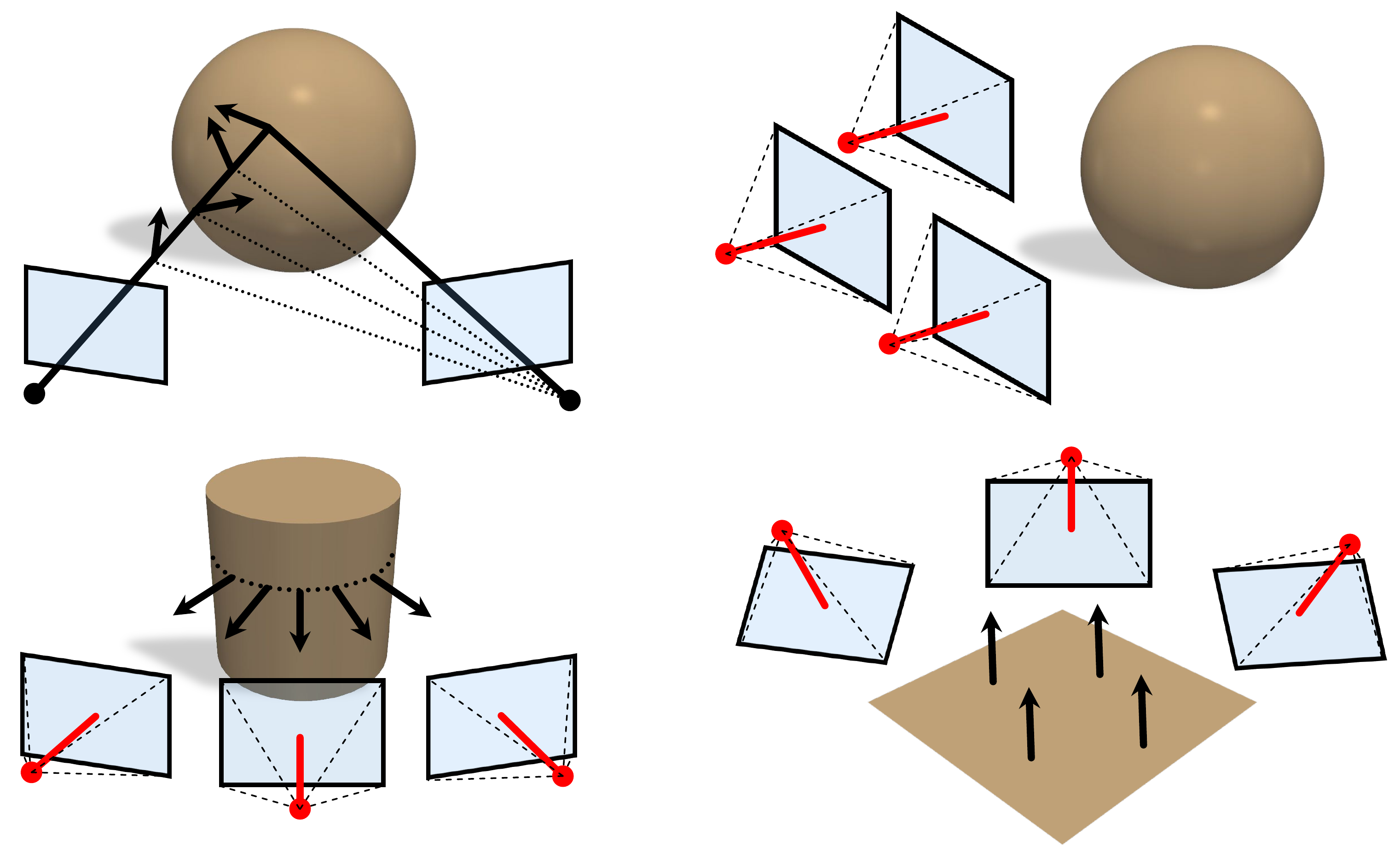}
	\caption{Degeneration scenarios where TSC cannot distinguish good correspondence from bad ones. \textbf{(Top Left)} Two-view observations,  \textbf{(Top Right)} frontal parallel cameras with parallel optical axes (red pins), \textbf{(Bottom Left)} cameras with coplanar optical axes observe coplanar surface normals, and  \textbf{(Bottom Right)} planar surface regions. }
	\label{fig.degerated_cases}
\end{figure}

To effectively distinguish surface/non-surface points using TSC, a non-planar surface must be observed by at least three cameras with non-parallel optical axes. These requirements indicate four degeneration scenarios, as shown in \cref{fig.degerated_cases} and discussed below. \Cref{tab.degeneration_tsc} summarizes the variations of rank$(\stackedTangentVectors)$ in these scenarios.
\vspace{-1em}
\paragraph{Number of viewpoints}
For TSC to be effective, the rank of $\stackedTangentVectors$ is expected to be $3$ for non-surface points. However, when only two views are available, the rank of  \stackedTangentVectors is impossible to achieve $3$ since $\stackedTangentVectors \in \R^{2\times 3}$. In this case, rank$(\stackedTangentVectors)\leq2$ is satisfied for arbitrary correspondence. Consequently, TSC cannot distinguish surface points from non-surface points in the two-view case.
\vspace{-1em}
\paragraph{Camera setups}
TSC requires the \projectedTangentVectors of a surface point can span the tangent space but not a tangent line.
This requirement breaks down when \projectedTangentVectors are observed from 
(1) frontal parallel cameras, or (2) cameras with coplanar optical axes.

\begin{table}
	\centering
	\scriptsize
	\caption{Rank of \stackedTangentVectors in four degenerate cases of TSC. 
	}
	\resizebox{\linewidth}{!}{
		\begin{tabular}{@{}lccc}
			\toprule
			Scenarios & Non-surface points & surface points & surface normal \\
			\midrule
			Two-view  & 2 & 2 & $\times$ \\
			Co-linear optical axes & 2 & 1 & $\times$\\
			Co-planar optical axes & 2 & 1 & $\triangle$ \\
			Planar surface & 2 & 2 &  $\checkmark$ \\
			\midrule
			TSC & 3 & 2 & $\checkmark$ \\
			\bottomrule
		\end{tabular}
	}
	\label{tab.degeneration_tsc}
	\vspace{-1.5em}
\end{table}

Frontal parallel cameras have parallel optical axes. 
By  \cref{eq.tangent_constraint}, multi-view projected tangent vectors of a surface point also become parallel. 
This reduces the rank of \stackedTangentVectors to 1, and TSC degrades to photo-consistency since all cameras should observe the same tangent vector for a surface point.

A more special case is when cameras with coplanar optical axes observe coplanar surface normals, such as a rotating camera observing a cylinder. In this case, the cross product of the coplanar normal and optical axis vectors yields co-linear projected tangent vectors. As such, the rank of \stackedTangentVectors is 1 for surface points, and TSC again degrades to photo-consistency. However, this degradation does not occur for non-coplanar surface normals, meaning TSC can still be effective for general surfaces.

\vspace{-1em}
\paragraph{Surface types} 
TSC breaks down for a planar surface.
At any location on the planar surface, $\normal(\point)$ is the same and rank$(\stackedTangentVectors
)$ is identically $2$ for arbitrary correspondence.
However, the normal direction of this plane can still be correctly determined in the case rank$(\stackedTangentVectors) = 2$, \ie, at least three non-frontal parallel views.
The planar surface can be seen as the counterpart to the textureless region for photo-consistency. 
However, unlike photo-consistency, TSC can still determine the surface normal\footnote{A similar phenomenon exists in Helmholtz stereopsis~\cite{zickler2002helmholtz}, where wrong correspondence might still result in the correct normal estimation.}.
\vspace{-1em}
\paragraph{$\pi$-invariance} 
TSC remains effective when the azimuth angle is changed by $\pi$.
By~\cref{eq.normal_tangent}, the sign of the projected tangent vector will be reversed:
\begin{equation}
	\tangent(\phi + \pi) = - \V{r}_1 \sin \phi + \V{r}_2 \cos \phi = -\tangent(\phi).
\end{equation}
Intuitively, reversing the direction of a tangent vector still places it in the same tangent space, as $\normal ^\top (-\tangent)= 0$ when $\normal ^\top \tangent=0$.
Mathematically, reversing the signs of arbitrary rows in \stackedTangentVectors does not affect the rank of \stackedTangentVectors. This $\pi$-invariance can be particularly useful for polarization imaging, as they can only measure azimuth angles up to a $\pi$ ambiguity.

\subsection{Multi-view azimuth stereo}
\label{sec.sdf_optimization}
We propose the following TSC-based functional for multi-view geometry reconstruction:
\begin{equation}
	\mathcal{J} = \oiint_{\surface} \frac{\sum_{i=1}^{\cameraNum} \visibility(\point) \left(\normal(\point)^\top\tangent_i(\point)\right)^2}{\sum_{i=1}^{\cameraNum}\visibility(\point)} \, d\surface.
	\label{eq.functional_long}
\end{equation}
Here, \surface is the surface embedded in the 3D space, and $d\surface$ is the infinitesimal area on the surface. 
$\visibility(\point)$ is a binary function indicating the visibility of the point \point from the $i$-th viewpoint:
\begin{equation}
	\visibility(\point) = \begin{cases}
		1 \quad \text{if \point is visible to $i$-th camera} \\
		0 \quad \text{otherwise}
	\end{cases}.
\end{equation}
We can simplify \cref{eq.functional_long} as follows:
\begin{equation}
	\mathcal{J} = \oiint_{\surface} \normal^\top \tilde{\V{T}}\normal \, d\surface \quad \text{with} \quad \tilde{\V{T}} = \frac{\sum_{i=1}^{\cameraNum}\visibility \tangent_{i}\tangent_{i}^\top}{\sum_{i=1}^{\cameraNum}\visibility},
	\label{eq.functional_simple}
\end{equation}
where we omit the dependence on the surface point \point for clarity.
As discussed in \cref{sec.functional},  accurate surface points and normals are both necessary to minimize the functional.

We represent the surface implicitly using a signed distance function (SDF) and optimize the SDF based on the framework of implicit differentiable renderer (IDR)~\cite{idr2020multiview}.
We parameterize the SDF by a multi-layer perceptron~(MLP) as~$f(\V{x};\Vg{\theta}):\R^3\times \R^{d}\rightarrow \R$, where $\point \in \R^3$ is the 3D point coordinate, and $\Vg{\theta} \in \R^d$ are MLP parameters.
The surface \surface is implicitly represented as the zero-level set of the SDF
\begin{equation}
	\surface(\Vg{\theta}) = \{\V{x} \mid f(\V{x}; \Vg{\theta})=0 \},
\end{equation} 
which varies depending on the MLP parameters.

To optimize the MLP, we use a loss function that consists of the tangent space consistency loss, the silhouette loss, and the Eikonal regularization:

\begin{equation}
	\loss = \loss_{\textrm{TSC}} + \lambda_1 \loss_{\textrm{silhouette}} + \lambda_2 \loss_{\textrm{Eikonal}}.
	\label{eq.total_loss}
\end{equation}

In each batch of the optimization, we randomly sample a set of \batchsize pixels from all views, cast camera rays from these pixels into the scene, and find the first ray-surface intersections. We evaluate the TSC loss for pixels with ray-surface intersections located inside the silhouette, denoted as $\V{X}$. We evaluate the silhouette loss for pixels that do not have ray-surface intersections or are located outside the silhouette, denoted as $\V{\tilde{X}}$.

\input{sections/figures_tables/TSC_loss.tex}
\vspace{-1em}
\paragraph{Tangent space consistency loss}
Based on \cref{eq.functional_simple}, we define the TSC loss as
\begin{equation}
	\loss_{\textrm{TSC}} = \frac{1}{\batchsize} \sum_{\point \in \V{X}} \normal(\point;\Vg{\theta})^\top \tilde{\V{T}}(\point) \normal(\point;\Vg{\theta}).
\end{equation}
To evaluate the TSC loss, we need to evaluate the surface normal and construct the matrix $\tilde{\V{T}}$.
According to the property of SDF~\cite{osher2004level}, the surface normal direction is the gradient evaluated at a zero-level set point:
\begin{equation}
	\V{n}(\V{x};\Vg{\theta}) = \nabla_{\V{x}}f(\V{x};\Vg{\theta}).
\end{equation}
Here, the surface normal can still be represented analytically as the MLP parameters~\cite{igr2020icml,siren2020sitzmann}.
Therefore, the gradient of the loss functions can be backpropagated to MLP parameters via surface normals.

We then compute $\V{T}(\point)$ for the point \point from all visible views. 
First, we project the surface points onto all views and check their visibility in each, as shown in \cref{fig.neuralSDF_optimization}.
To determine the visibility, we march the surface points toward the camera center and check whether there is a negative distance on the ray; see the supplementary material for more details.
Then in visible views, we compute the projected tangent vectors from input azimuth maps.

\vspace{-1em}
\paragraph{Silhouette loss}
Following IDR~\cite{idr2020multiview}, we use the input masks to constrain the visual hull of the shape\footnote{IDR~\cite{idr2020multiview} refers to it as mask loss, but we prefer to use ``silhouette loss'' after shape-from-silhouette~\cite{visualhull1994}.}.
We find the minimal distance on the rays for pixels that do not have ray-surface intersections, denoted as $f^*$.
The silhouette loss is then
\begin{equation}
	\loss_{\textrm{silhouette}} = \frac{1}{\alpha \batchsize} \sum_{\point \in \V{\tilde{X}}} \Psi \bigl(\mask(\Pi(\point)), \sigma(\alpha f^*)\bigr),
\label{eq.silhouette_loss}
\end{equation}
where $\Psi$ is the cross entropy function, and $\sigma(\cdot)$ is a sigmoid function with $\alpha$ controlling its sharpness.
\vspace{-1em}
\paragraph{Eikonal regularization}
Following IGR~\cite{igr2020icml}, we use the Eikonal loss to regularize the gradient of SDF such that the gradient norm is close to $1$ everywhere~\cite{osher2004level}:
\begin{equation}
	\loss_{\textrm{Eikonal}} = \mathbb{E}_\point \left( \left(\norm{\V{n}}_2 - 1\right)^2 \right).
\end{equation}
To apply Eikonal regularization, we randomly sample points within the object bounding box and compute the mean squared deviation from $1$-norm.

None of the three loss functions explicitly constrain the surface points.
It is the TSC loss that implicitly encourages good correspondence.

%% file: sections/figures_tables/TSC_loss.tex
\begin{figure}[t]
	\centering
	\includegraphics[width=\linewidth]{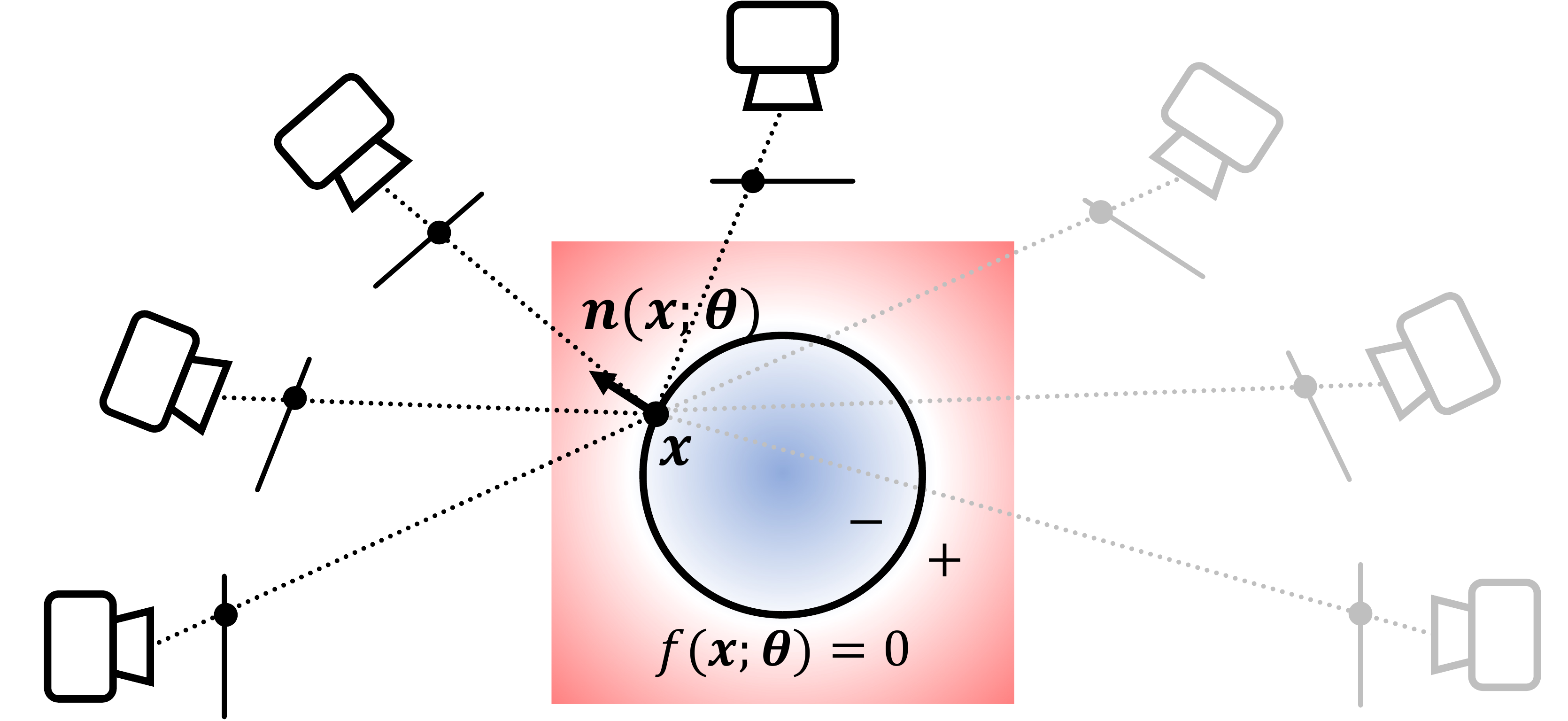}
	\caption{
		To optimize the neural SDF, we project the surface point onto all views and enforce the surface normal to be perpendicular to all visible \projectedTangentVectors.
  }
	\label{fig.neuralSDF_optimization}
\end{figure}

%% file: sections/04_experiments.tex
\section{Experiments}
\label{sec.experiments}
We evaluate \mvas in three experiments: comparing with MVPS methods quantitatively for surface and normal reconstruction in \cref{sec.exp_mvps}, applying \mvas to a photometric stereo method which struggles with \zenith estimation in \cref{sec.exp_symps}, and using \mvas with passive polarization imaging in \cref{sec.exp_polar}.
Implementation details are in the supplementary material.

\subsection{\mvas versus MVPS}
\label{sec.exp_mvps}
\paragraph{Baselines} 
We assess \mvas against multiple MVPS methods using the \diligentmv benchmark~\cite{li2020multi}. The MVPS methods include the coarse mesh refinement method \rmvps~\cite{park2016robust}, the benchmark method \bmvps~\cite{li2020multi}, the depth-normal fusion-based method \uanet~\cite{kaya2022uncertainty}, and the neural inverse rendering method \psnerf~\cite{yang2022psnerf}. \diligentmv~\cite{li2020multi} captures $20$ views under $96$ different lights for five objects. We use $15$-view azimuth maps for optimization and leave out $5$ views for testing, following \psnerf~\cite{yang2022psnerf}. The azimuth maps are computed from the normal maps estimated by the self-calibrated photometric stereo method \sdps~\cite{chen2019SDPS_Net}. 
\vspace{-1em}
\paragraph{Evaluation metrics} 
We use Chamfer distance (CD) and F-score for geometry accuracy~\cite{kaya2022uncertainty,knapitsch2017tanks}, and mean angular error (MAE) for normal accuracy~\cite{yang2022psnerf}. 
For CD and F-score, we only consider visible points by casting rays for all pixels and finding the first ray-mesh intersections\footnote{Different strategies for computing CD yield different results to the original papers. \uanet crops the invisible bottom face and uses mesh vertices~\cite{kaya2022uncertainty}; \psnerf~\cite{yang2022psnerf} samples $10000$ points from the mesh surface. }.
\vspace{-1em}
\paragraph{Results and discussions}
\Cref{tab.chamfer_mvps_kaya} reports the geometry accuracy of the recovered \diligentmv surfaces.
\bmvps~\cite{li2020multi} achieves the best scores in $4$ objects due to the usage of calibrated light information.
\uanet~\cite{kaya2022uncertainty} distorts the surface reconstruction by not considering the multi-view consistency.
\mvas outperforms \psnerf~\cite{yang2022psnerf} in $3$ objects without modeling the rendering process.

\Cref{fig.vis_comp_mvps} visually compares recovered ``Buddha'' and ``Reading'' objects.
Despite not having the best numerical scores, our method produces comparable results.
Lower scores for these objects are mainly due to our method's sensitivity to inaccurate silhouette masks provided by \diligentmv~\cite{li2020multi}.
We project the GT surface onto the image plane and find up to $10$-pixel inconsistency between the projected region and the GT mask.
Thus, the silhouette loss \cref{eq.silhouette_loss} encourages our reconstructed surfaces to shrink to align with the smaller silhouettes.

Our method requires less effort for shape recovery than \bmvps~\cite{li2020multi} and \psnerf~\cite{yang2022psnerf}. 
While \bmvps~\cite{li2020multi} calibrates $96$ light directions and intensities, we use a self-calibrated PS method for input azimuth maps. 
\psnerf~\cite{yang2022psnerf} uses $15$ view $\times$ $96$ light $=1440$ images to optimize multiple MLPs that model shape and appearance, which requires a high computational cost.
It takes \psnerf~\cite{yang2022psnerf} over $20$ hours per object on an RTX 3090 GPU.
In contrast, our approach optimizes a single MLP with $15$ azimuth maps, taking approximately $3$ hours per object on an RTX 2080Ti GPU.

\Cref{tab.mae_normal} reports MAE for $5$ test and all $20$ viewpoints, and \cref{fig.comp_mvps_normal} visually compares recovered normal maps. \mvas improves normal accuracy compared to \sdps~\cite{chen2019SDPS_Net} and outperforms \psnerf~\cite{yang2022psnerf} in $4$ objects, demonstrating TSC's effectiveness in constraining surface normals from multi-view observations.
Since \tsc imposes a direct constraint on surface normals, it is more effective than modeling a rendering process as in \psnerf~\cite{yang2022psnerf}.

\input{sections/figures_tables/cd_mvps}
\input{sections/figures_tables/compare_mvps}
\input{sections/figures_tables/mae_normal.tex}
\input{sections/figures_tables/visualization_normal.tex}
\input{sections/figures_tables/symps_setup.tex}
\input{sections/figures_tables/mvs_comparison}
\input{sections/figures_tables/pandora_horizontal}

\subsection{\mvas for symmetric-light photometric stereo}
\label{sec.exp_symps}
Some photometric stereo methods can estimate azimuth angles well but struggle with zenith angles\cite{chandraker2012differential, minami2022symmetric}. This section shows how \mvas can be used for an uncalibrated photometric stereo setup to eliminate the need for tedious zenith estimation while allowing full surface reconstruction. 

We use the setup shown in \cref{fig.symps_setup} to obtain multi-view azimuth maps.
We place four lights symmetrically around the camera and the target object on a rotation table.
In each view, we capture one ambient-light image and four lit images.
The ambient-light images are used for SfM~\cite{schoenberger2016sfm} to obtain the camera poses and are input to MVS~\cite{schoenberger2016mvs} for comparison.
Using the four lit images, The azimuth angles can be trivially computed from the ratio of the vertical to the horizontal difference image~\cite{minami2022symmetric}.

\Cref{fig.mvs_comparison} compares reconstructed surfaces and normals by Colmap~\cite{schoenberger2016mvs} and \mvas.
The first object shows a scene with challenging white planar faces. Photo-consistency-based MVS fails to recover the textureless region, while \tsc succeeds in the planar region. This is possibly due to that \tsc can still determine surface normals with wrong correspondences in a planar region, as discussed in~\cref{sec.functional}.
The second object has a dark surface, which is also challenging for photo-consistency, and Colmap~\cite{schoenberger2016mvs} struggles to recover the correct surface normals.

\subsection{\mvas with polarization imaging}
\label{sec.exp_polar}
This section shows the application of \mvas on azimuth maps obtained passively by a snapshot polarization camera, which makes the capture process as simple as MVS.
Since \tsc is $\pi$-invariant, \mvas eliminates the need to correct the $\pi$-ambiguity~\cite{miyazaki2003polarization}.
\Cref{fig.pandora} compares the surface and normal reconstruction on the multi-view polarization image dataset~\cite{dave2022pandora}.
We input the color images into Colmap~\cite{schoenberger2016mvs} and reproduce the results of the polarimetric inverse rendering method PANDORA using their codes~\cite{dave2022pandora}.
We modify our TSC loss to account for \halfpi ambiguity in polar-azimuth maps; see the supplementary material for details.

As shown in \cref{fig.pandora}, MVS~\cite{schoenberger2016mvs} breaks down for highly specular objects.
Polar-azimuth observations are robust to such specularity and allow \mvas for faithful reconstruction.
The comparison to PANDORA~\cite{dave2022pandora} shows that surfaces can be recovered without considering the degree of polarization or reflectance-light modeling.

%% file: sections/figures_tables/cd_mvps.tex
\begin{table}
	\scriptsize
	\definecolor{Gray}{gray}{0.85}
	\newcolumntype{g}{>{\columncolor{Gray}}r}
	\centering
	\caption{\textbf{(Top)} Chamfer distance ($\downarrow$) and \textbf{(Bottom)} F-score~($\uparrow$)~\cite{kaya2022uncertainty,knapitsch2017tanks} of recovered geometry on \diligentmv benchmark~\cite{li2020multi}.}
	\resizebox{\linewidth}{!}{ 
		\begin{tabular}{@{}lrrrrrg}
			\toprule
			& Bear & Buddha & Cow & Pot2 & Reading & Average \\
			\midrule
			\rmvps~\cite{park2016robust}  & 1.070 & 0.397 & 0.440 & 1.504 & 0.561 & 0.794  \\
			\bmvps~\cite{li2020multi} & \textbf{0.212} & \textbf{0.254} &\textbf{0.091} & 0.201 & \textbf{0.259} &  \textbf{0.203}\\
			\uanet~\cite{kaya2022uncertainty} &  0.414 & 0.452 & 0.326 & 0.414 & 0.382 & 0.398\\
			\psnerf~\cite{yang2022psnerf} & 0.260 & 0.314 & 0.287 & 0.254 & 0.352 & 0.293\\ 
			\mvas (ours) & 0.243  & 0.357 & 0.216 & \textbf{0.197} & 0.522 & 0.307\\
			\midrule
   			\rmvps~\cite{park2016robust}  & 0.262 & 0.698 & 0.760 & 0.198 & 0.519 & 0.487  \\
			\bmvps~\cite{li2020multi} & \textbf{0.958} & \textbf{0.902} & \textbf{0.986} & 0.946 & \textbf{0.914} & \textbf{0.941}\\
			\uanet~\cite{kaya2022uncertainty} & 0.707 & 0.669 & 0.798 & 0.731 & 0.762  & 0.733\\
			\psnerf~\cite{yang2022psnerf} & 0.898  & 0.806 & 0.856 & 0.919 & 0.785 & 0.853\\ 
			\mvas (ours) &0.909 & 0.754 & 0.907 & \textbf{0.962} & 0.546 & 0.816 \\
			\bottomrule
		\end{tabular}
	}
\label{tab.chamfer_mvps_kaya}
\end{table}

%% file: sections/figures_tables/compare_mvps.tex
\begin{figure}[t!]
	\tiny
	\newcommand{\figwidtha}{0.16}
	\centering
	\begin{tabular}{c@{}c@{}c@{}c@{}c@{}c}
		R-MVPS~\cite{park2016robust} & \bmvps~\cite{li2020multi} & \uanet~\cite{kaya2022uncertainty} & \psnerf~\cite{yang2022psnerf} & \mvas (ours)  & GT \\
		\includegraphics[width=\figwidtha\linewidth]{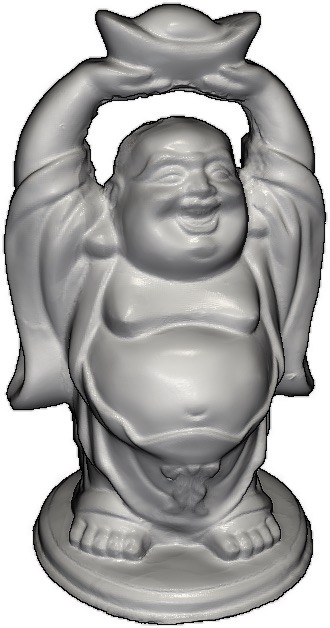} &
		\includegraphics[width=\figwidtha\linewidth]{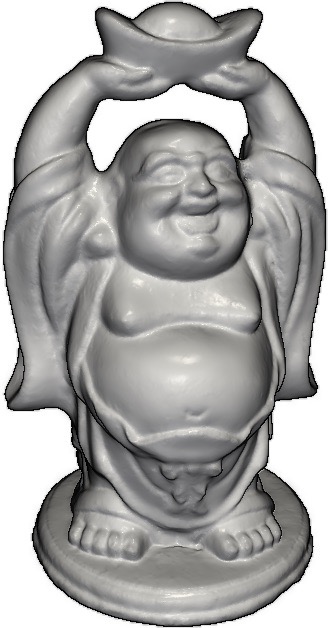} &
		\includegraphics[width=\figwidtha\linewidth]{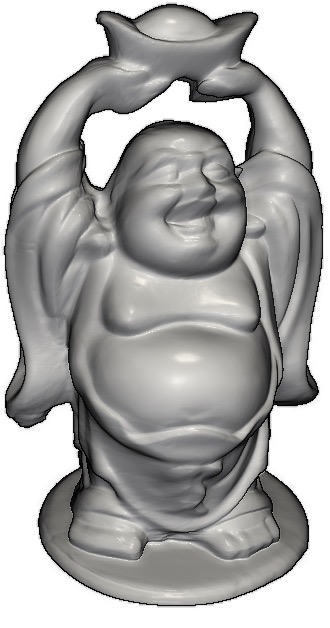} &
		\includegraphics[width=\figwidtha \linewidth]{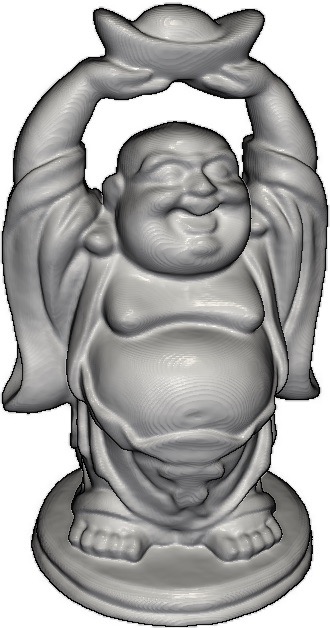} &
		\includegraphics[width=\figwidtha \linewidth]{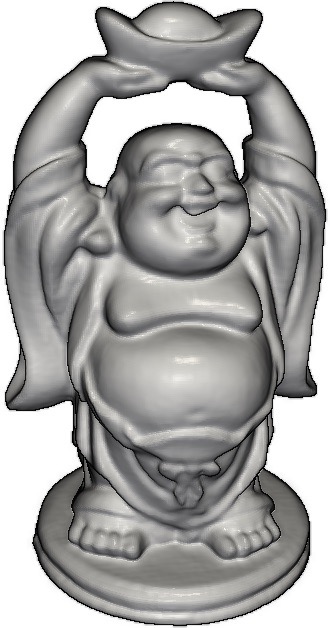} &
		\includegraphics[width=\figwidtha \linewidth]{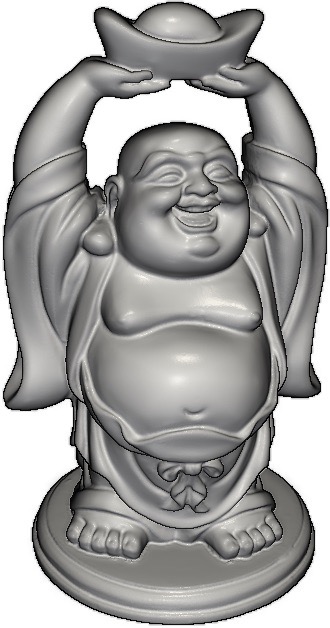}  \\
		
		\includegraphics[width=\figwidtha\linewidth]{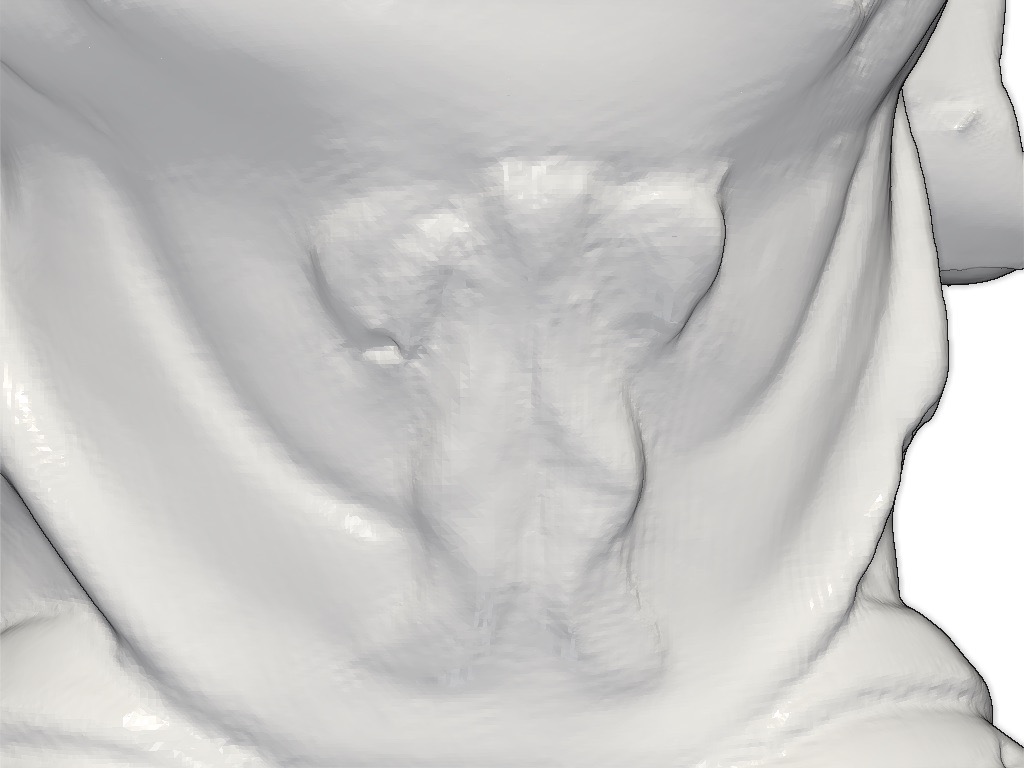} &
		\includegraphics[width=\figwidtha\linewidth]{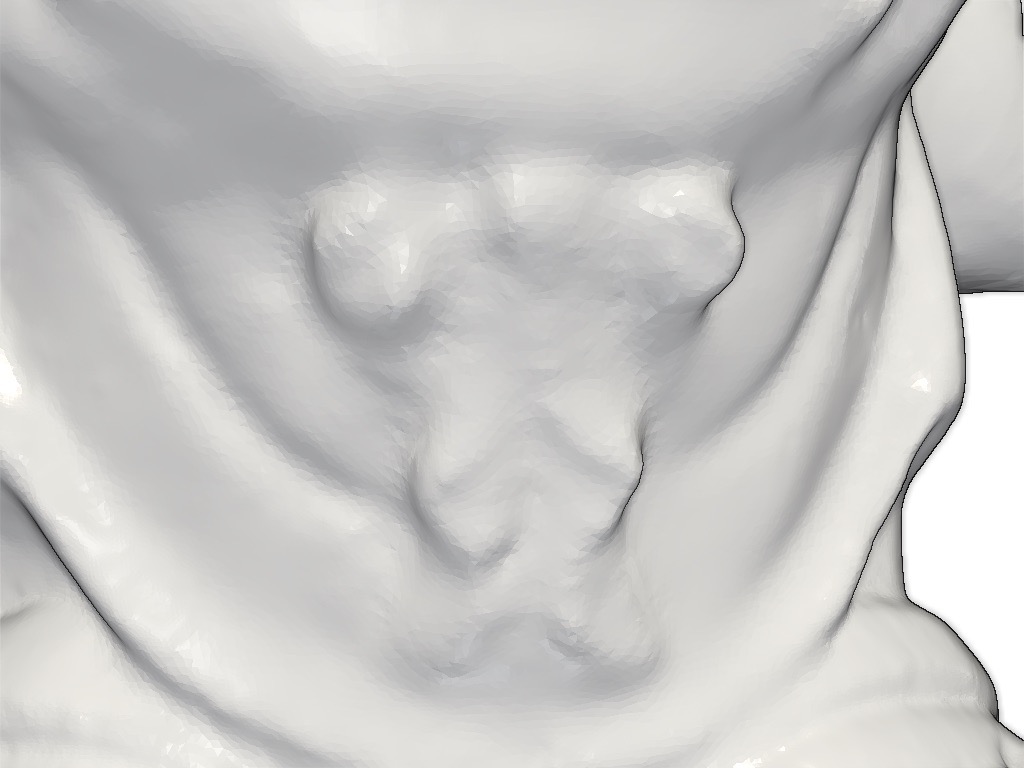} &
		\includegraphics[width=\figwidtha\linewidth]{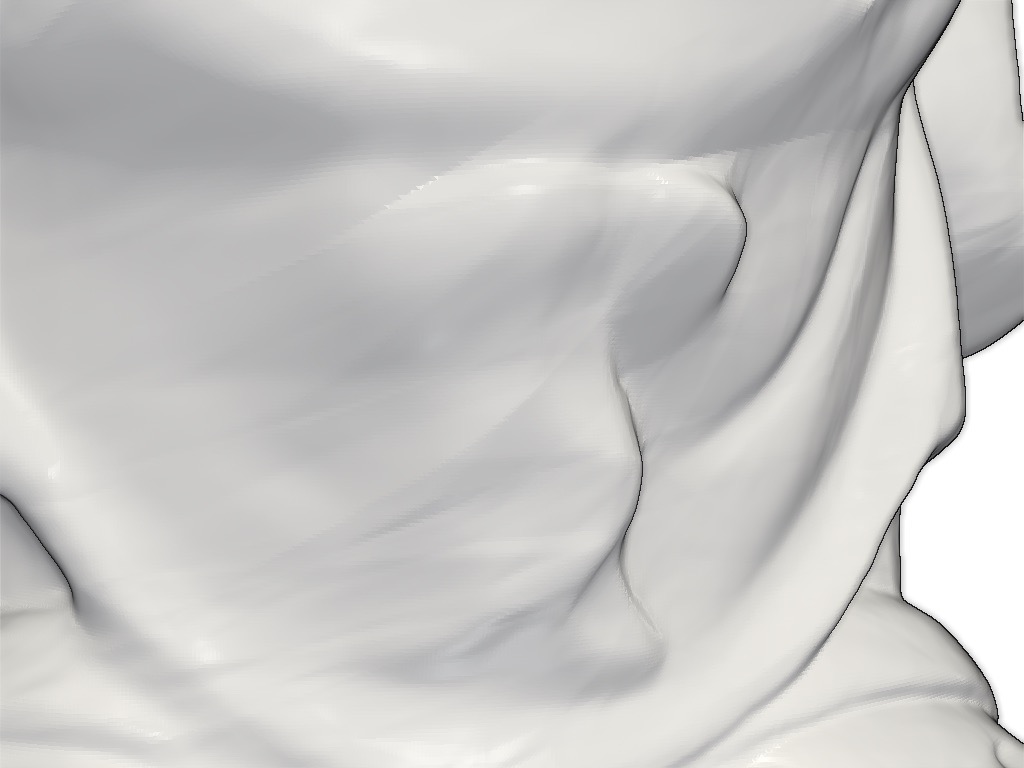} &
		\includegraphics[width=\figwidtha \linewidth]{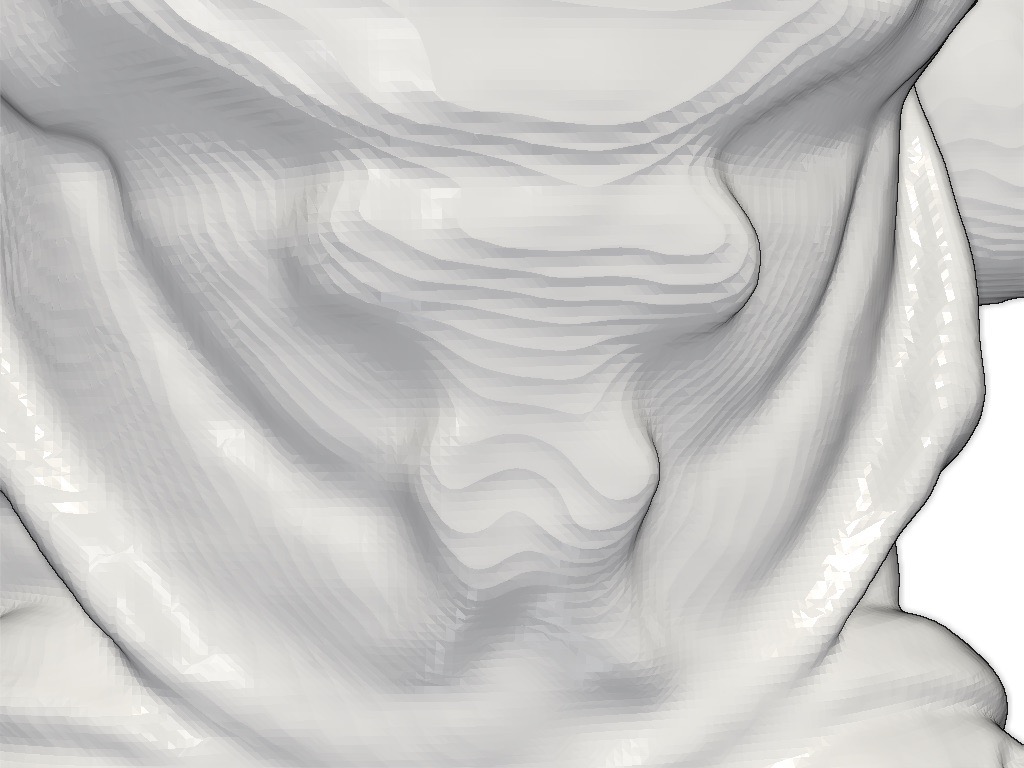} &
		\includegraphics[width=\figwidtha \linewidth]{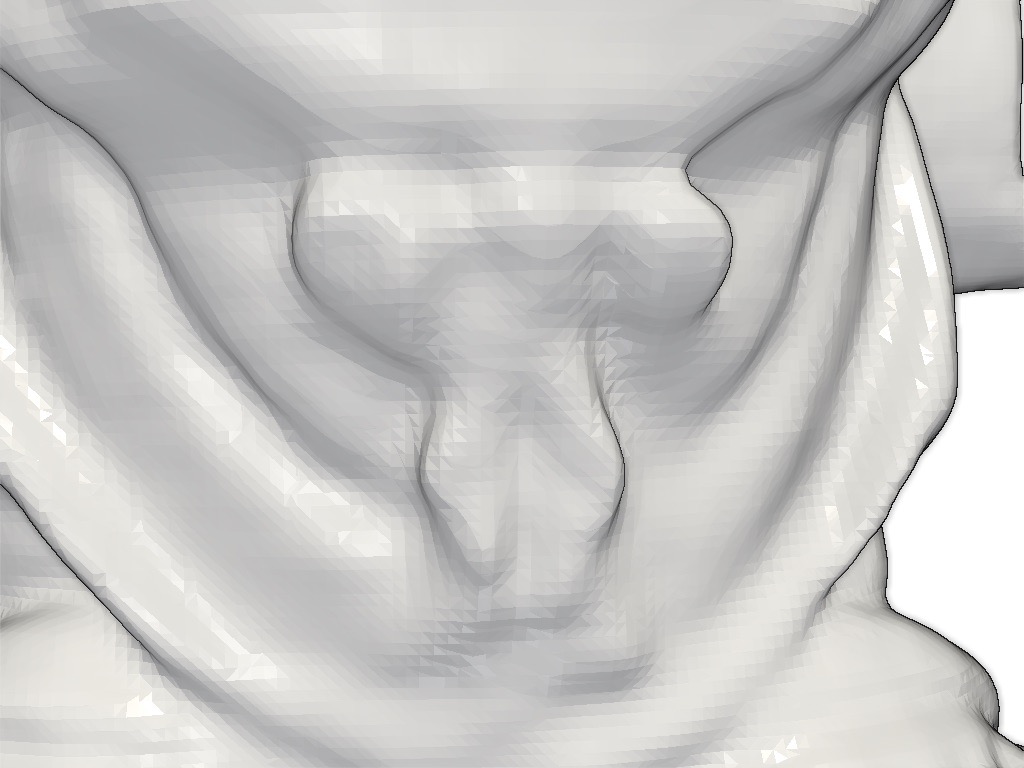} &
		\includegraphics[width=\figwidtha \linewidth]{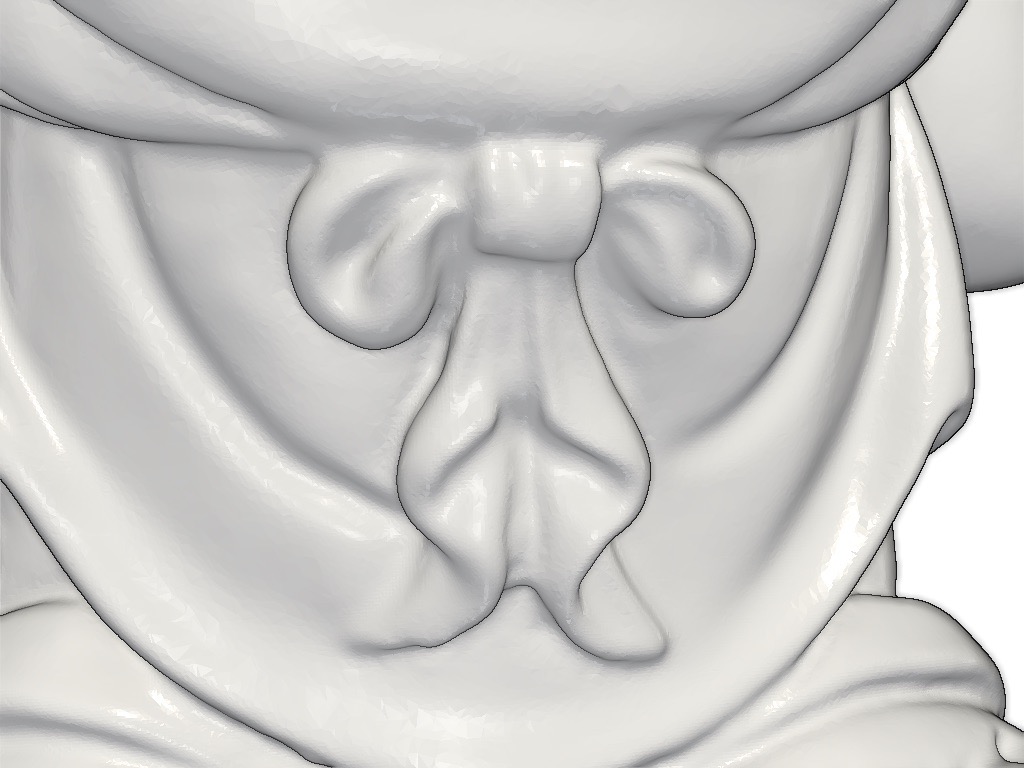} \\
		\includegraphics[width=\figwidtha\linewidth]{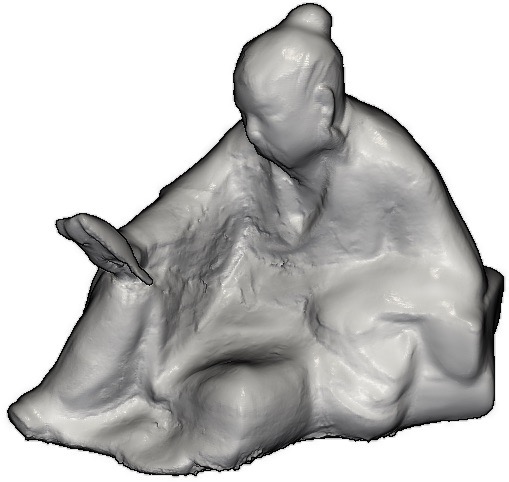} &
		\includegraphics[width=\figwidtha\linewidth]{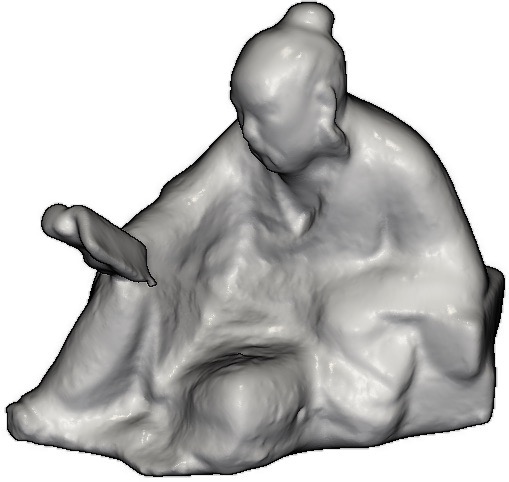} &
		\includegraphics[width=\figwidtha\linewidth]{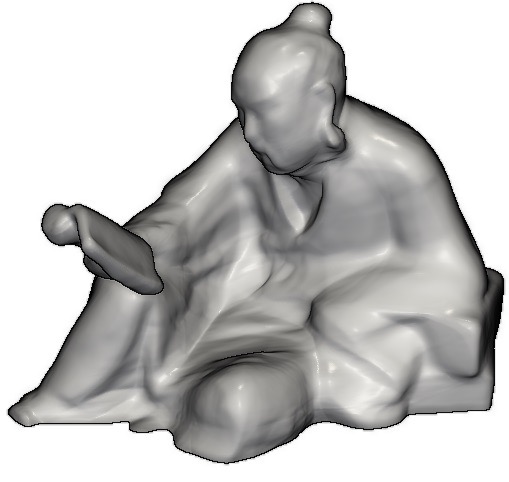} &
		\includegraphics[width=\figwidtha \linewidth]{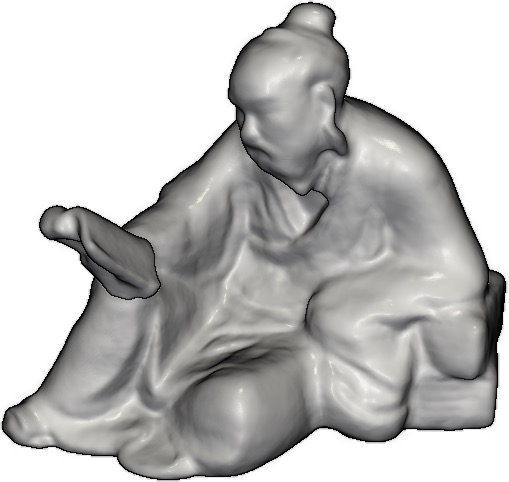} &
		\includegraphics[width=\figwidtha \linewidth]{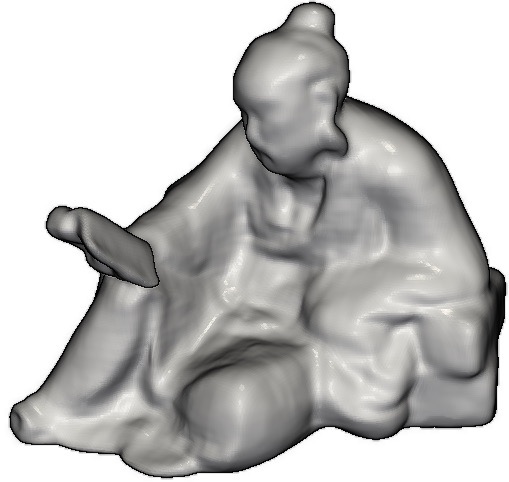} &
		\includegraphics[width=\figwidtha \linewidth]{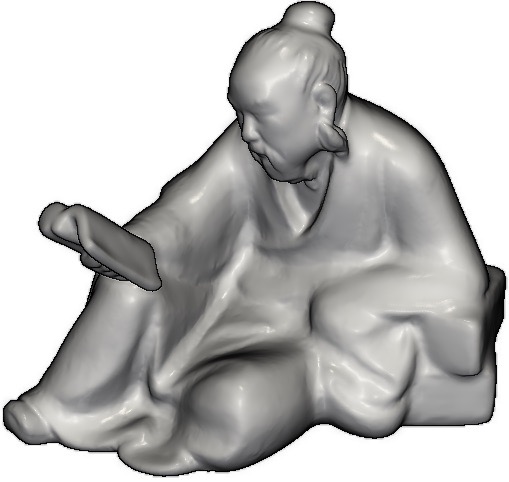} \\
		\includegraphics[width=\figwidtha\linewidth]{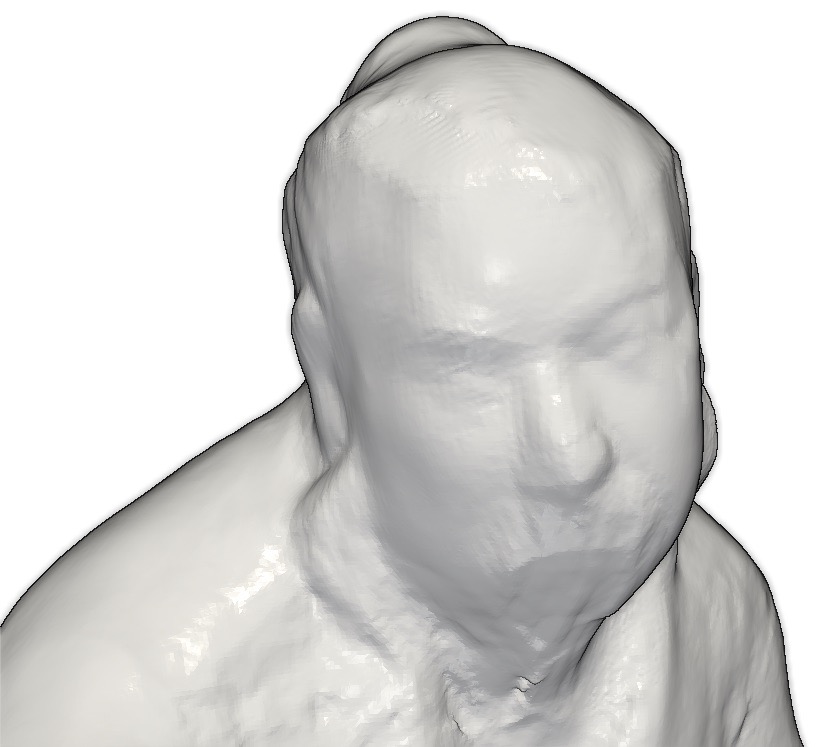} &
		\includegraphics[width=\figwidtha\linewidth]{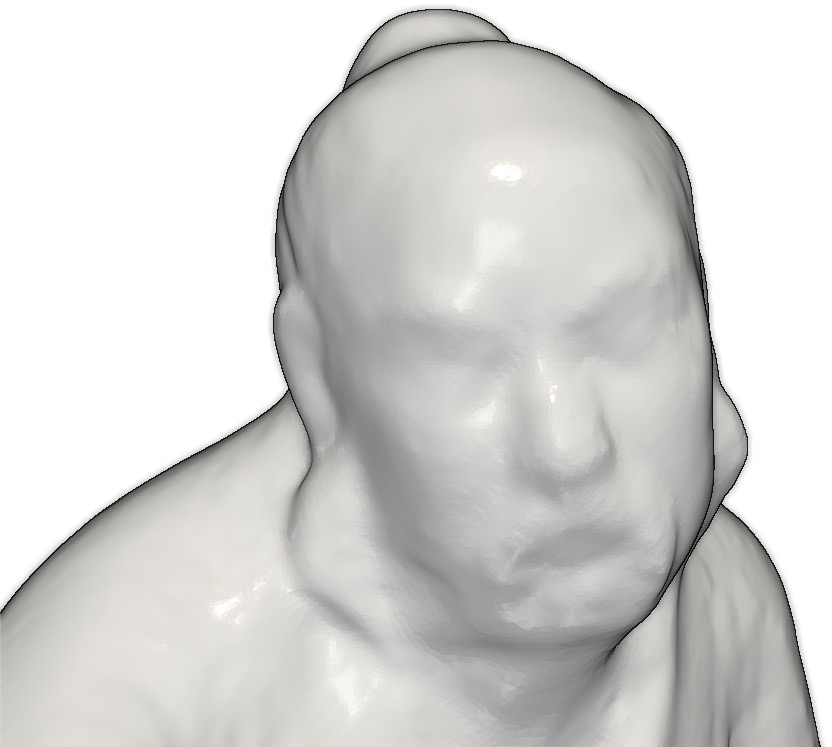} &
		\includegraphics[width=\figwidtha\linewidth]{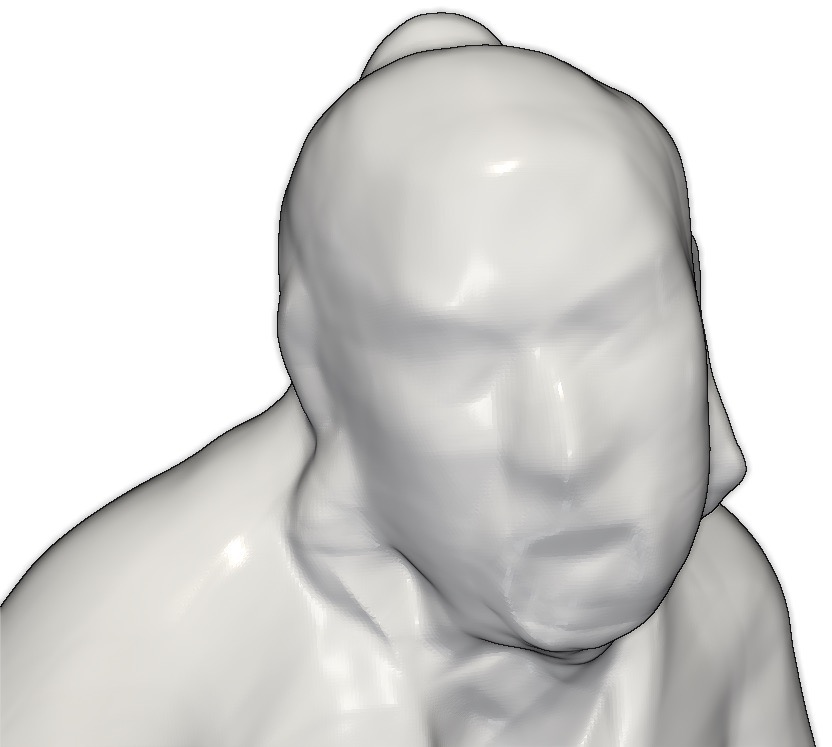} &
		\includegraphics[width=\figwidtha \linewidth]{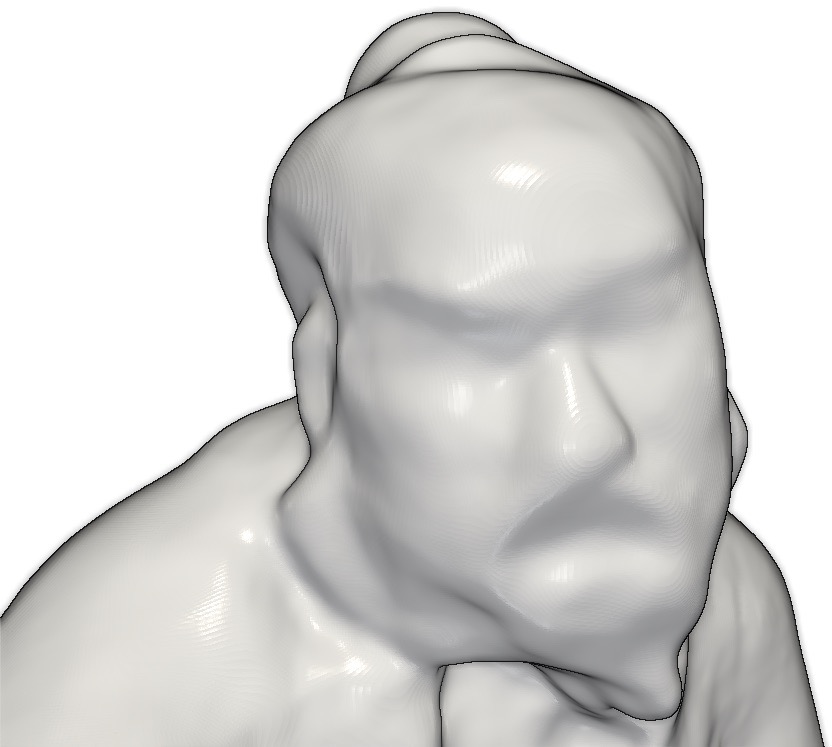} &
		\includegraphics[width=\figwidtha \linewidth]{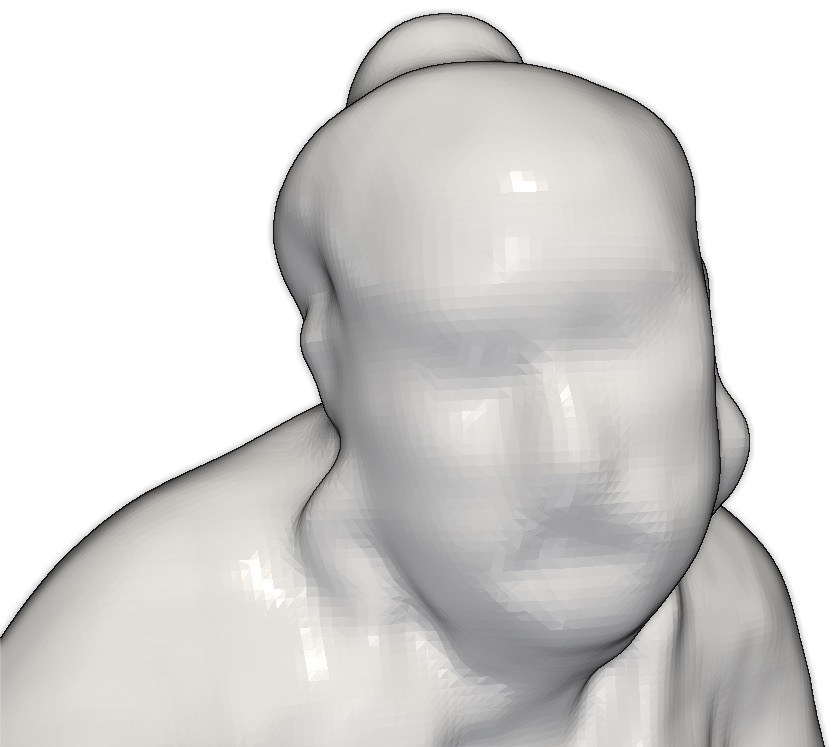} &
		\includegraphics[width=\figwidtha \linewidth]{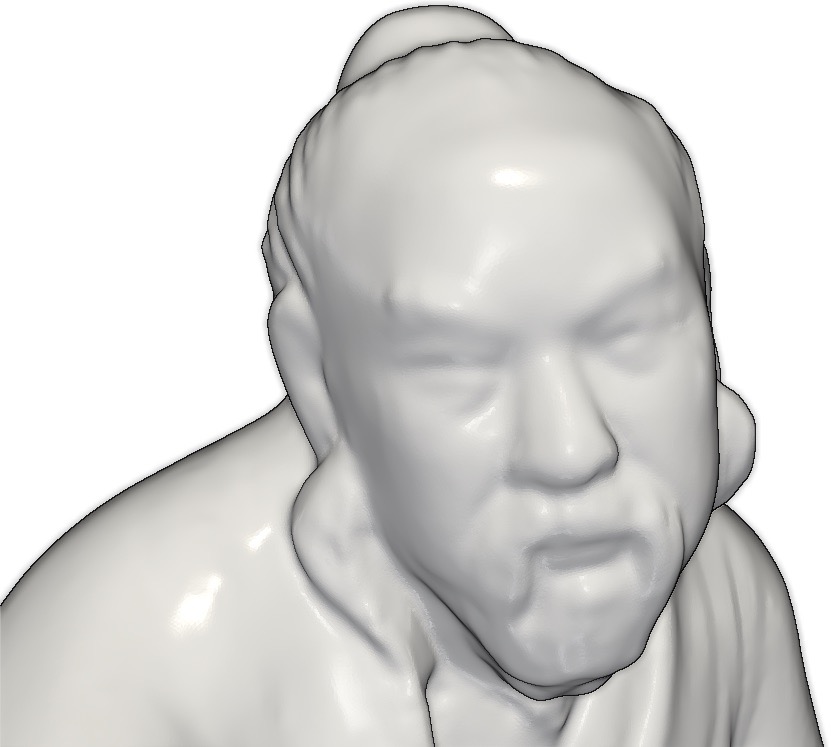}
	\end{tabular}
	\caption{Visual comparison of recovered geometry.  \rmvps~\cite{park2016robust}, \bmvps~\cite{li2020multi}, and \uanet~\cite{kaya2022uncertainty} require coarse geometry and use all 20 views for optimization, while \psnerf~\cite{yang2022psnerf} and ours use a sphere initialization and 15 views.}
\label{fig.vis_comp_mvps}
\end{figure}

%% file: sections/figures_tables/mae_normal.tex
\begin{table}
	\definecolor{Gray}{gray}{0.85}
	\newcolumntype{g}{>{\columncolor{Gray}}r}
	\centering
	\caption{Mean angular error ($\downarrow$) of recovered normal maps~\cite{li2020multi}, evaluated using \textbf{(Top)} $5$ test views and \textbf{(Bottom)} all $20$ views.}
	\resizebox{\linewidth}{!}{
		\begin{tabular}{@{}llrrrrrg}
			\toprule
			Methods& \# views & Bear & Buddha & Cow & Pot2 & Reading & Average \\
			\midrule
			\rmvps~\cite{park2016robust}  & \multirow{4}{*}{5}& 12.80 & 13.67 & 10.81 & 14.99 & 11.71 &  12.80 \\
			\bmvps~\cite{li2020multi}  &  & 3.80 & 10.57 & \textbf{2.83} & 5.76 & \textbf{6.90} & \textbf{5.97}\\
			\psnerf~\cite{yang2022psnerf} & & 3.45  & 10.25  & 4.35 & 5.94 & 9.36 & 6.67\\ 
			\sdps~\cite{chen2019SDPS_Net} && 7.59 & 11.16 & 9.46 & 7.95 & 16.16 & 10.46\\
			\mvas (ours) & & \textbf{3.08} & \textbf{9.90} & 3.72 & \textbf{5.07} &10.02& 6.36\\
			\midrule
			\rmvps~\cite{park2016robust}  & \multirow{4}{*}{20}& 12.70 & 13.63 & 10.92 & 14.91 & 11.79 &   12.79\\
			\bmvps~\cite{li2020multi}  &  & 3.81 & 10.58 & \textbf{2.86} & 5.72 & \textbf{6.98} & \textbf{5.99}\\
			\psnerf~\cite{yang2022psnerf} & & 3.32  & 10.55 & 4.21 & 5.88 &  8.97 & 6.59 \\ 
			\sdps~\cite{chen2019SDPS_Net} & & 7.72 & 11.03 & 9.65 & 8.14 & 15.59 & 10.42\\ 
			\mvas (ours) & &\textbf{3.09} & \textbf{9.78} & 3.74 & \textbf{5.04} &10.06& 6.34\\
			\bottomrule
		\end{tabular}
	}
	\label{tab.mae_normal}
\end{table}

%% file: sections/figures_tables/visualization_normal.tex
\begin{figure}
\scriptsize
\newcommand{\figwidthNormalVis}{0.16}
\centering
\begin{tabular}{@{}c@{}c@{}c@{}c@{}c@{}c@{}}
\rmvps~\cite{park2016robust} & \bmvps~\cite{li2020multi} & \psnerf~\cite{yang2022psnerf} & \sdps~\cite{chen2019SDPS_Net} & \mvas (ours) & GT \\
\includegraphics[width=\figwidthNormalVis\linewidth]{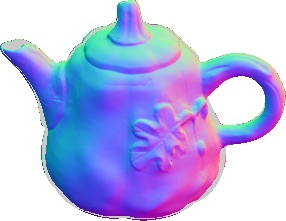}&
\includegraphics[width=\figwidthNormalVis\linewidth]{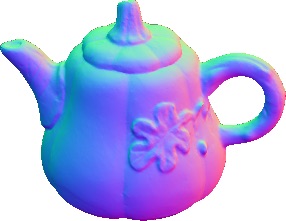}
&
\includegraphics[width=\figwidthNormalVis\linewidth]{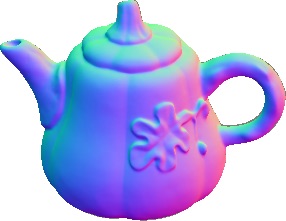}
&
\includegraphics[width=\figwidthNormalVis\linewidth]{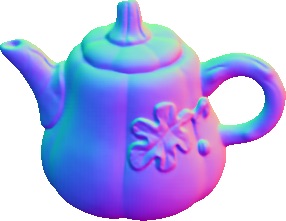}
&
\includegraphics[width=\figwidthNormalVis\linewidth]{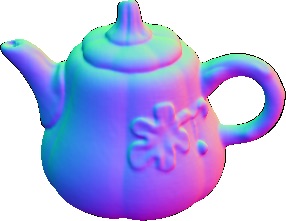}
& 
\includegraphics[width=\figwidthNormalVis\linewidth]{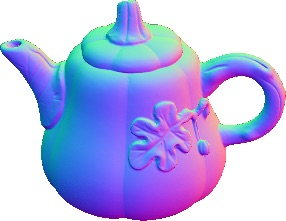}
\\
\includegraphics[width=\figwidthNormalVis\linewidth]{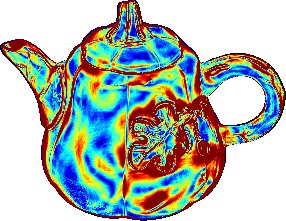}&
\includegraphics[width=\figwidthNormalVis\linewidth]{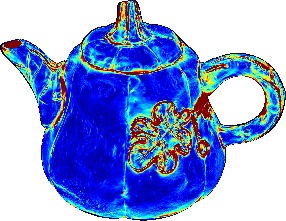}
&
\includegraphics[width=\figwidthNormalVis\linewidth]{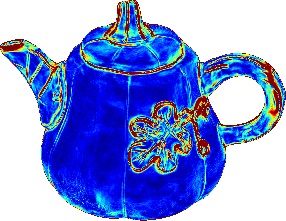}
&
\includegraphics[width=\figwidthNormalVis\linewidth]{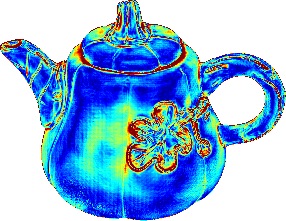}
&
\includegraphics[width=\figwidthNormalVis\linewidth]{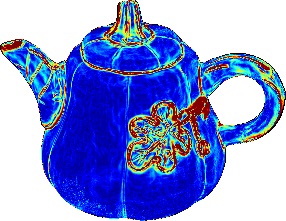}
& 
\colorbar{0.12}{$20^\circ$}{15}
\\
\includegraphics[width=\figwidthNormalVis\linewidth]{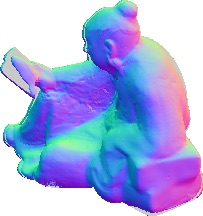}&
\includegraphics[width=\figwidthNormalVis\linewidth]{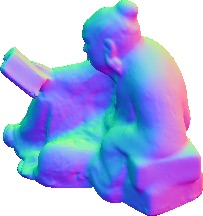}
&
\includegraphics[width=\figwidthNormalVis\linewidth]{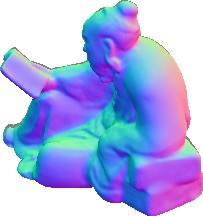}
&
\includegraphics[width=\figwidthNormalVis\linewidth]{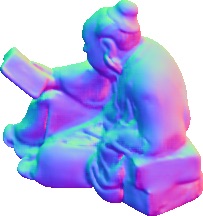}
&
\includegraphics[width=\figwidthNormalVis\linewidth]{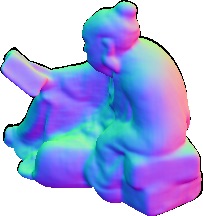}
& 
\includegraphics[width=\figwidthNormalVis\linewidth]{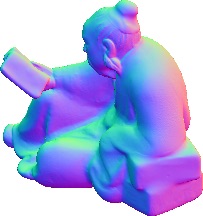}
\\
\includegraphics[width=\figwidthNormalVis\linewidth]{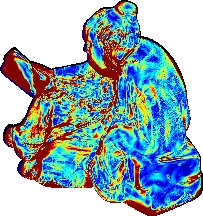}&
\includegraphics[width=\figwidthNormalVis\linewidth]{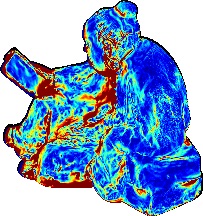}
&
\includegraphics[width=\figwidthNormalVis\linewidth]{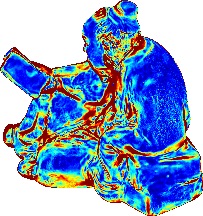}
&
\includegraphics[width=\figwidthNormalVis\linewidth]{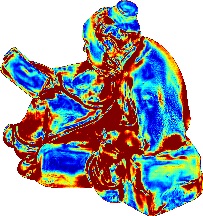}
&
\includegraphics[width=\figwidthNormalVis\linewidth]{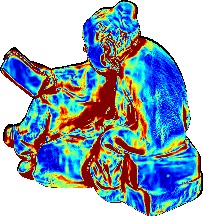}
& 
\colorbar{\figwidthNormalVis}{$20^\circ$}{25}
\end{tabular}
\caption{Visual comparison of recovered normal maps and angular error maps from the first view of \diligentmv~\cite{li2020multi} on the object ``Pot2'' and ``Reading.''}
\label{fig.comp_mvps_normal}
\end{figure}

%% file: sections/figures_tables/symps_setup.tex
\begin{figure}
	\centering
	\includegraphics[width=\linewidth]{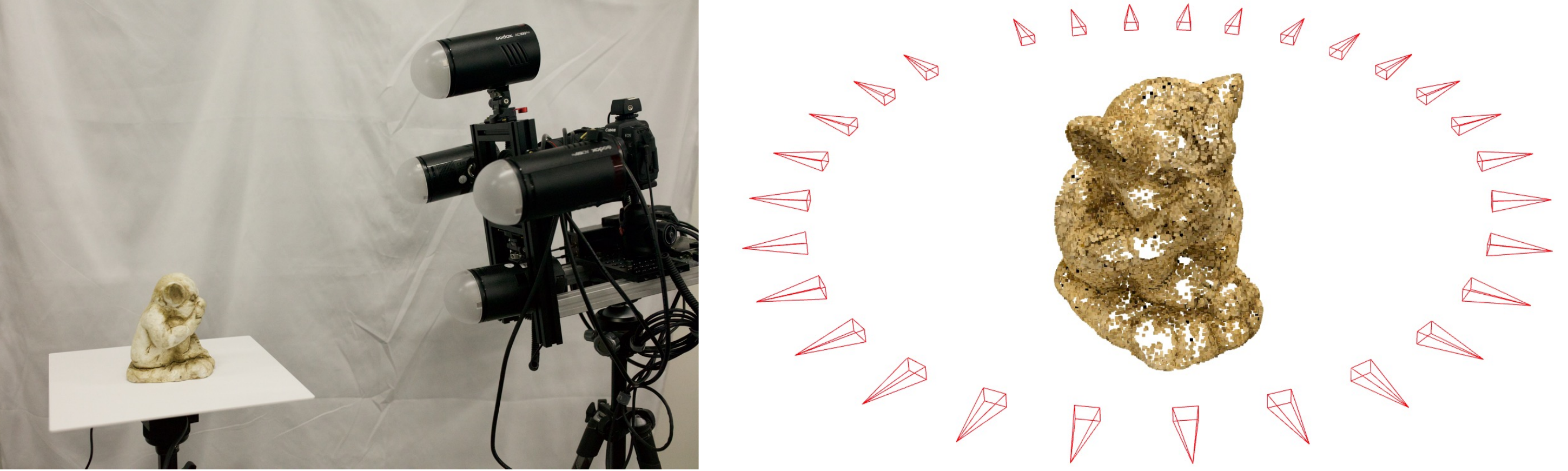}
	\caption{Our uncalibrated symmetric-light photometric stereo setup. Four lights are mounted symmetrically around the camera. We put the target object on a rotation table and capture about $30$ views $\times$ $5$ images in each view.}
\label{fig.symps_setup}
\end{figure}

%% file: sections/figures_tables/mvs_comparison.tex
\begin{figure*}
	\centering
	\scriptsize
	\newcommand{\figwidthB}{0.15}
	\begin{tabular}{@{}c@{}c@{}cc@{ }c@{ }c@{}}
		\begin{tabular}{c}
			Color images \& \\
			PS-azimuth~\cite{minami2022symmetric}
		\end{tabular} &  Colmap~\cite{schoenberger2016mvs} & \mvas (ours) & 		
		\begin{tabular}{c}
			Color images \& \\
			PS-azimuth~\cite{minami2022symmetric}
		\end{tabular}  & Colmap~\cite{schoenberger2016mvs} & \mvas (ours) \\
		\includegraphics[width=0.16\linewidth]{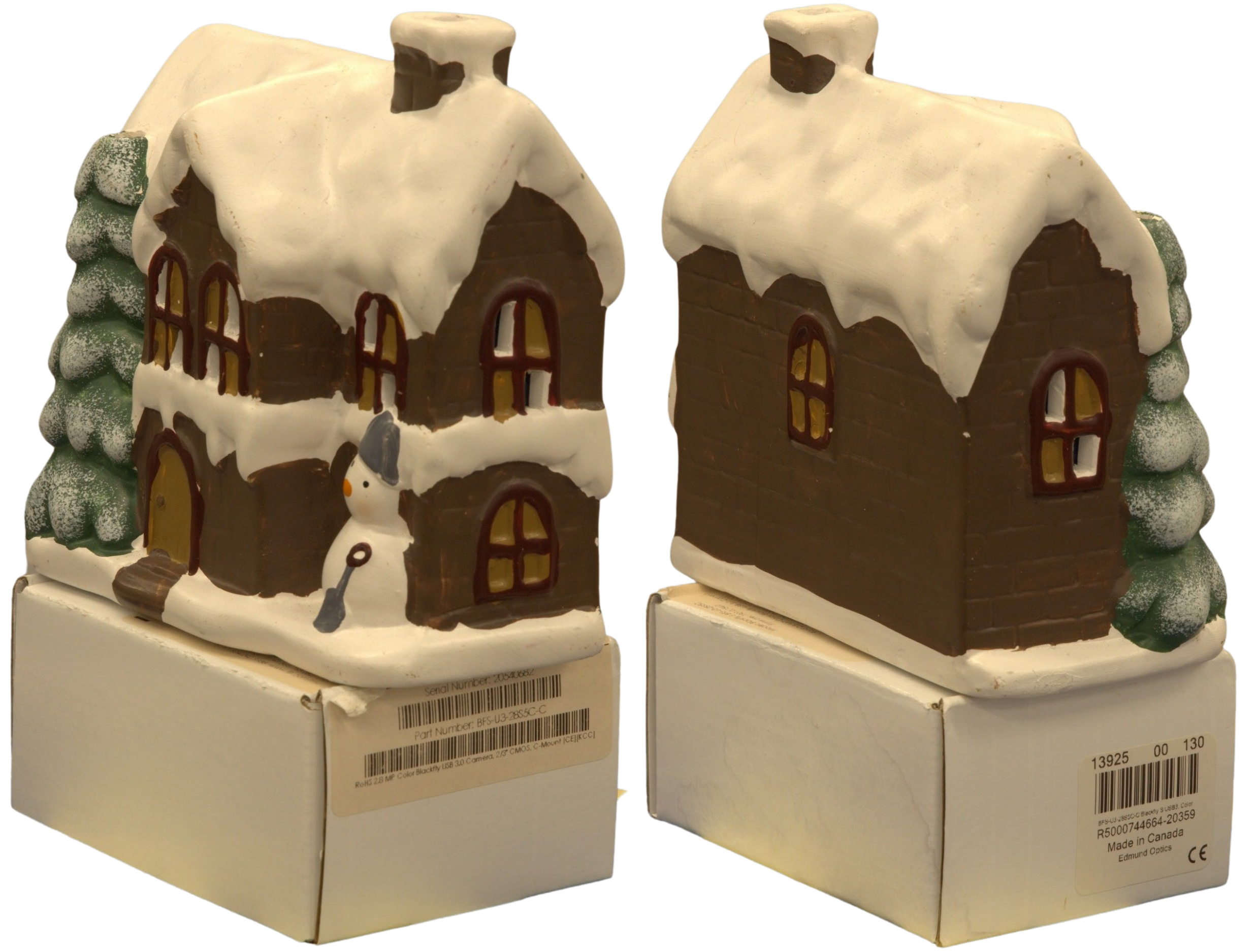} &
		\includegraphics[width=0.16\linewidth]{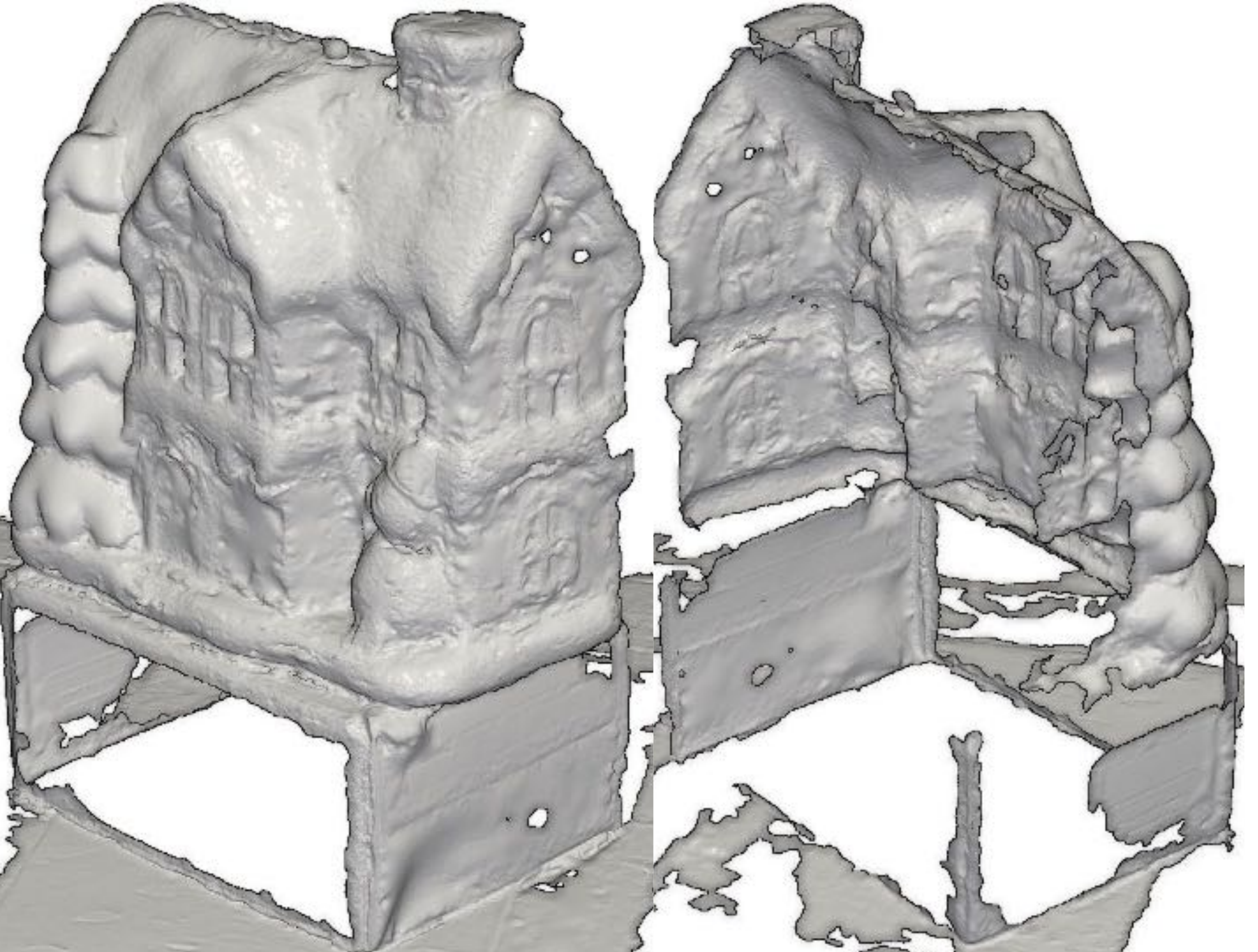} &
		\includegraphics[width=0.16\linewidth]{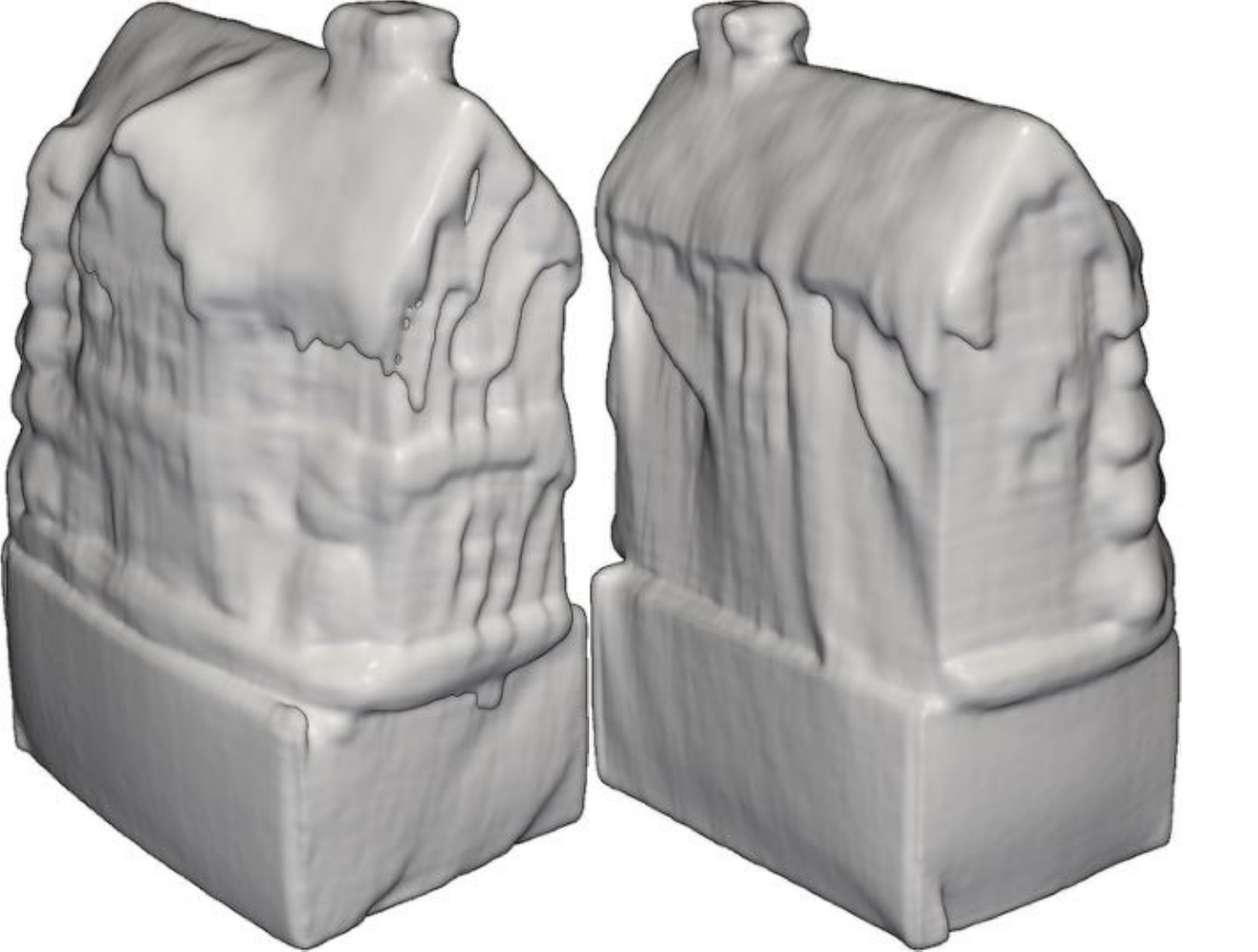} &
		\includegraphics[width=0.16\linewidth]{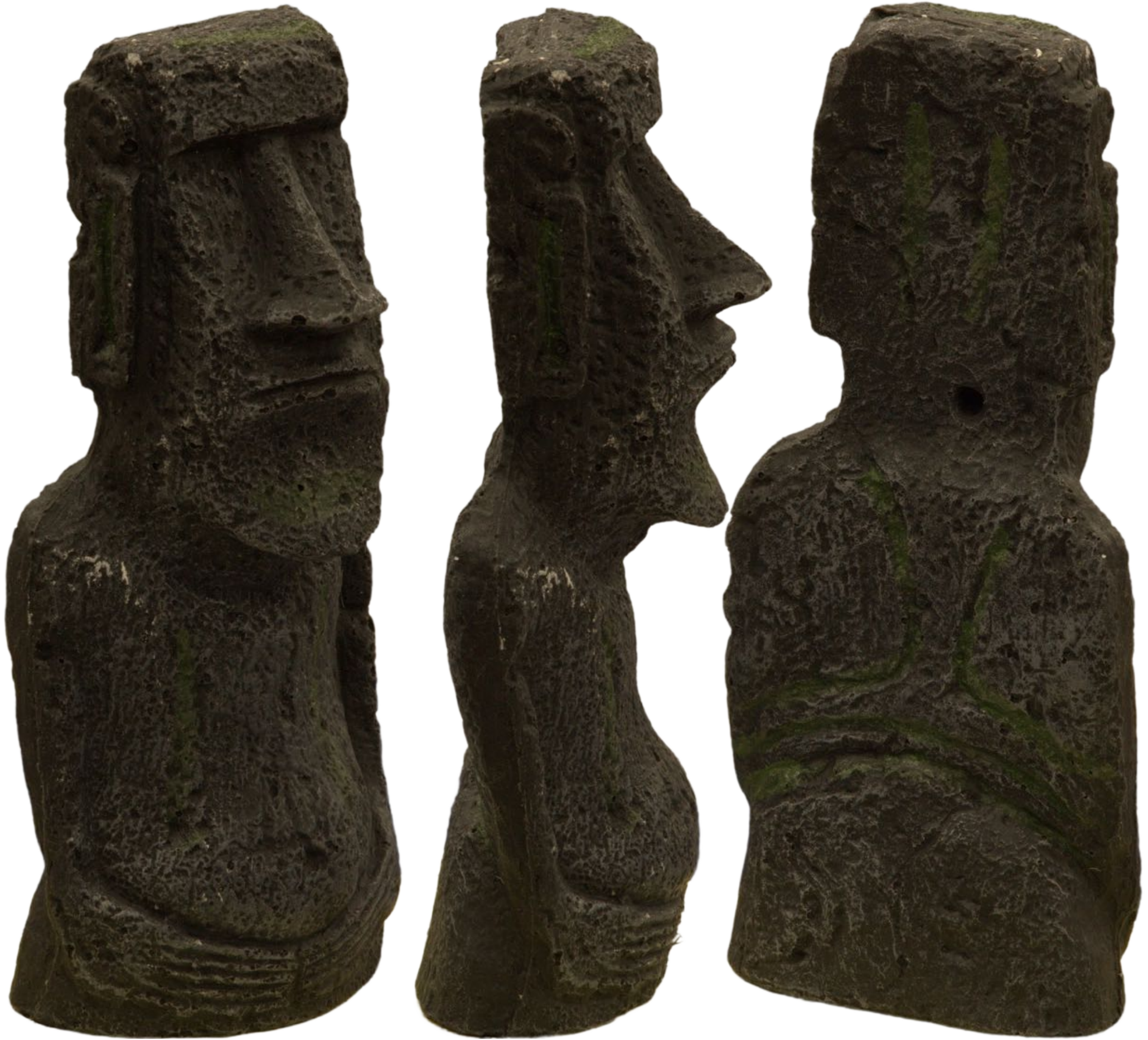} &
		\includegraphics[width=0.16\linewidth]{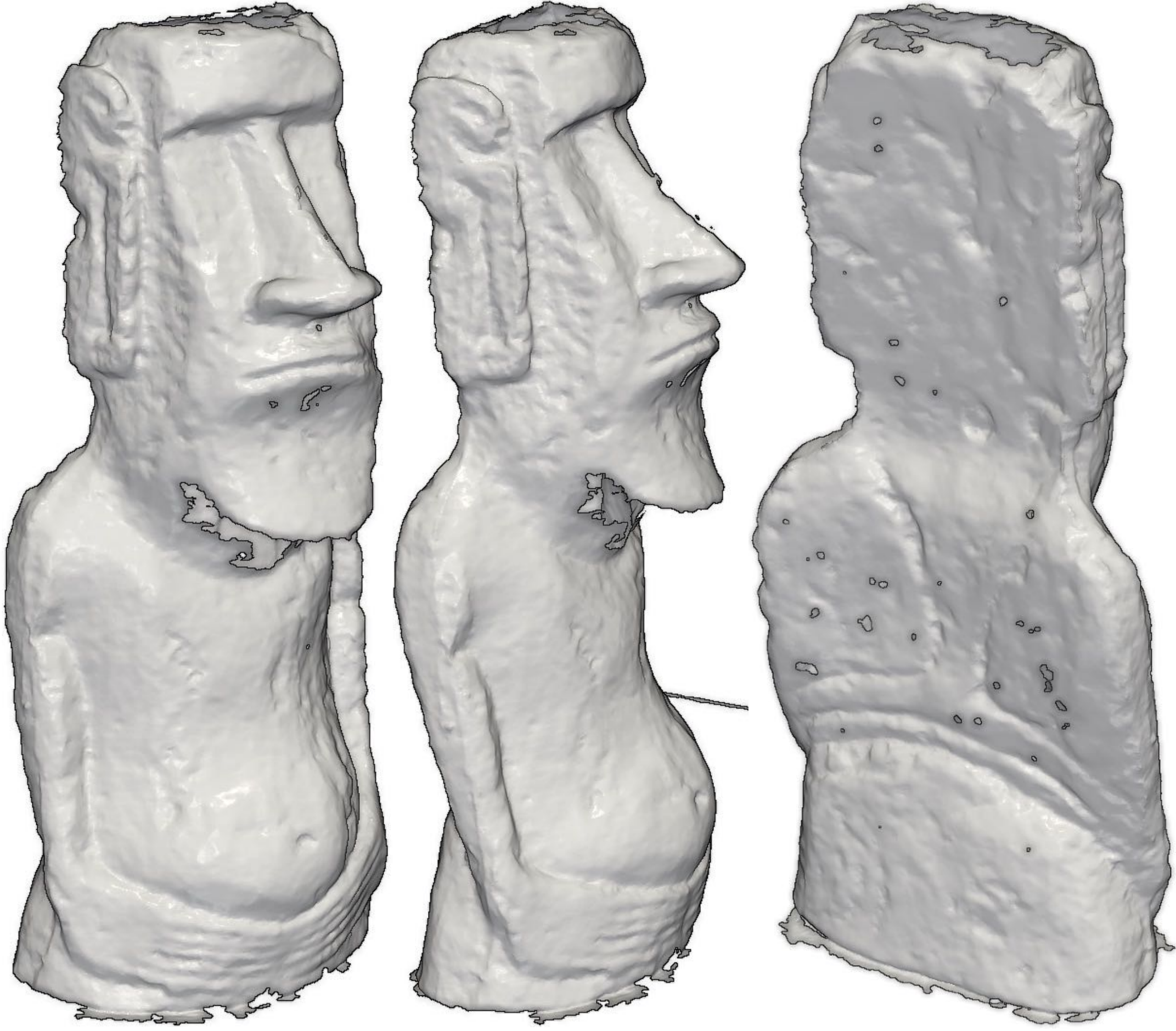} &
		\includegraphics[width=0.16\linewidth]{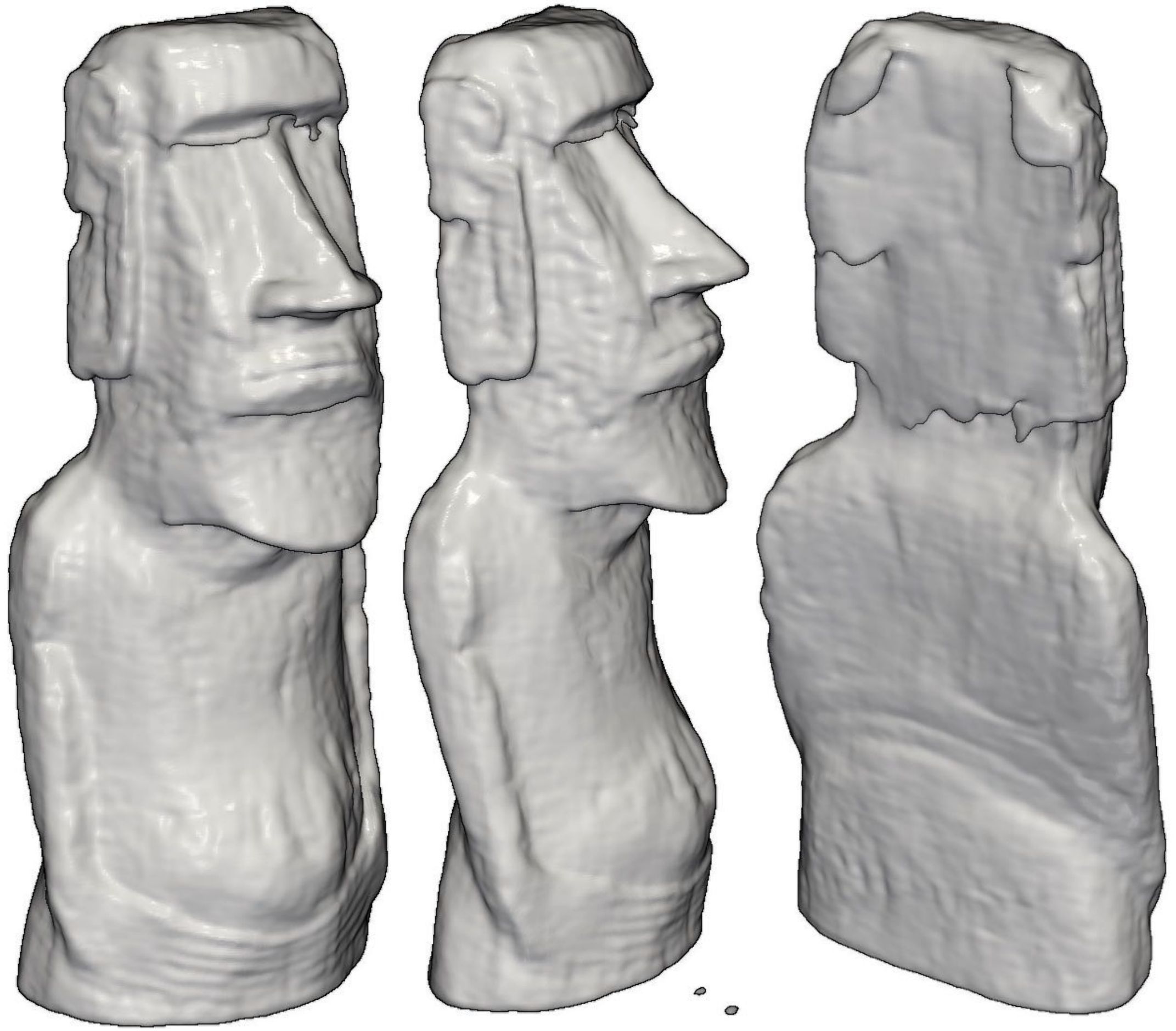} 
		\\
		\includegraphics[width=0.16\linewidth]{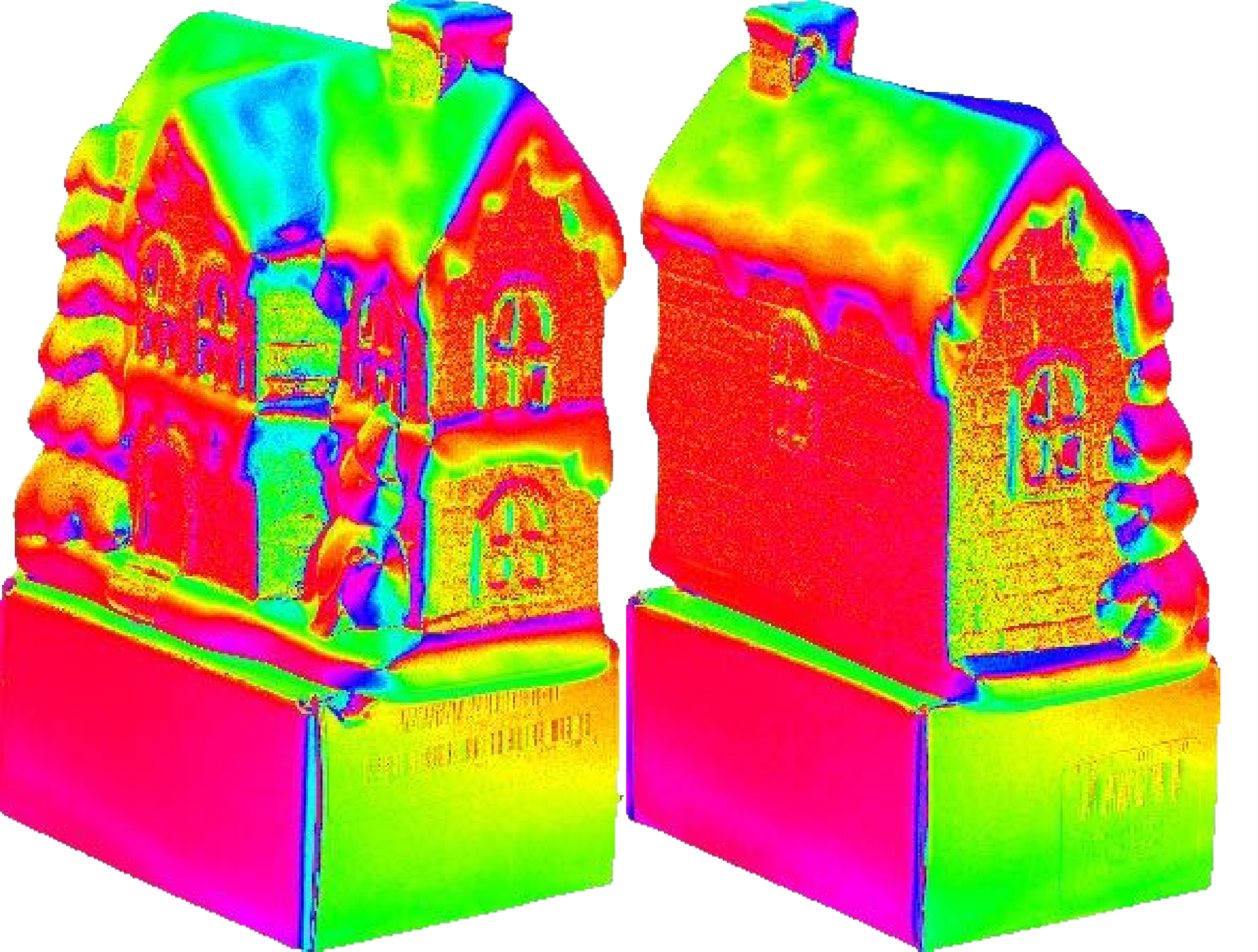} &
		\includegraphics[width=0.16\linewidth]{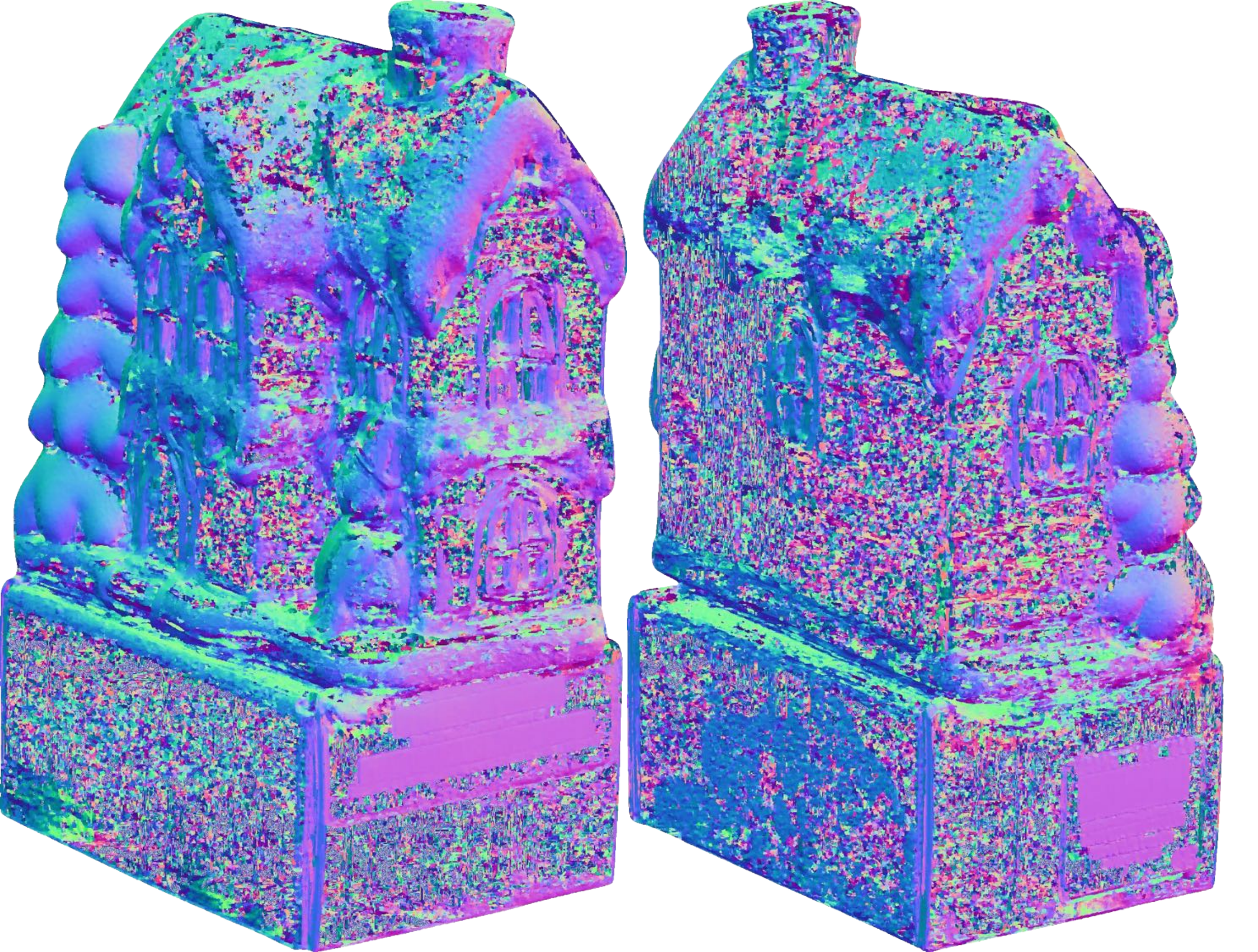} &
		\includegraphics[width=0.16\linewidth]{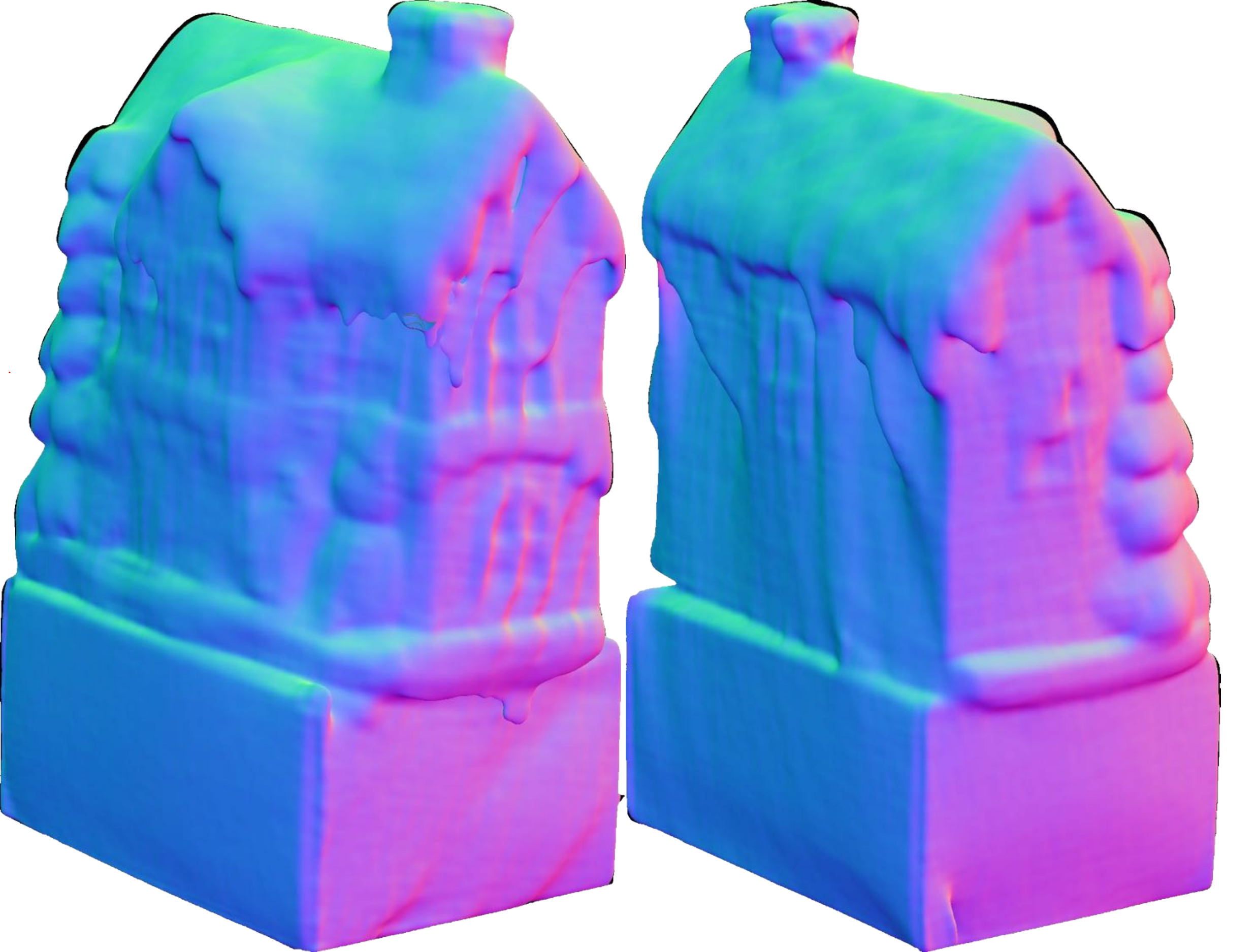} &
		\includegraphics[width=0.16\linewidth]{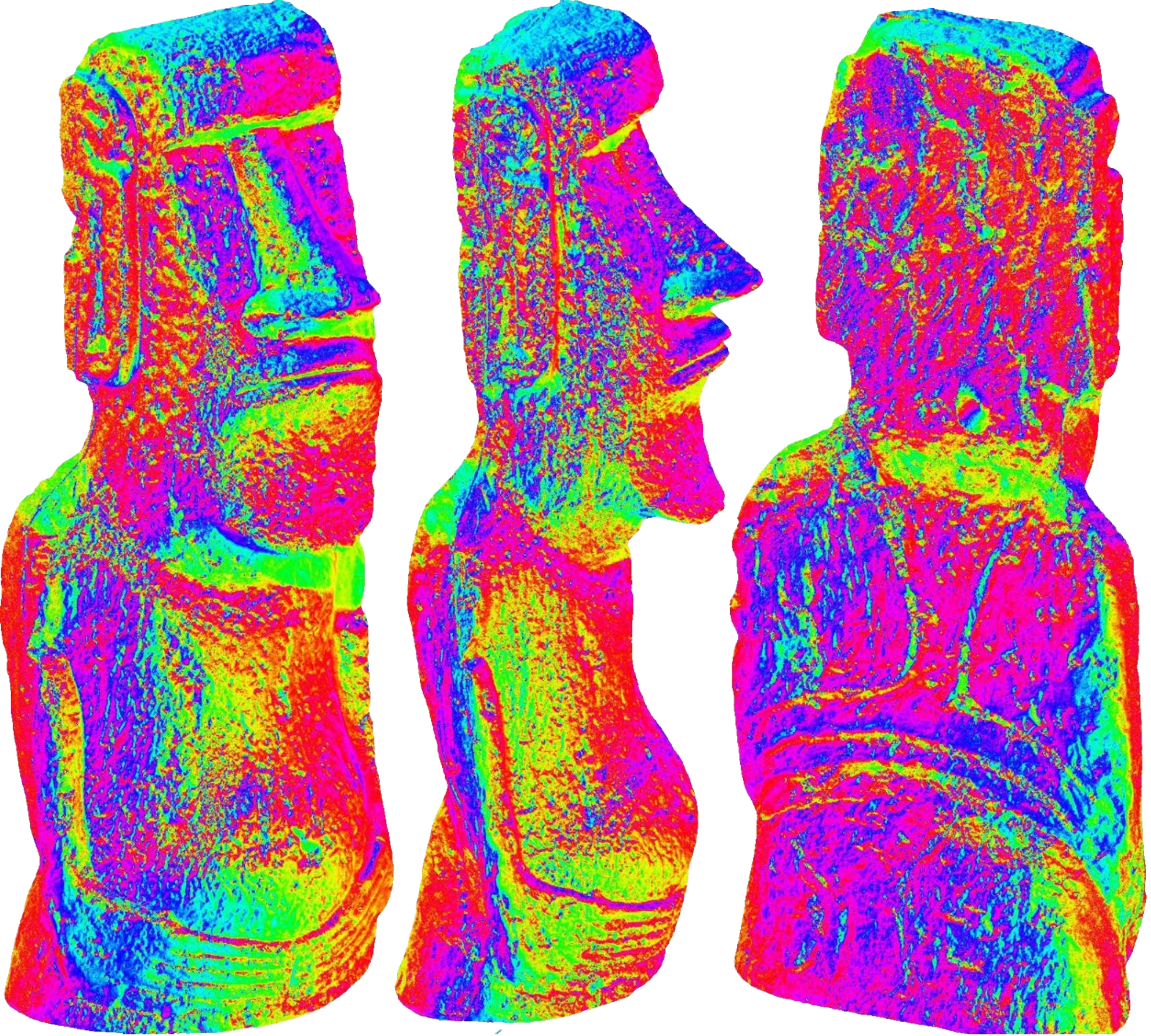} &
		\includegraphics[width=0.16\linewidth]{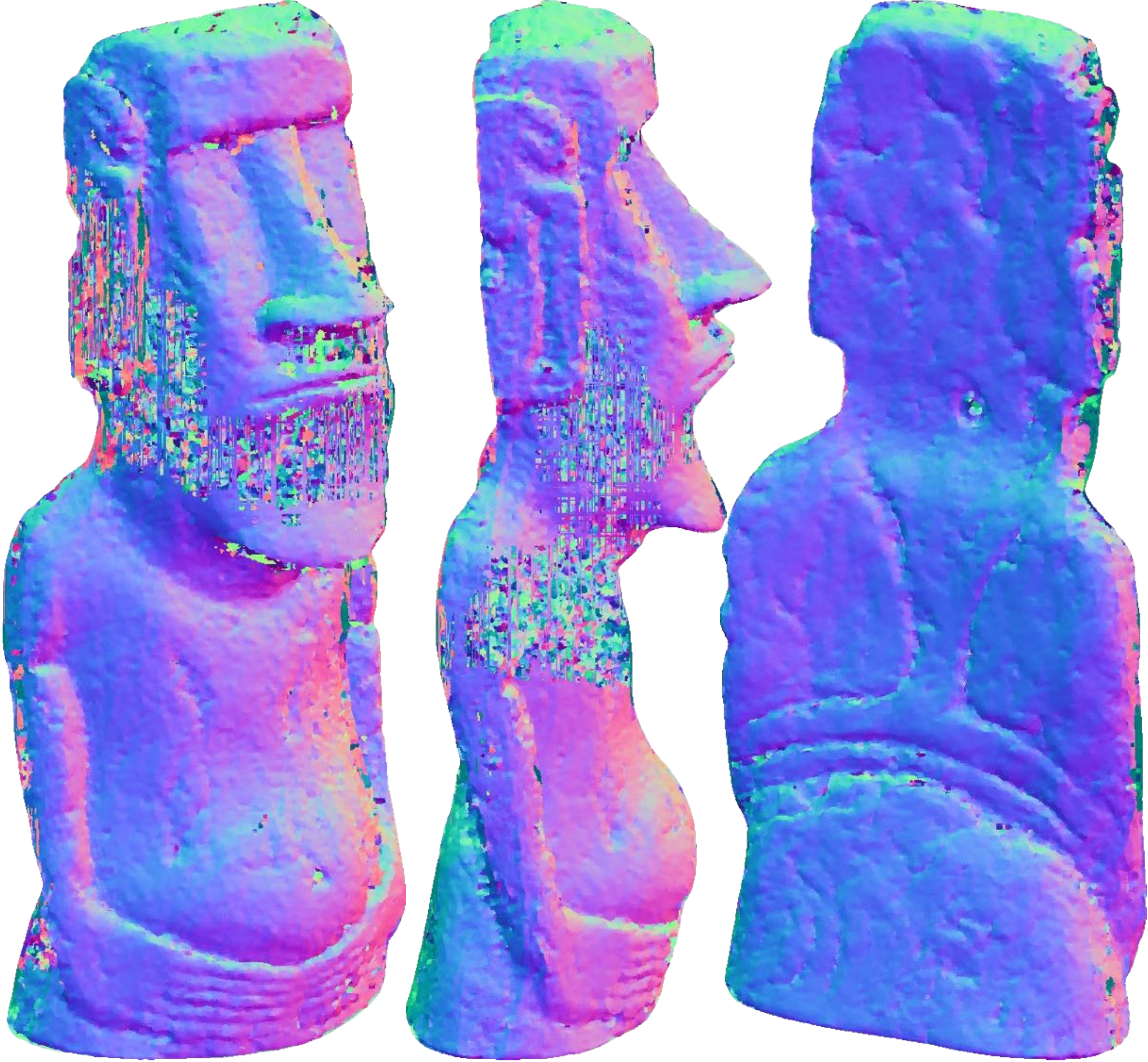} &
		\includegraphics[width=0.16\linewidth]{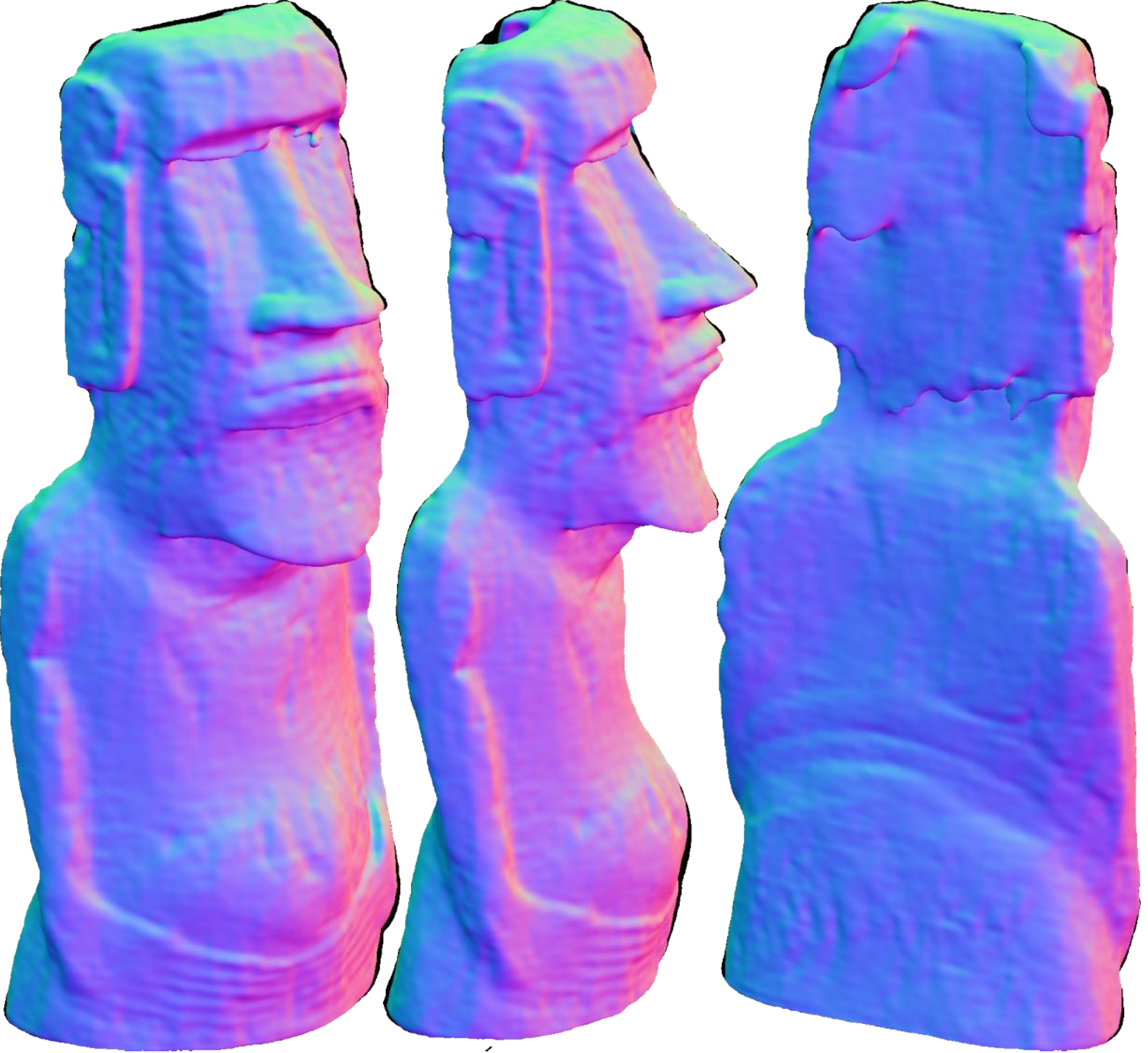} 
	\end{tabular}
\vspace{-0.6em}
	\caption{Visual comparison between \mvas and MVS on surface and normal reconstruction.}
	\label{fig.mvs_comparison}
\end{figure*}

%% file: sections/figures_tables/pandora_horizontal.tex
\begin{figure*}
	\scriptsize
	\centering
	\newcommand{\figwidth}{0.12}
	\begin{tabular}{@{}c@{}c@{}c@{}c c@{}c@{}c@{}c@{}}
		\begin{tabular}{c}
			Color images \& \\
		Polar-azimuth~\cite{dave2022pandora}
		\end{tabular} & 
		Colmap~\cite{schoenberger2016mvs} & 
		PANDORA~\cite{dave2022pandora} &
		\mvas (ours) &
		\begin{tabular}{c}
			Color images \& \\
			Polar-azimuth~\cite{dave2022pandora}
		\end{tabular} & 
		Colmap~\cite{schoenberger2016mvs} & 
		PANDORA~\cite{dave2022pandora} &
		\mvas (ours)
		\\
		\begin{minipage}{0.11 \linewidth}  \includegraphics[width= \linewidth]{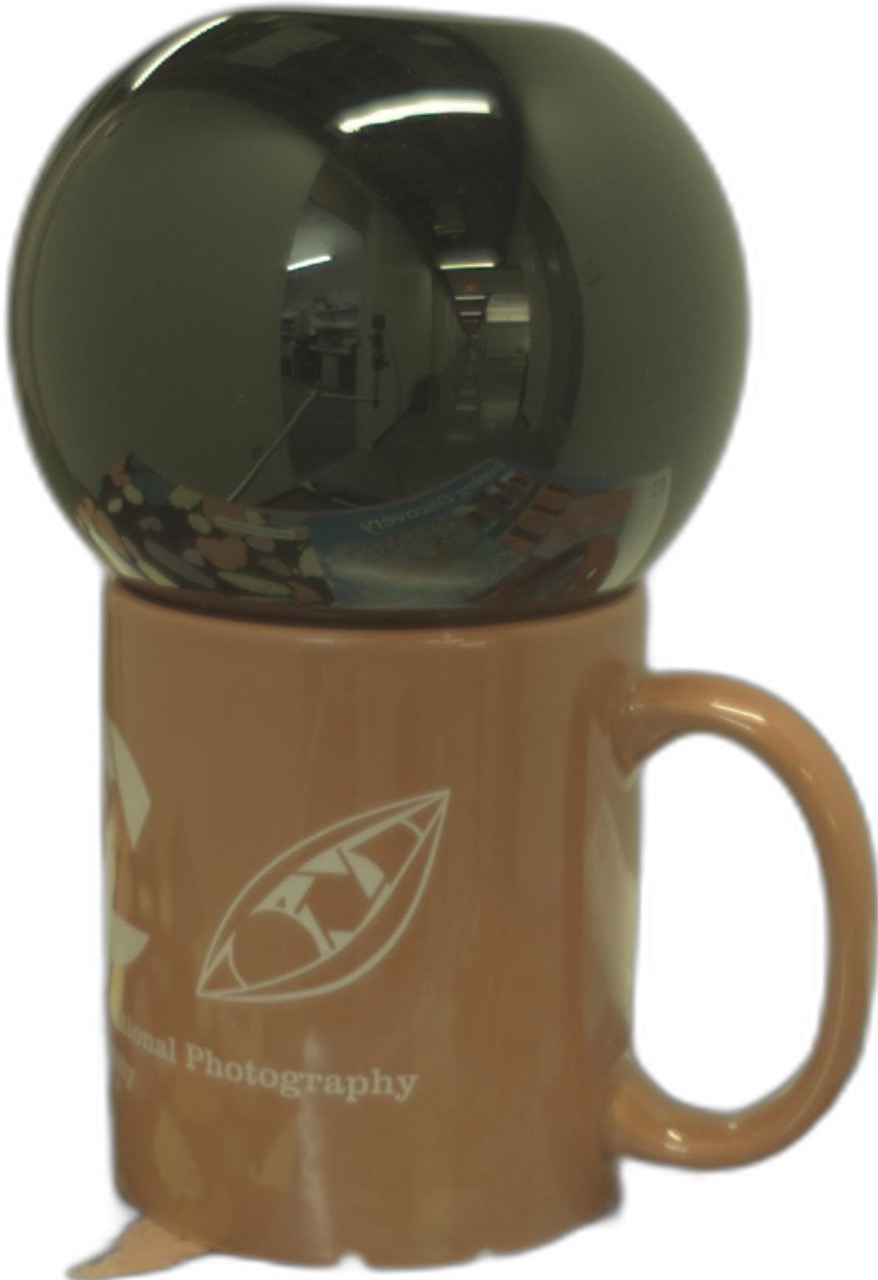} \end{minipage}& 
		\begin{minipage}{\figwidth \linewidth}   \includegraphics[width=\linewidth]{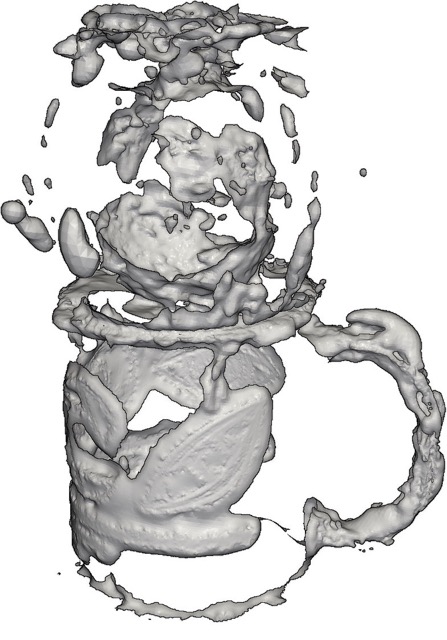}  \end{minipage}& 
		\begin{minipage}{0.11 \linewidth}   \includegraphics[width=\linewidth]{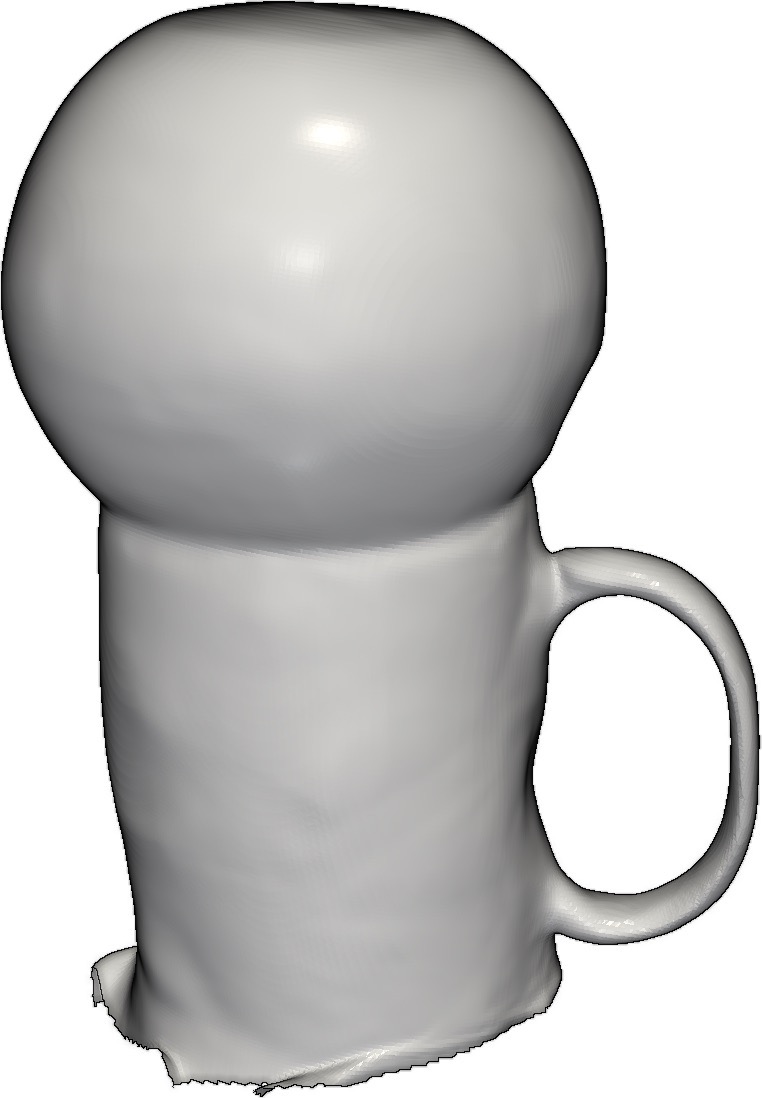} \end{minipage}&
		\begin{minipage}{\figwidth \linewidth}   \includegraphics[width= \linewidth]{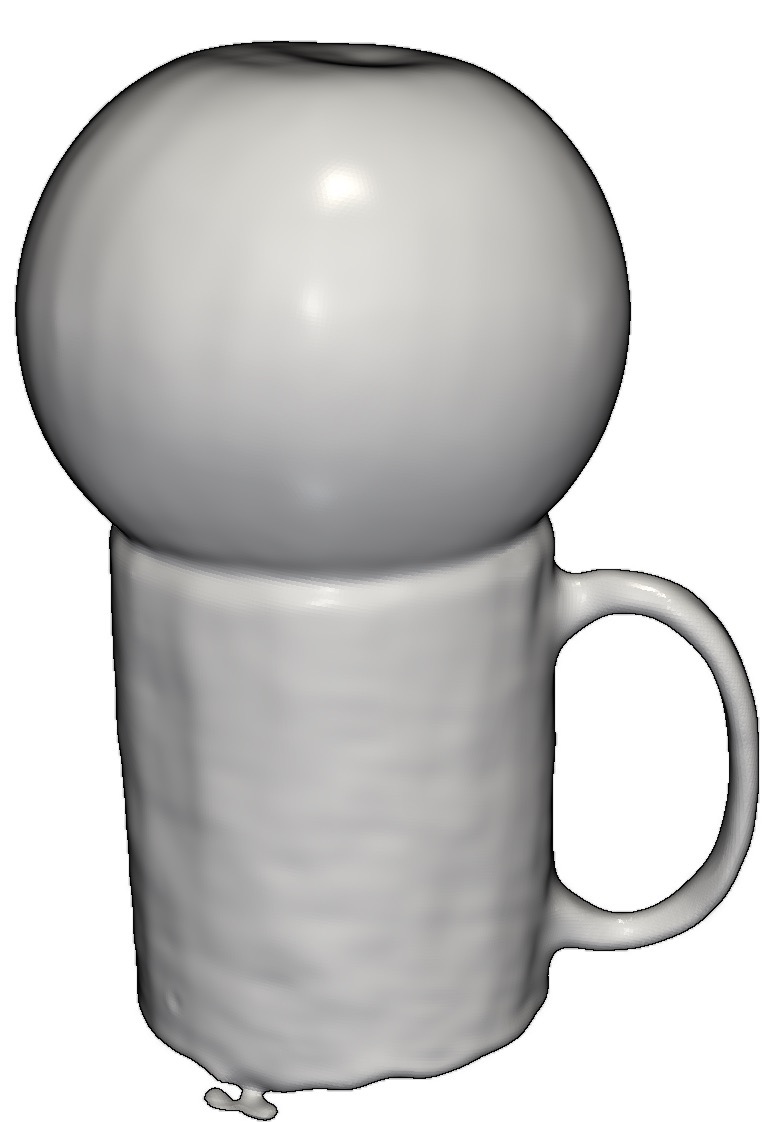}  \end{minipage} &
		\begin{minipage}{0.11 \linewidth} \includegraphics[width=\linewidth]{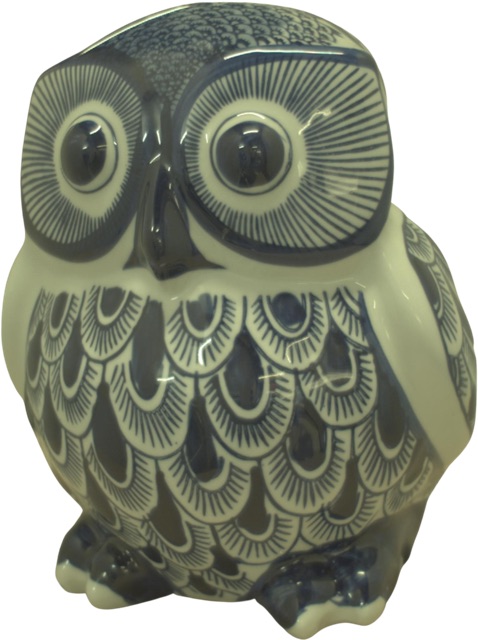}  \end{minipage}&
		\begin{minipage}{0.115 \linewidth}  \includegraphics[width=\linewidth]{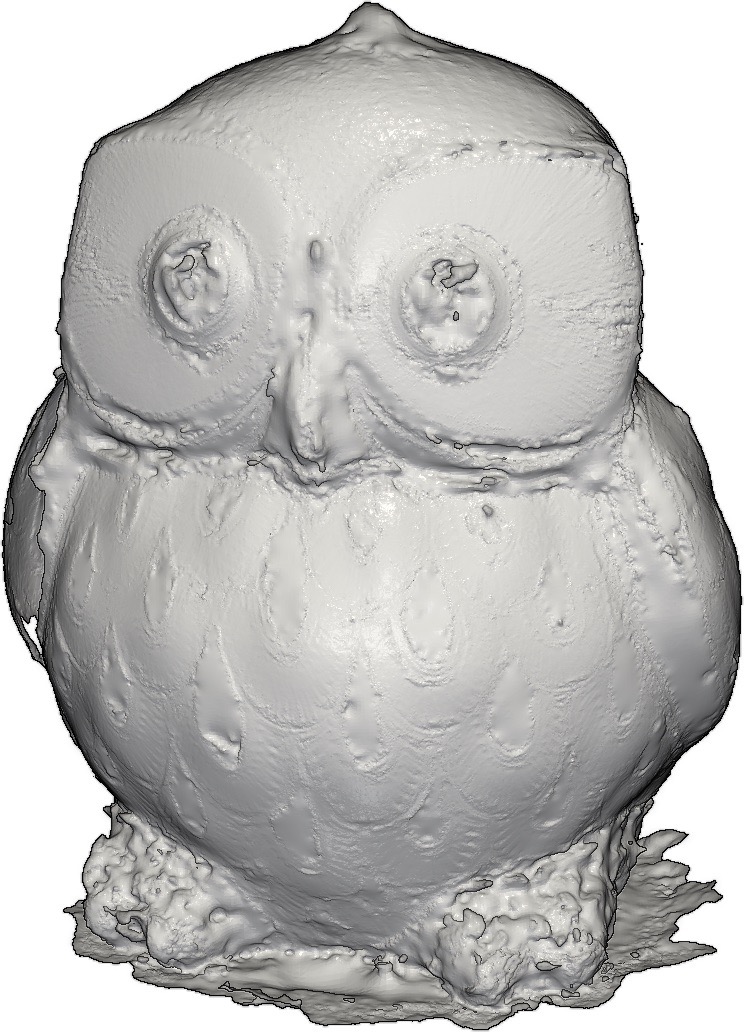}  \end{minipage}& 
		\begin{minipage}{\figwidth \linewidth} 	\includegraphics[width=\linewidth]{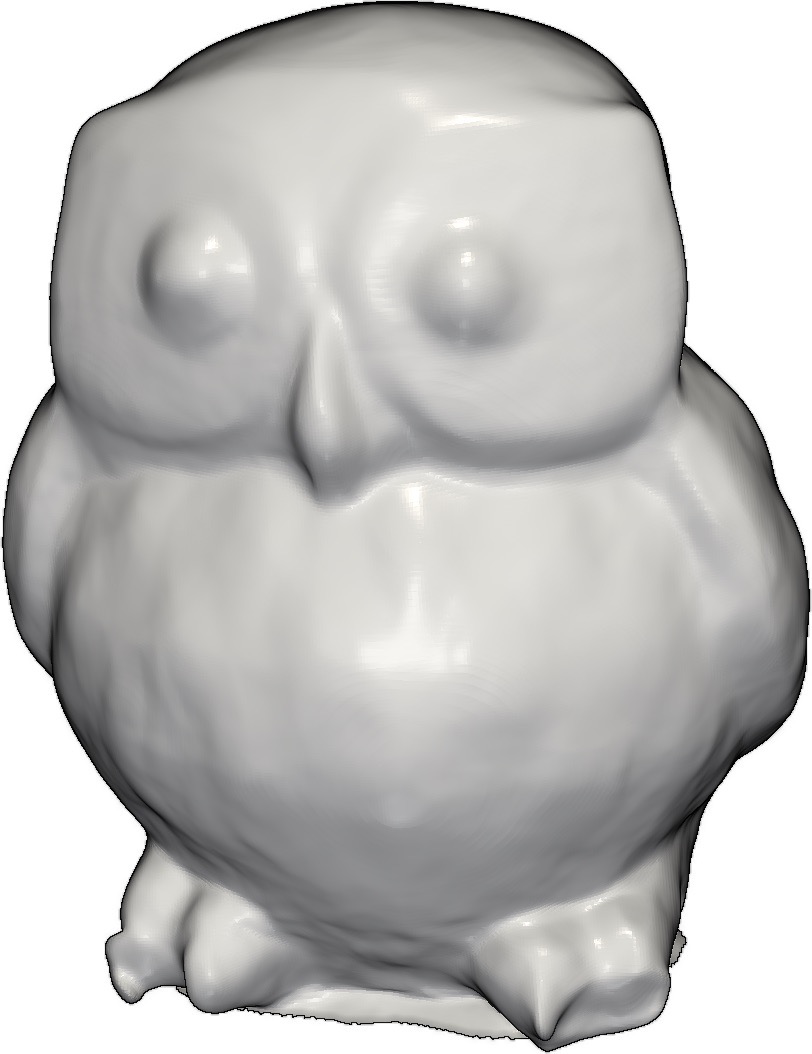}  \end{minipage}&
		\begin{minipage}{\figwidth \linewidth} 	\includegraphics[width=\linewidth]{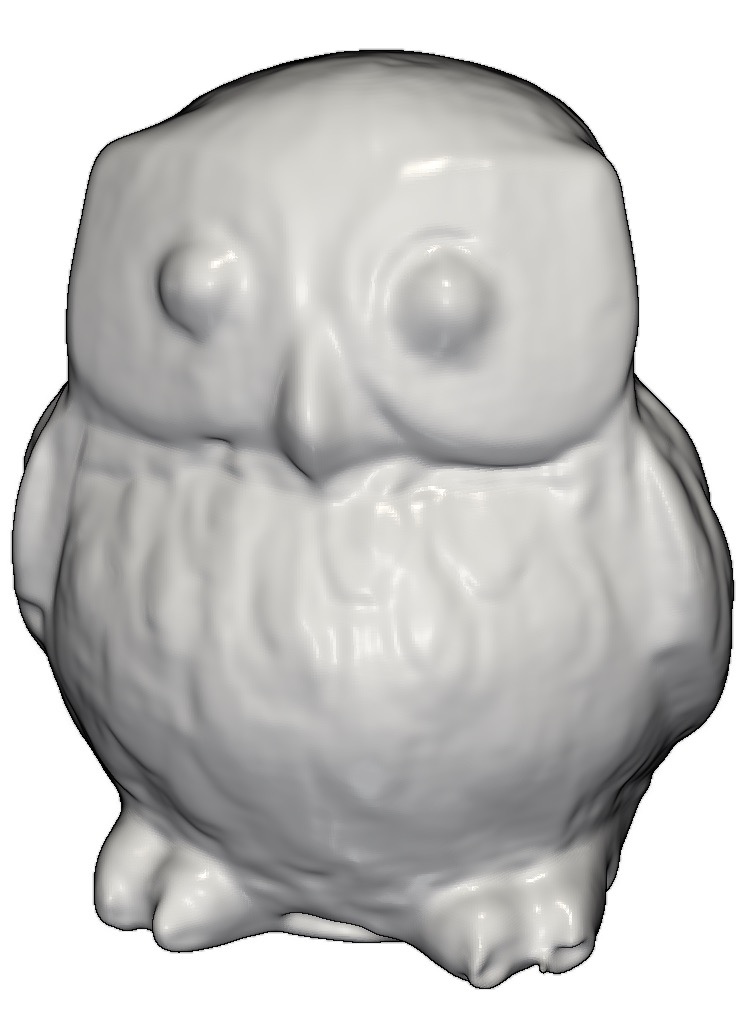}  \end{minipage}
		\\
		\begin{minipage}{\figwidth \linewidth}  \includegraphics[width=\linewidth]{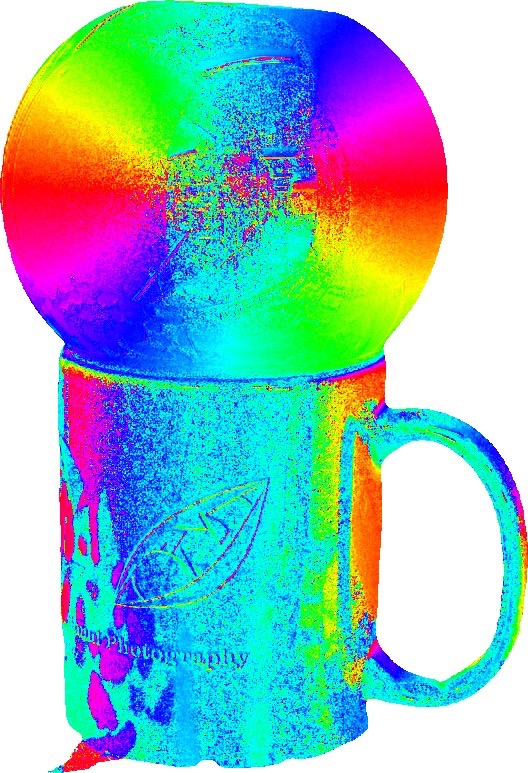}  \end{minipage}&
		\begin{minipage}{\figwidth \linewidth} 	\includegraphics[width=\linewidth]{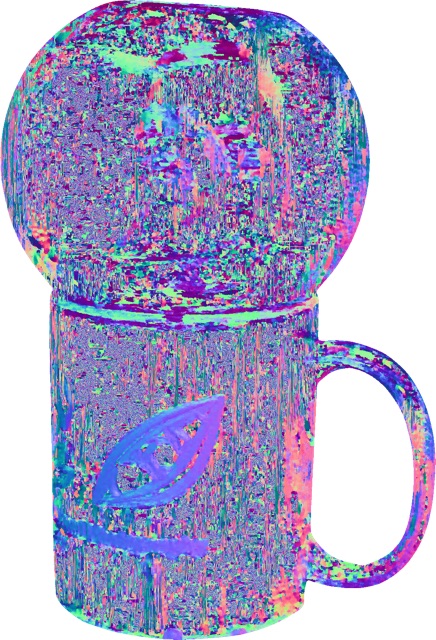}  \end{minipage}&
		\begin{minipage}{\figwidth \linewidth} 	\includegraphics[width=\linewidth]{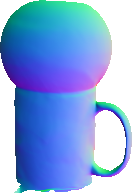}  \end{minipage}&
		\begin{minipage}{\figwidth \linewidth} 	\includegraphics[width=\linewidth]{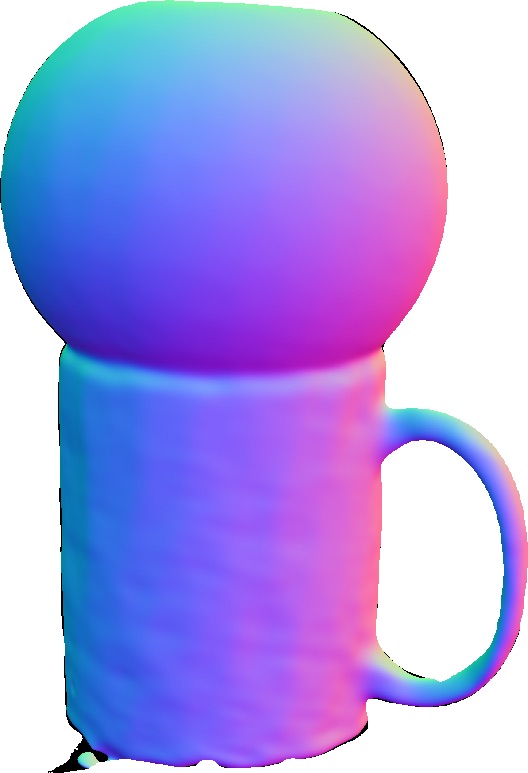}  \end{minipage}&
		\begin{minipage}{\figwidth \linewidth} \includegraphics[width=\linewidth]{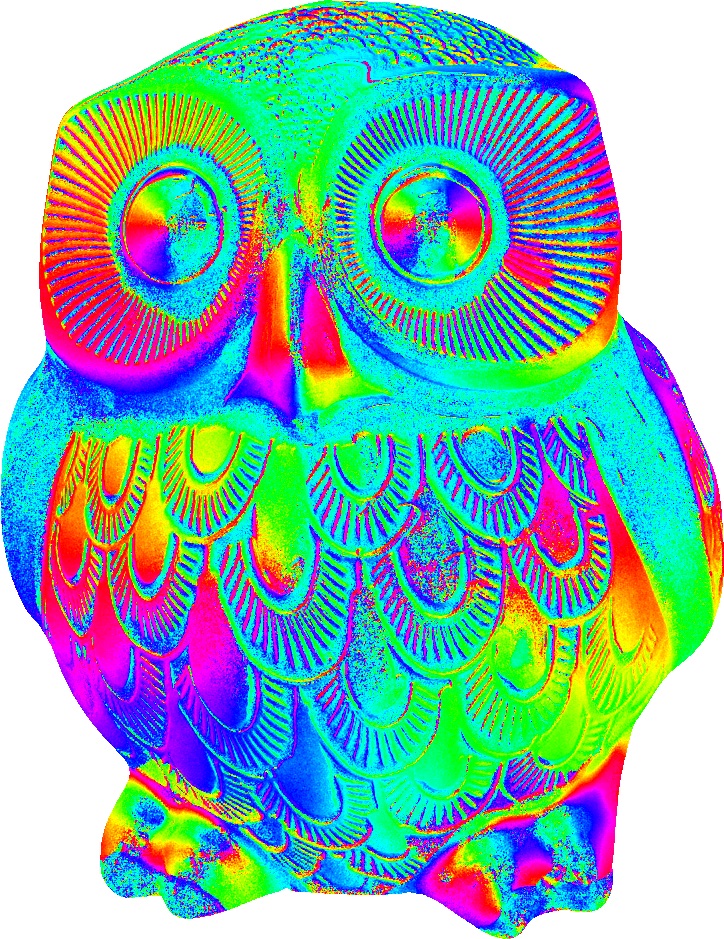}  \end{minipage}&
		\begin{minipage}{\figwidth \linewidth} \includegraphics[width=\linewidth]{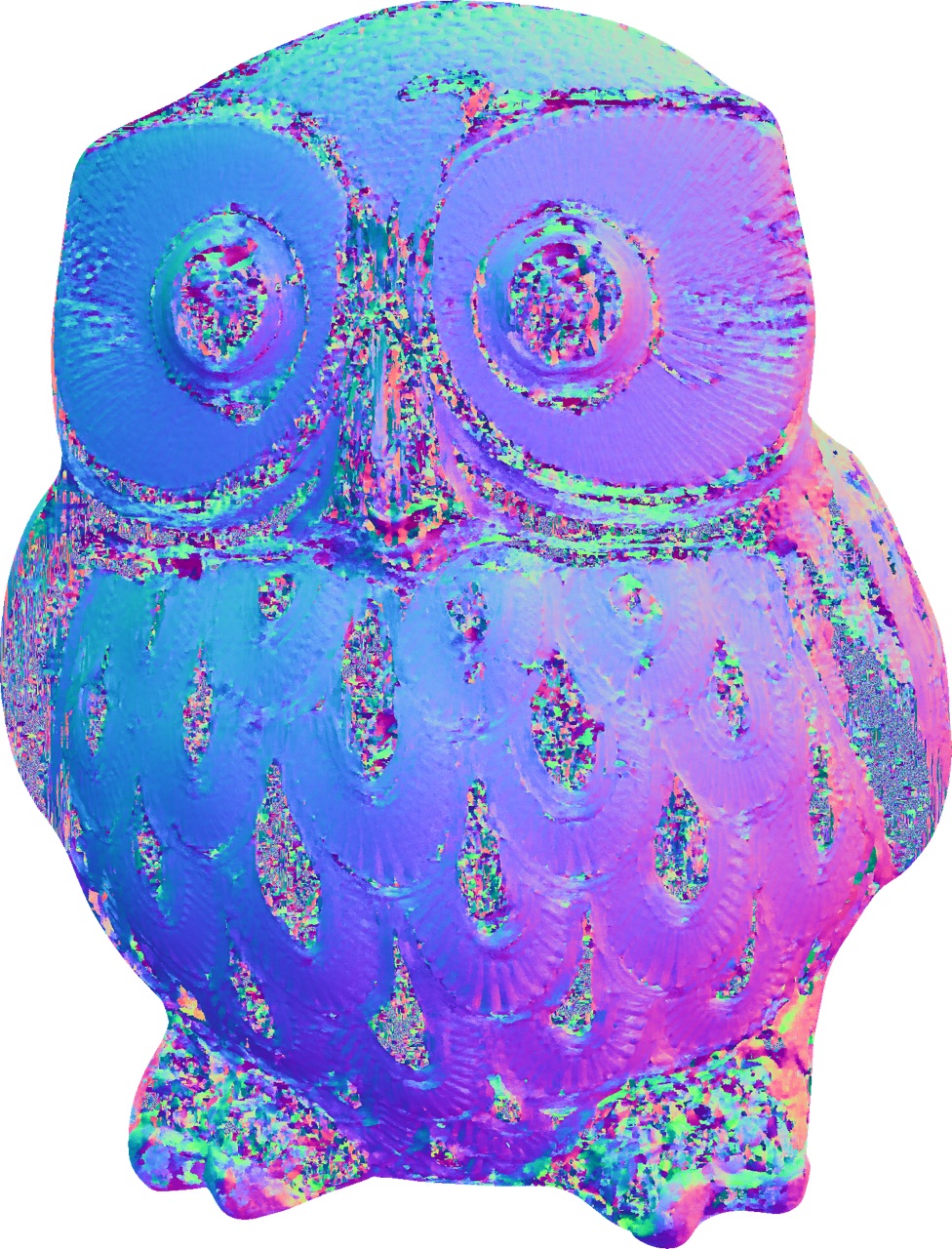}  \end{minipage}&
		\begin{minipage}{\figwidth \linewidth} \includegraphics[width=\linewidth]{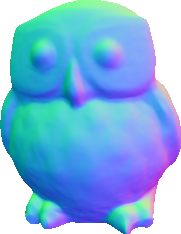}  \end{minipage}&
		\begin{minipage}{\figwidth \linewidth} \includegraphics[width=\linewidth]{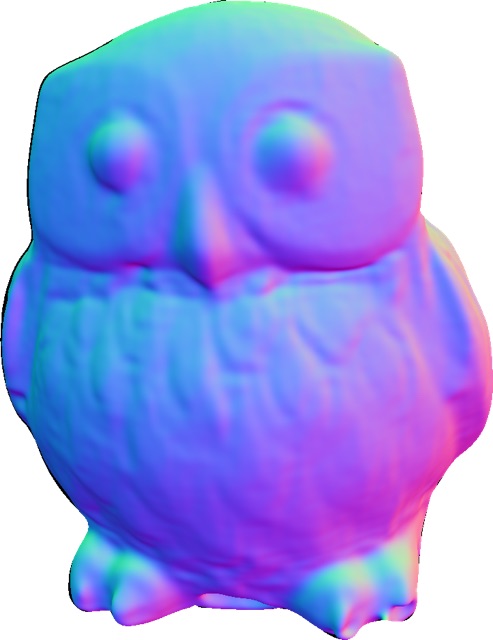}  \end{minipage}
		\\
	\end{tabular}
\vspace{-0.6em}
	\caption{Qualitative comparison of recovered surfaces and normals using a polarization camera~\cite{dave2022pandora}. $35$ views are used. 
	}
	\label{fig.pandora}
\end{figure*}

%% file: sections/06_conclusions.tex
\section{Discussions}
\label{sec.conclusion}

We present MVAS, an approach for reconstructing surfaces from multi-view azimuth maps. 
By establishing multi-view consistency in the tangent space and optimizing a neural SDF with the TSC loss, MVAS achieves comparable results to MVPS methods without zenith information. 
We verify MVAS's effectiveness with real-world azimuth maps obtained by symmetric-light photometric stereo and polarization measurements. 
Our results suggest that MVAS can enable high-fidelity reconstruction of shapes that have been challenging for traditional MVS methods.

Today, azimuth maps are still more expensive to obtain than ordinary color images, which may limit the application of \mvas. However, the situation will be changed when commercial polarimetric cameras are more accessible.

%% file: supp/supp.tex
\begin{appendices}
\label{appendices}
	
\setcounter{section}{0}
\setcounter{figure}{10}
\setcounter{equation}{20}
\setcounter{table}{2}
\renewcommand{\thesection}{\Alph{section}}

We provide more details and analysis of the proposed method, as listed below.
\startcontents

{
	\hypersetup{linkcolor=black}
	\printcontents{}{1}{}
}\
\section{Analysis of \tsc loss}
This section provides more details about our modification for \tsc loss to account for the \halfpi ambiguity in polarimetric azimuth observations, discusses the necessity of considering multi-view consistency, and provides more details and an efficiency analysis of our visibility determination strategy.
	
\subsection{Accounting for \halfpi ambiguity in \tsc loss}
We modify our \tsc loss to account for \halfpi ambiguity in polarimetric observations.
Given an observed polarimetric phase angle \phaseangle , the surface azimuth angle \azimuthangle is either \phaseangle \halfpi or $\phaseangle (=  \phaseangle+ \pi)$
depending on whether the surface point is polarimetric specular or diffuse reflection dominated~\cite{Cui_2017_CVPR,miyazaki2003polarization}.
Unfortunately, labeling the specular or diffuse domination is non-trivial~\cite{Cui_2017_CVPR, zhao2020polarimetric, dave2022pandora, zhu2019depth}.
In our approach, although \tsc is invariant to $\pi$ ambiguity, the \halfpi ambiguity still requires specific handling for polarimetric observations.
	
	\input{sections/figures_tables/supp_tsc_halfpi_comparison}
	
	Our idea is to allow both possibilities in the TSC loss.
	The \halfpi ambiguity introduces one more candidate tangent vector, and the surface normal should be perpendicular to either of the vectors deduced from $\pi$ or $\halfpi$ phase angles.
	By main paper's Eq.~(6), the projected tangent vector $\tangent^{\prime}$ from the \halfpi phase angle is
	\begin{equation}\label{eq.halfpi_tangent}
		\begin{aligned}
			\tangent^{\prime}(\azimuthangle) =\tangent\left(\phaseangle + \frac{\pi}{2}\right) &= \V{r}_1 \sin\left(\phaseangle+\frac{\pi}{2}\right) - \V{r}_2\cos\left(\phaseangle+\frac{\pi}{2}\right)  \\
			& = -\V{r}_1 \cos\left(\phaseangle\right) - \V{r}_2 \sin
		\left(\phaseangle\right).
		\end{aligned}
	\end{equation}
	Because $\tangent^{\prime}$ is also parallel to the image plane, $\tangent^{\prime}$  can be obtained by rotating \tangent by \halfpi in the image plane.
	At this point, however, we cannot fully determine which vector, \tangent or $\tangent^{\prime}$, is the actual tangent vector.
	We only know that the surface normal is perpendicular to either of the vectors:
	\begin{align}\label{eq.halfpi_tangent_normal}
		\normal \perp \tangent \quad \text{or} \quad \normal \perp \tangent^{\prime}.
	\end{align}
	Putting together the notations in the main paper's Eqs.~(12) and (17), we can rewrite our TSC loss as
	\begin{align}\label{eq.tsc_loss_extended}
		\loss_{\textrm{TSC}} = 
		\frac{1}{\batchsize} \sum_{\point \in \V{X}} \frac{\sum_{i=1}^{\cameraNum} \visibility \left(\normal^\top\tangent_i\right)^2}{\sum_{i=1}^{\cameraNum}\visibility}.
	\end{align}
	Based on \cref{eq.halfpi_tangent_normal}, we modify \cref{eq.tsc_loss_extended} as 
	\begin{align}
		\loss_{\textrm{TSC}}^{\prime} = 
		\frac{1}{\batchsize}\sum_{\point \in \V{X}} \frac{\sum_{i=1}^{\cameraNum} \visibility \left(\normal^\top\tangent_i\right)^2\left(\normal^\top\tangent^{\prime}_{i}\right)^2}{\sum_{i=1}^{\cameraNum}\visibility}.
	\end{align}
The modified \tsc loss allows the surface normal to be perpendicular to either of the two candidate tangent vectors.
	
\Cref{fig.tsc_halfpi} shows that this strategy yields better reconstruction quality, which gives us the results presented in the main paper's Fig.~10.
If we do not deal with \halfpi ambiguity, the recovered shapes appear twisted due to wrong tangent vectors (\ie, rotated by \halfpi from actual tangent vectors in the image space).
	
\input{sections/figures_tables/ablation}
\subsection{Ablation study on multi-view consistency}
Accumulating \projectedTangentVectors from all visible views  to compute the \tsc loss is necessary for accurate shape recovery.
Without considering multi-view consistency, we can simplify our original \tsc loss from \cref{eq.tsc_loss_extended} to
\begin{align}
	\loss_{\textrm{TSC}}^{''} = 
	\frac{1}{\batchsize} \sum_{\point \in \V{X}} (\normal(\point)^\top\tangent(\azimuthangle(\Pi(\point))))^2,
\label{eq.tsc_simplified}
\end{align}
where the \projectedTangentVector \tangent is computed from the input pixel location, and visibility or tangent vectors in other views need no longer be considered.

This simplified loss \cref{eq.tsc_simplified}, however, can lead to convex-concave ambiguity in the recovered surfaces, as shown in \cref{fig.ablation_on_reprojection}.
Without multi-view consistency, the tangent vector from one view can only constrain the surface normal loosely on a plane and cannot constrain the surface positions correctly.
Therefore, locally concave or convex surfaces with the same tangent vectors can both minimize the simplified loss, thus resulting in the ambiguity.
	
\subsection{More details on visibility determination}
We determine the visibility of a surface point in a view by marching the point toward the corresponding camera, \ie, performing sphere tracing~\cite{sphere1996hart} in the reverse direction.

\begin{figure}[t]
	\centering
	\includegraphics[width=\linewidth]{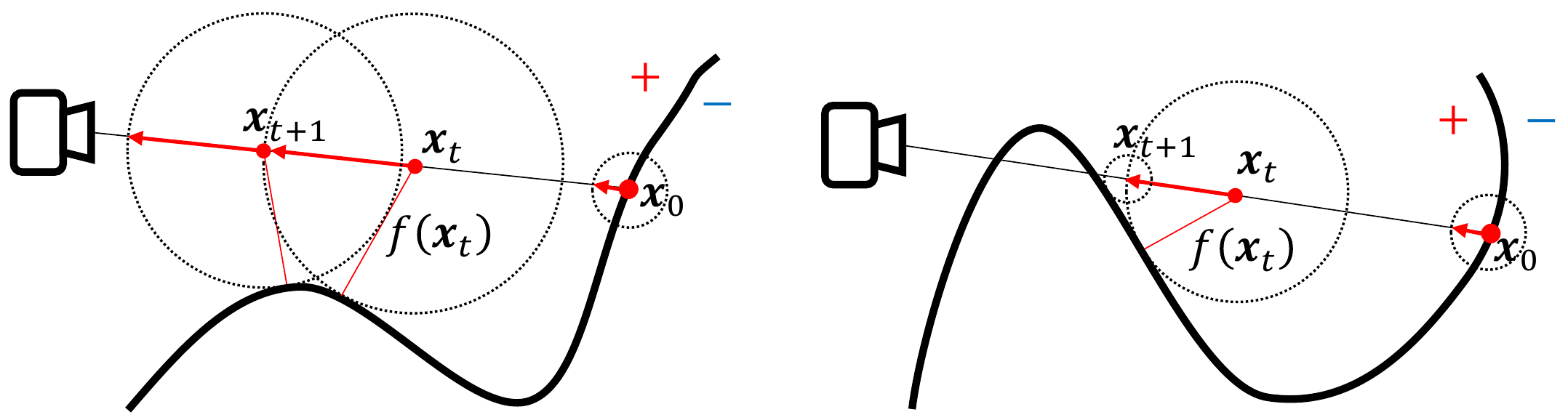}
	\caption{Visibility determination via reverse sphere tracing. We march a surface point $\point_0$ towards the camera center. At each step, the marching distance is the signed distance $f(\point_t)$ from the current point $\point_t$ to the surface, which requires one MLP evaluation. \textbf{(Left)} The marching diverges quickly towards the camera if $\point_0$ is visible. \textbf{(Right)} The marching converges to another surface point as ordinary sphere tracing~\cite{sphere1996hart} if $\point_0$ is occluded.}
	\label{fig.reverse_sphere_tracing}
\end{figure}

\begin{figure}[t]
	\centering
	\begin{tabular}{cc}
		\includegraphics[height=0.4\linewidth]{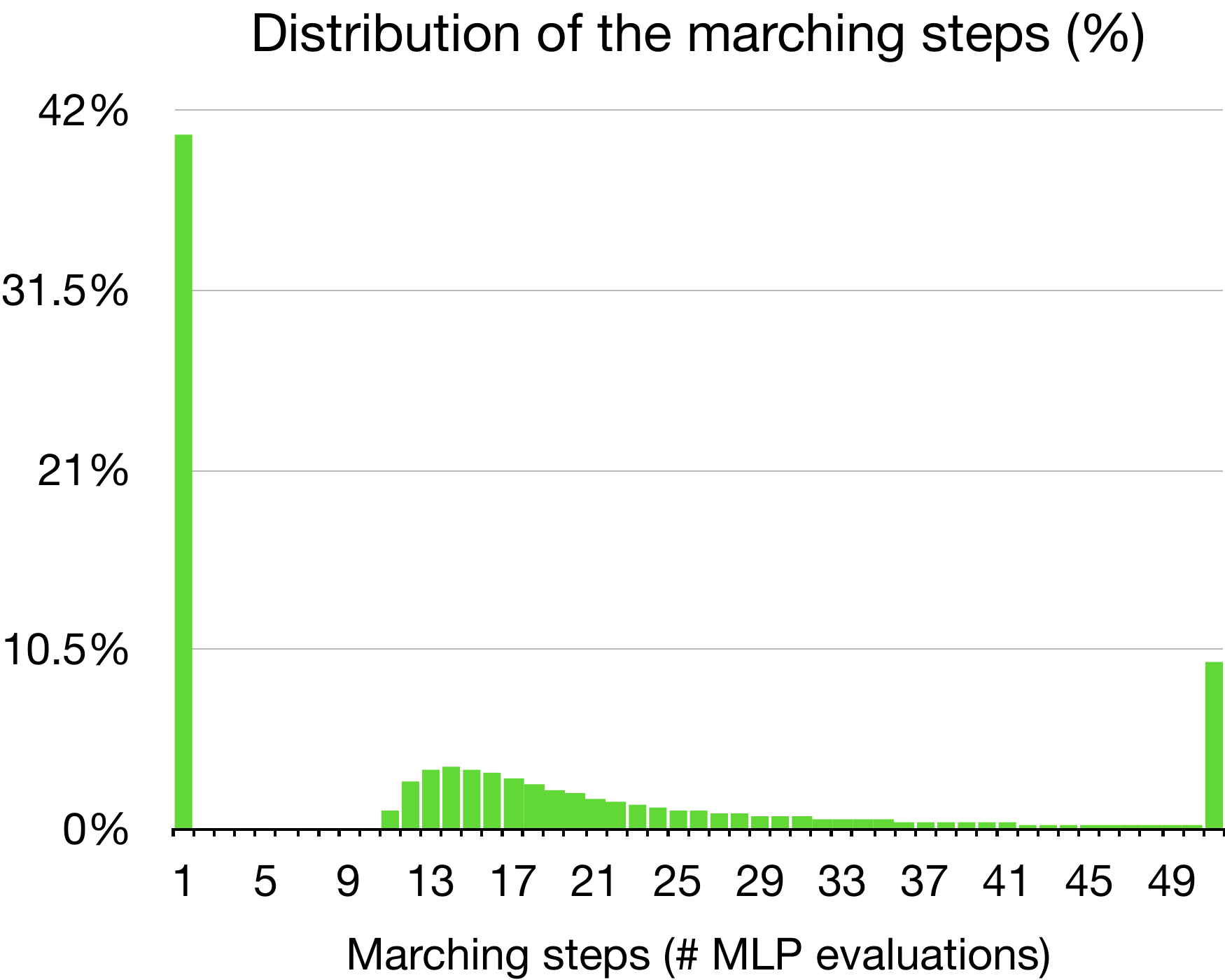} &
		\includegraphics[height=0.4\linewidth]{images/shape_vis/buddha/draw_time_2022_10_16_14_31_28/2022_10_16_01_23_08}
	\end{tabular}
	\caption{
		The distribution of marching steps required to determine the visibility of surface points of the \diligentmv object ``Buddha''~\cite{li2020multi}.
		On average, $16$ MLP evaluations are required per surface point per view over the training.
	}
	\label{fig.vis_distribution}
\end{figure}

We consider four conditions when marching the surface point. 
Initially, we push the surface point $\point_0$ by a tiny distance ($1\times 10^{-3}$ in our experiments) to the camera.
(1) The surface point is invisible if the signed distance becomes negative, as the marching direction is towards inside the surface.
As long as the marching point is outside the surface, we move the point $\point_t$ at step $t$ by a distance $f(\point_t)$ towards the camera.
The surface point is (2) visible if the marching point goes beyond the camera center (\cref{fig.reverse_sphere_tracing} left) or (3) invisible if the marching point hits another surface point (\cref{fig.reverse_sphere_tracing} right). 
(4) We treat the surface point as invisible if the marching is not terminated within certain steps.

This strategy is advantageous in both efficiency and accuracy compared to other visibility determination strategies used in neural rendering methods.
First, it avoids densely evaluating an MLP on the point-to-camera rays~\cite{nerfactor2021,facerelighting2022}.
The marching quickly terminates and only requires a few MLP evaluations, \eg, $16$ MLP evaluations on average ( \cref{fig.vis_distribution}).
Second, it does not rely on the visibility predicted by an additional trainable MLP~\cite{yang2022psnerf}.

\section{Evaluation on \diligentmv}
This section provides more details of our evaluation metrics, additional visual comparisons on \diligentmv benchmark~\cite{li2020multi}, and investigates the effect of number of input viewpoints.

\subsection{More details on evaluation metrics}
The definition of our evaluation metrics follow~\cite{kaya2022uncertainty,knapitsch2017tanks}.
We present their definitions here for completeness.
\paragraph{Chamfer distance}
Chamfer distance measures the point-set-to-point-set distance by accumulating the point-to-point-set distances.
Given two point sets \pointsetOne, and \pointsetTwo, the distance from a point  to another point set is defined as
\begin{equation}
	\begin{aligned}
	&d_{\pointOne \rightarrow \pointsetTwo} = \min_{\pointTwo \in \pointsetTwo} \norm{\pointOne - \pointTwo}_2 \quad \text{and} \\ 
	&d_{\pointTwo \rightarrow \pointsetOne} = \min_{\pointOne \in \pointsetOne} \norm{\pointOne - \pointTwo}_2.
	\end{aligned}
\end{equation}
The Chamfer distance \chamferDist is then 
\begin{align}
	\chamferDist = \frac{1}{2\abs{\pointsetOne}} \sum_{\pointOne \in \pointsetOne} d_{\pointOne \rightarrow \pointsetTwo} + \frac{1}{2\abs{\pointsetTwo}} \sum_{\pointTwo \in \pointsetTwo} d_{\pointTwo \rightarrow \pointsetOne}.
\end{align}

\paragraph{F-score}
F-score considers both the precision and recall of the recovered surfaces to the GT surfaces.
The precision and recall are defined based on the point-to-point-set distances as
\begin{equation}
	\begin{aligned}
	\precision &=  \frac{1}{\abs{\pointsetOne}} \sum_{\pointOne \in \pointsetOne}[d_{\pointOne \rightarrow \pointsetTwo} < \fscoreThreshold]  \quad \text{and}\\
	\recall &=  \frac{1}{\abs{\pointsetTwo}} \sum_{\pointTwo \in \pointsetTwo}[d_{\pointTwo \rightarrow \pointsetOne} < \fscoreThreshold].
	\end{aligned}
\end{equation}
Here, $[\cdot]$ is the Iverson bracket, and \fscoreThreshold is the distance threshold for a point to be considered close enough to a point set.
The F-score then takes the geometric average of precision and recall:
\begin{align}
	\fscore = \frac{2\precision \recall}{\precision + \recall}.
\end{align}
We set $\fscoreThreshold=\SI{0.5}{mm}$ in our evaluations.

As mentioned in the main paper, our evaluation takes the first ray-surface intersection points from all views as the input point sets to the Chamfer distance and F-score.
This puts more focus on evaluating visible surface regions in input images and avoids a heuristic crop of the surface~\cite{kaya2022uncertainty}.

\input{sections/figures_tables/diligent_inner}

\input{sections/figures_tables/diligent_shape_supp}

\input{sections/figures_tables/diligent_normal_supp}

Our evaluation metrics do not consider the \emph{cleanness} of inner space (\ie, correctness of inner topology) of the recovered surfaces.
To assess how accurate the inner space of the surfaces is, we visualize the inner space of the mesh in \cref{fig.diligent_inner}.
The visualization shows that our method does not produce unwilling structures inside recovered meshes.

\subsection{Additional visual comparisons}

\Cref{fig.vis_comp_mvps_supp,fig.comp_mvps_normal_supp} show the visual comparisons on \diligentmv objects~\cite{li2020multi} in addition to the ones presented in the main paper's Figs.~(7) and (8). 
Our method consistently recovers accurate and detailed shapes and normals.

\input{sections/figures_tables/diligent_normal_unseen}

\Cref{fig.comp_mvps_normal_supp_unseen} shows the comparison of surface normals to \psnerf~\cite{yang2022psnerf} from the $5$ unseen viewpoints during the training.
\psnerf~\cite{yang2022psnerf} use the $15$-view SDPS normal maps~\cite{chen2019SDPS_Net} to initialize shapes, therefore sharing the same access to underlying azimuth information as ours.
The comparison verifies that accurate shape and normal recovery can be realized using only azimuth maps without developing the rendering process for the multi-view case. 

\subsection{The effect of number of viewpoints}

\input{sections/figures_tables/diligent_num_views_quan}
\input{sections/figures_tables/diligent_num_views}
\mvas is robust to sparse view input.
As shown in \cref{tab.num_view} and \cref{fig.diligent_num_views}, we evaluate the shape and normal recovery accuracy by gradually reducing the number of input views.
\Cref{fig.diligent_num_views} shows that using as few as $5$-view azimuth maps can still achieve detailed reconstruction, while large errors are observed mainly at heavily occluded regions.

\section{Implementation details}
\label{sec.implementation_details}
This section describes the architecture of our neural SDF, the training details, and the camera normalization process.
\begin{figure}[t]
	\centering
	\includegraphics[width=\linewidth]{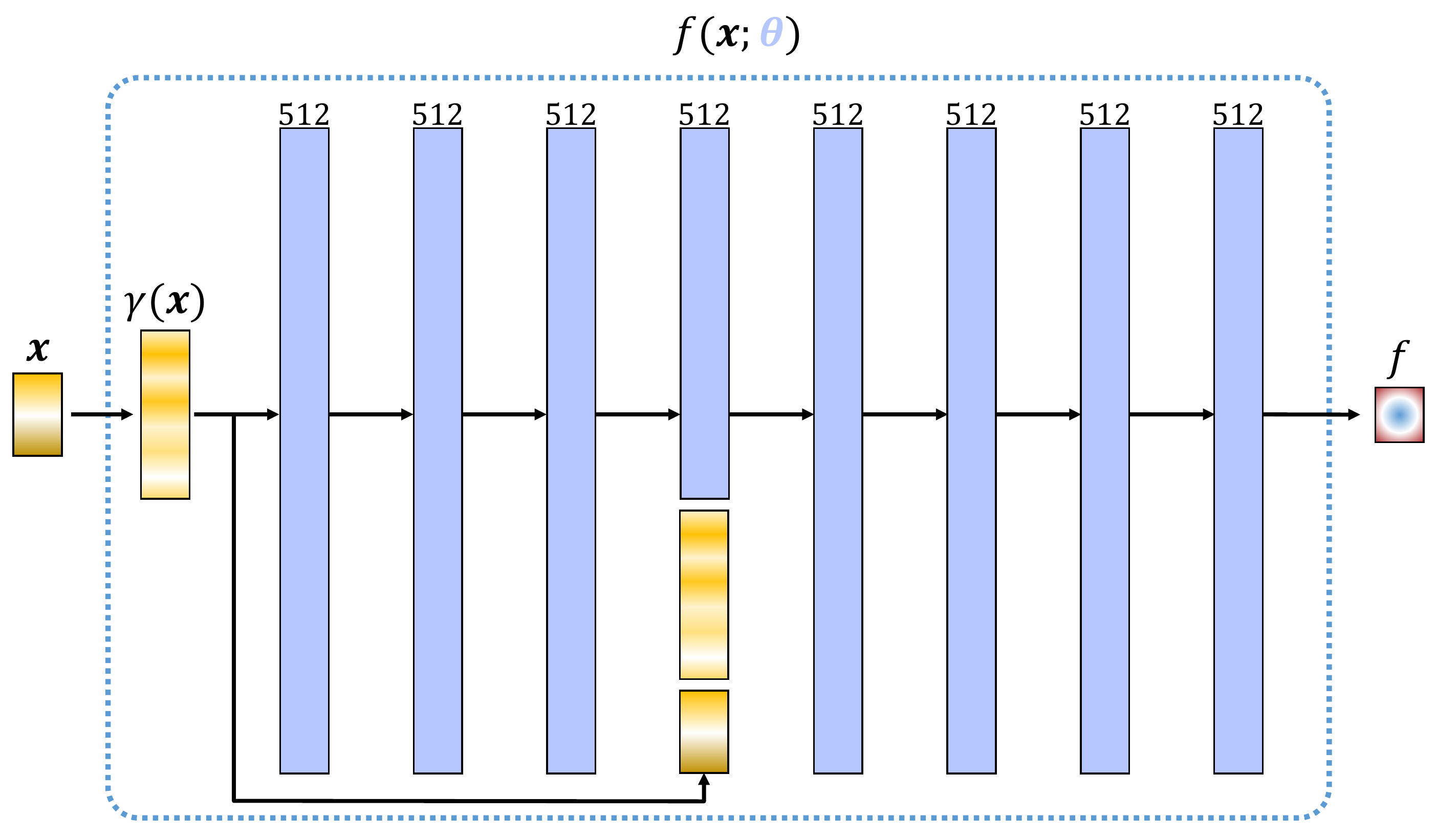}
	\caption{Our network consists of a positional encoding layer~$\gamma(\cdot)$ and an $8$-layer MLP with softplus activation functions. A skip connection is added to the $4$-th layer from the input. This is the only network we optimize.}
	\label{fig.mlp_architecture}
\end{figure}

\subsection{Neural network architecture}
Following IDR~\cite{idr2020multiview}, our neural SDF consists of a positional encoding layer~\cite{mildenhall2020nerf} followed by an $8$-layer MLP, as shown in \cref{fig.mlp_architecture}.
The positional encoding layer is defined as
\begin{equation}
	\begin{aligned}
		\gamma(\point) = [ \sin(2^0 \pi \point),& \cos(2^0\pi \point), ..., \\
		&\sin(2^{L-1} \pi \point), \cos(2^{L-1}\pi \point) ] .
	\end{aligned}
\end{equation}
We use $L=10$ in our experiments.
The input position \point and $\gamma(\point)$ are skip-connected to the $4$-th layer of the MLP.
For the activation functions in the MLP, we use the softplus function
\begin{align}
	\textrm{softplus}(x) = \frac{1}{\beta} \log\left(1+\exp(\beta x)\right)
\end{align}
with $\beta=100$.

The neural SDF shown in \cref{fig.mlp_architecture} is the only MLP we optimize.
Unlike recent works using additional rendering networks to model surface light field~\cite{idr2020multiview,volsdf2021yariv} or reflectance~\cite{yang2022psnerf} for computing re-rendering loss, multi-view azimuth maps directly regularize the geometry and eliminate the necessity to model a rendering process.

\input{sections/paragraph_implementation_details}

\subsection{Camera normalization}
Following \volsdf~\cite{volsdf2021yariv},  we normalize the world coordinates such that the object is bounded by a unit sphere.
As we cannot know the shape and its center position beforehand,  we approximate the object center location by the position that is closet to all camera principle axes.
This approximation assumes all cameras surrounding the target scene and is satisfied in our experiments.
We present the computation details here because we do not find such details in the \volsdf paper~\cite{volsdf2021yariv}.
The normalization is done by shifting and then scaling the camera center locations:
\begin{equation}
	\V{o}_i \leftarrow \frac{\V{o}_i - \V{x}_o}{s}.
\end{equation}
Here, $\V{o}_i$ is the $i$-th camera's center location in the world coordinates, $\V{x}_o$ and $s$ are the global offset and scale factor to be detailed in the following.

\paragraph{Camera centers' offset}
The offset applied to all camera center locations can be computed using a linear system.
Formally, let $\V{o}_i \in \R^3$ and $\V{z}_i \in \mathcal{S}^2 \subset \R^3$ be the \mbox{$i$-th} camera's center location and its principle axis direction in the world coordinates, respectively.
The principle axis can then be represented as $\point_i(t) = \V{o}_i + t\V{z}_i $ with $t \in \R_+$.
The shortest squared Euclidean distance from a point $\point \in \R^3$ to this principle axis is
\begin{equation}
	\begin{aligned}
	d^2\left(\point, \point_i(t)\right)	  & = \min_t \norm{\point - \point_i(t)}^{2}_2 \\
	& = (\point - \V{o}_i)^\top (\point - \V{o}_i) - \left((\point-\V{o}_i)^\top \V{z}_i\right)^2 \\
	& = \point^\top \V{Z}_i\point - 2\V{o}_i^\top \V{Z}_i\point + \V{o}_i^\top\V{Z}_i\V{o}_i,
	\end{aligned}
\end{equation}
where $\V{Z}_i = \V{I} - \V{z}_i\V{z}_i^\top$. To approximate the object center, we find the point that is the closest to all camera principle axes:
\begin{equation}
	\begin{aligned}
		\point_o &= \argmin_\point \sum_i d^2\left(\point, \point_i(t)\right) \\
		&= \point^\top \left(\sum\limits_{i=1}^\cameraNum\V{Z}_i\right)\point - 2\left(\sum\limits_{i=1}^\cameraNum \V{o}_i^\top \V{Z}_i\right)\point + \sum\limits_{i=1}^\cameraNum \V{o}_i^\top\V{Z}_i\V{o}_i.
	\end{aligned}
	\label{eq.camera_center}
\end{equation}
The global optimum $\point_o$ is attained by solving the following normal equation of~\cref{eq.camera_center}:
\begin{equation}
	\begin{aligned}
		\V{A}^\top\V{A}\point &= \V{A}^\top \V{b} \\ 
		\text{with} \quad \V{A}&=\sum\limits_{i=1}^\cameraNum\V{Z}_i, \quad\V{b}=\sum\limits_{i=1}^\cameraNum \V{Z}_i\V{o}_i
	\end{aligned}
\end{equation}

\paragraph{Camera centers' scale}
After centering the scene, we apply a global scale to all camera center locations to ensure a unit sphere bounds the scene.
We assume that all cameras surround the object.
Then we can compute the global scale factor as the maximal camera center norm scaled by a suitable value $s_r$:
\begin{equation}
	s = \max \{\norm{\V{o}_i - \point_o}_2\} / s_r.
\end{equation}
We chose $s_r$ such that it is slightly larger than the ratio of the camera-to-object distance to the object size.
For \diligentmv~\cite{li2020multi} objects, we set $s_r=10$ as they are captured about \SI{1.5}{m} away from about \SI{20}{cm} height objects.
For \pandora~\cite{dave2022pandora} and our objects, we set $s_r=3$.

\end{appendices}

%% file: sections/figures_tables/supp_tsc_halfpi_comparison.tex
\begin{figure}
	\small
	\centering
	\begin{tabular}{@{}c@{}c@{}c@{}c@{}}
		Color image & Polar-azimuth & w/o \halfpi & w/ \halfpi \\
		\includegraphics[width=0.24\linewidth]{../images/pandora_results/vase/02} &
		\includegraphics[width=0.24\linewidth]{../images/pandora_results/vase/azimuth/crop_azimuth_view_2_large} &
		\includegraphics[width=0.24\linewidth]{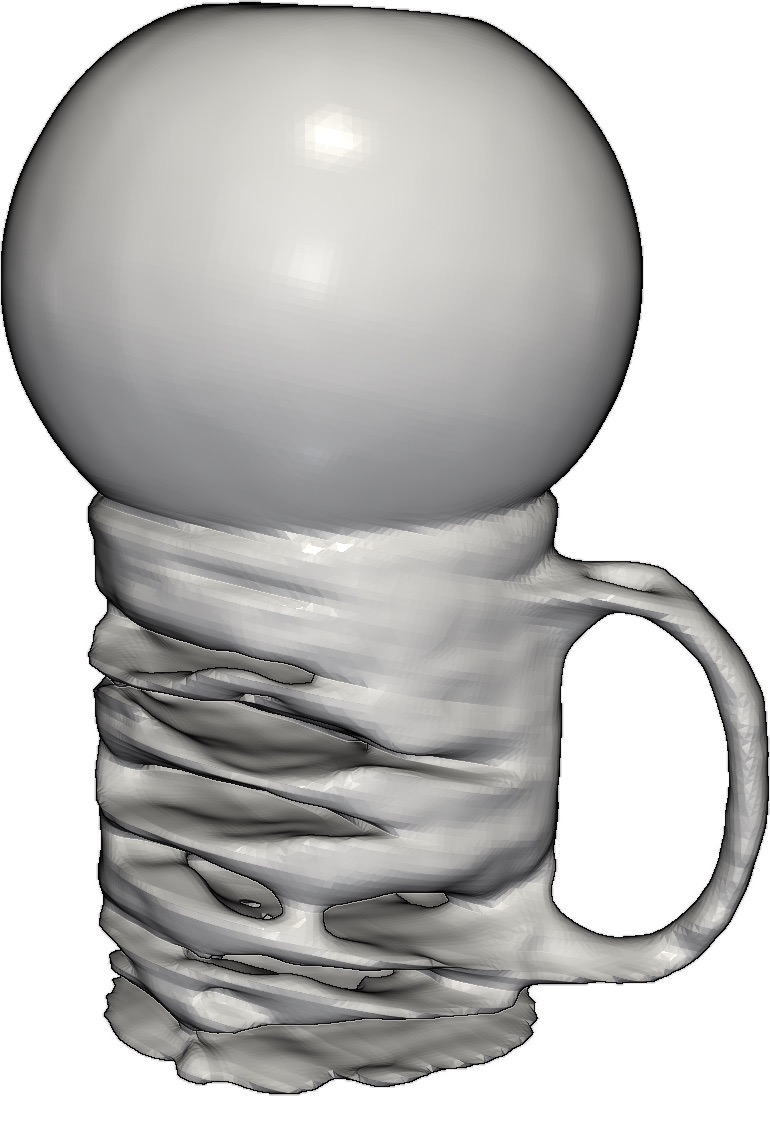} &
		\includegraphics[width=0.24\linewidth]{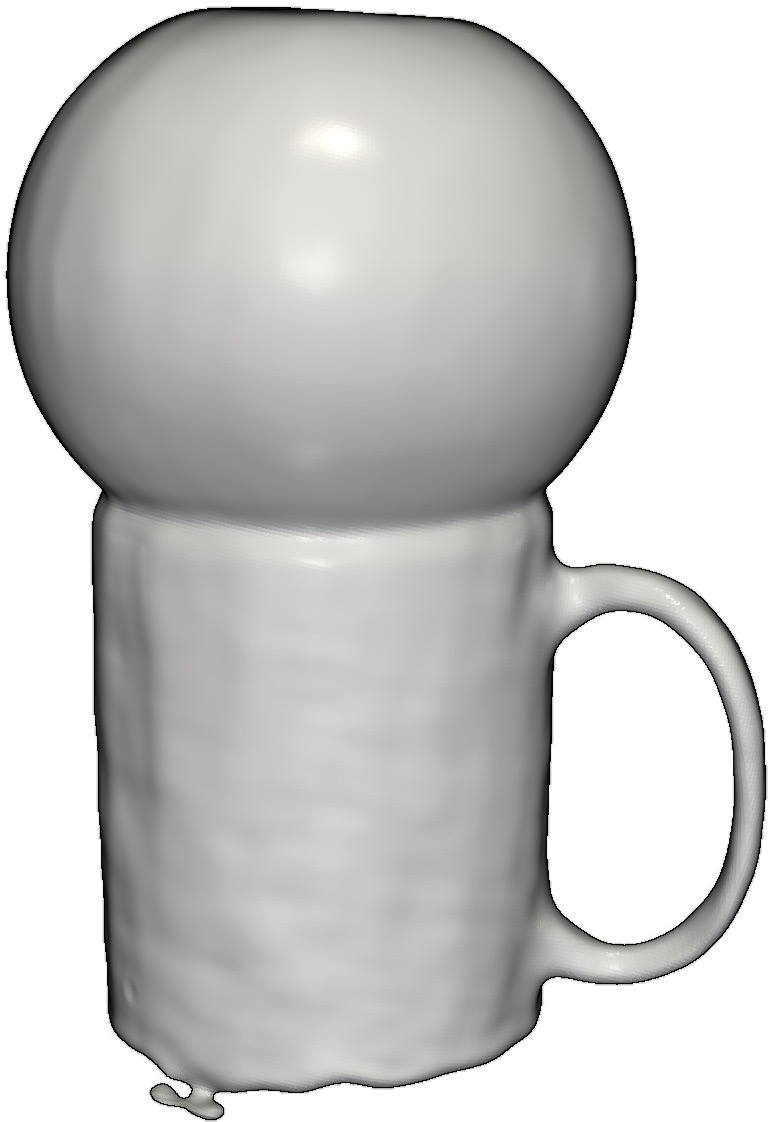}
		\\
		\includegraphics[width=0.24\linewidth]{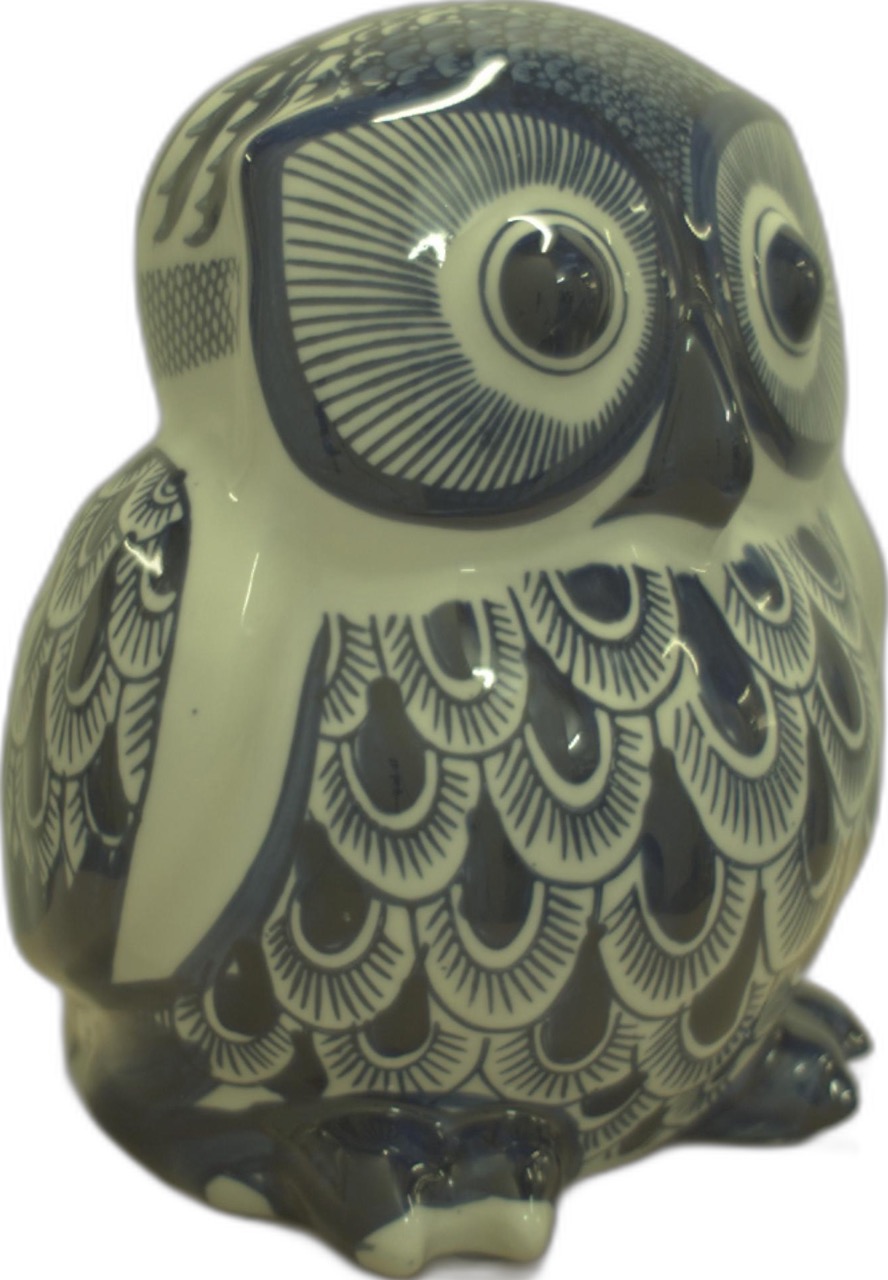} &
		\includegraphics[width=0.24\linewidth]{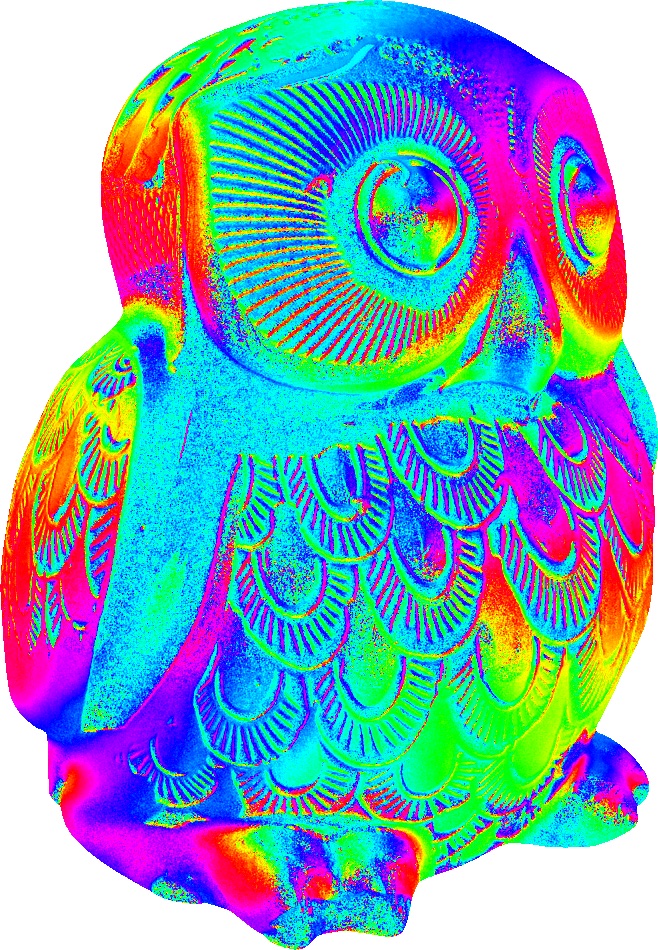} &
		\includegraphics[width=0.24\linewidth]{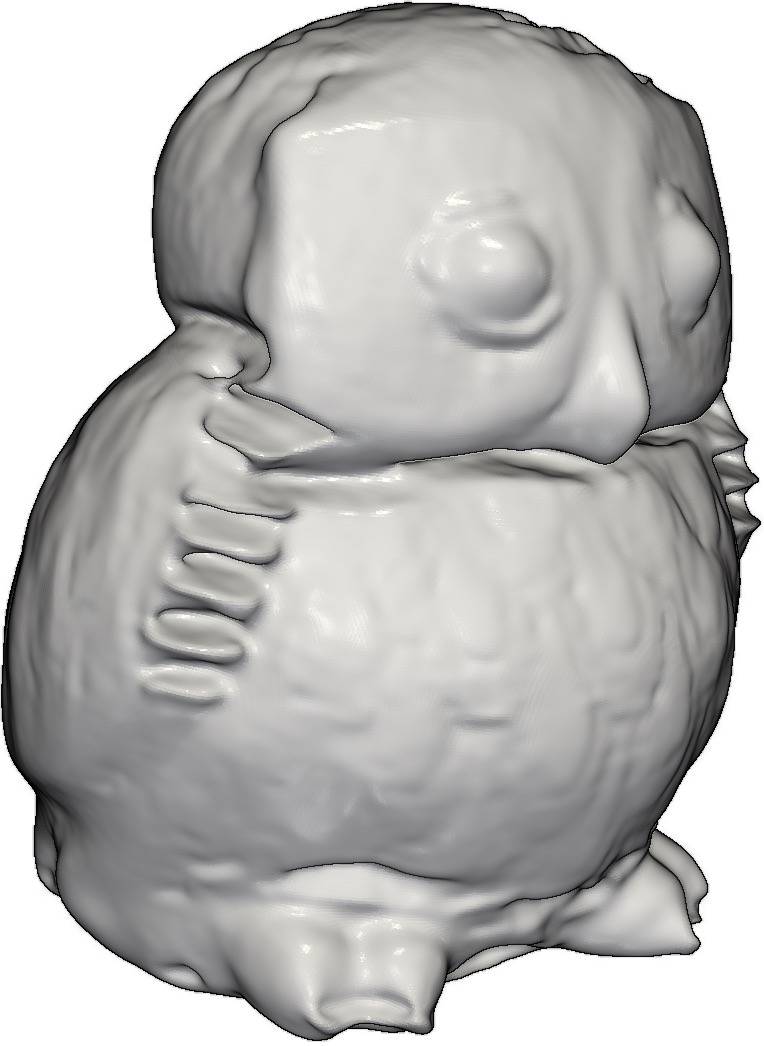} &
		\includegraphics[width=0.24\linewidth]{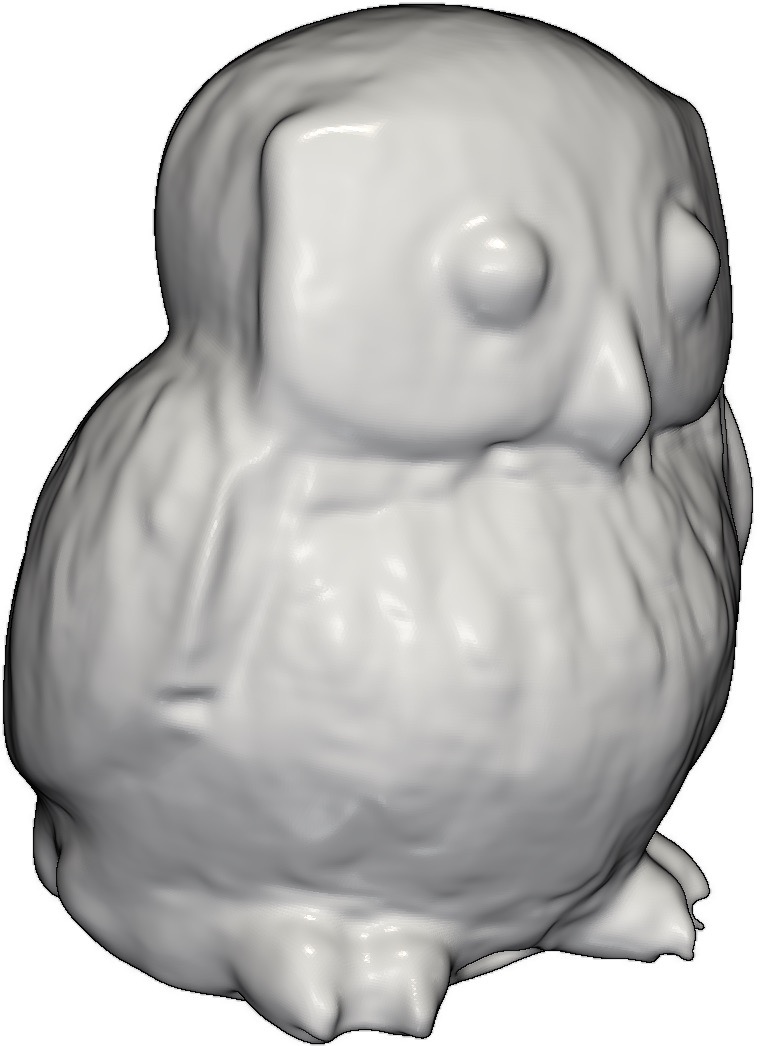}
	\end{tabular}
\caption{Accounting for the \halfpi ambiguity in \tsc loss resolves the \emph{twisted} surface problem. The results by dealing with the \halfpi ambiguity are presented in the main paper Fig.~11. }
\label{fig.tsc_halfpi}
\end{figure}

%% file: sections/figures_tables/ablation.tex
\begin{figure}[t]
	\scriptsize
	\centering
	\begin{tabularx}{\linewidth}{
			>{\centering\arraybackslash}X
			>{\centering\arraybackslash}X
			>{\centering\arraybackslash}X}
		w/o multiview consistency &
		w/ multiview consistency & 
		GT 
	\end{tabularx}
	\includegraphics[width= \linewidth]{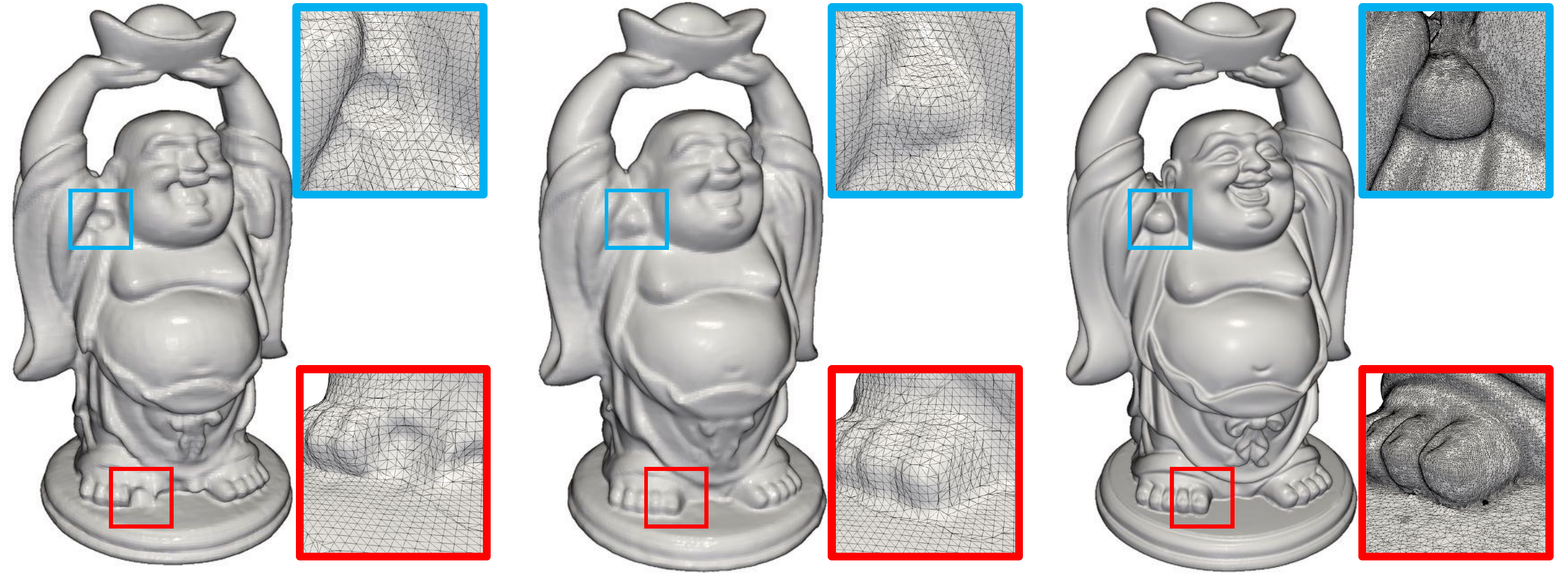}
	\caption{Considering multiview consistency resolves the convex-concave ambiguity because it encourages accurate correspondence.}
	\label{fig.ablation_on_reprojection}
\end{figure}

%% file: sections/figures_tables/diligent_inner.tex
\begin{figure}
	\small
	\centering
	\begin{tabular}{@{}c@{}c@{}c@{}c@{}c@{}}
		Bear & Buddha & Cow & Pot2 & Reading \\
			\includegraphics[height=0.2\linewidth]{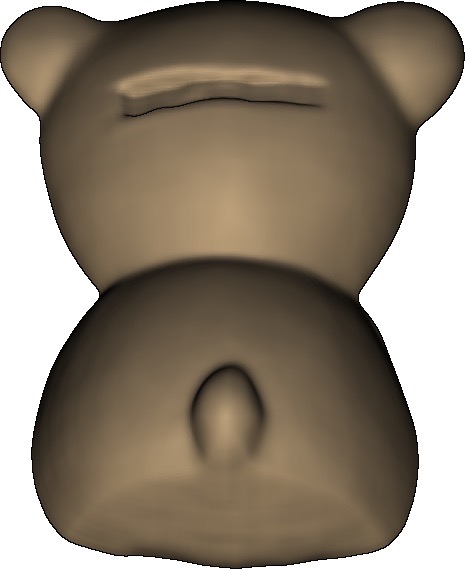} &
			\includegraphics[height=0.2\linewidth]{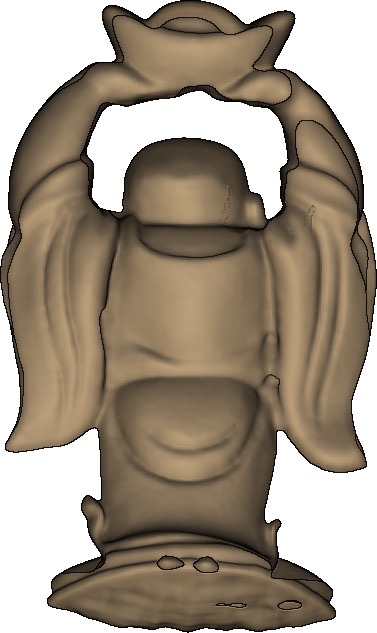} &
			\includegraphics[height=0.2\linewidth]{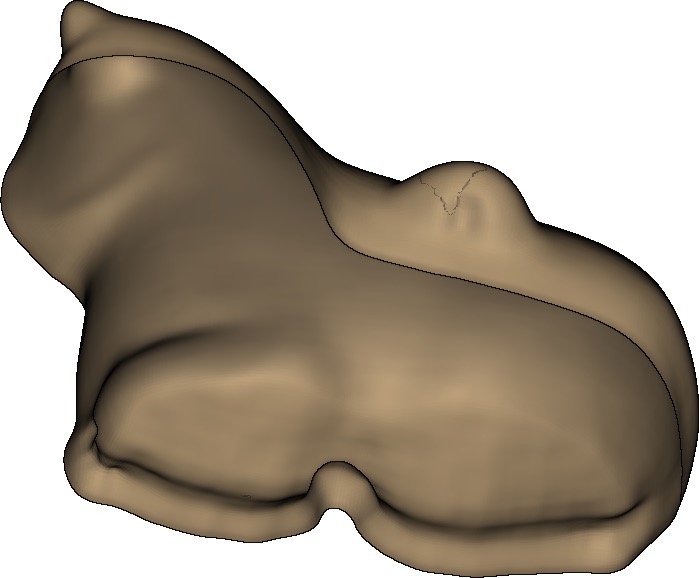} &
			\includegraphics[height=0.2\linewidth]{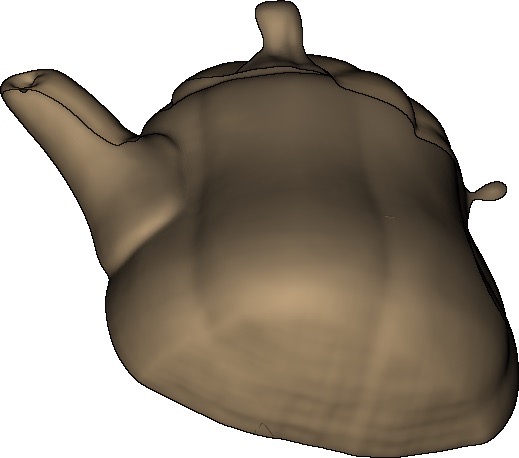} &
			\includegraphics[height=0.2\linewidth]{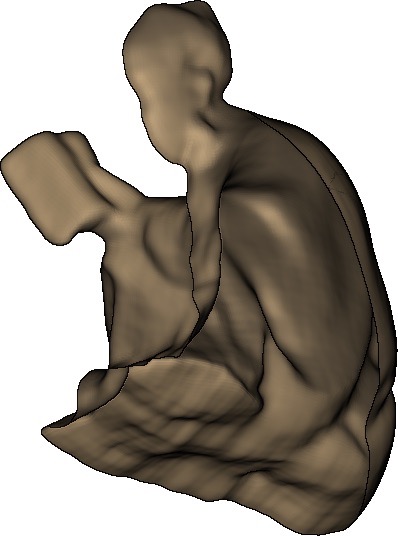} \\
			\includegraphics[height=0.2\linewidth]{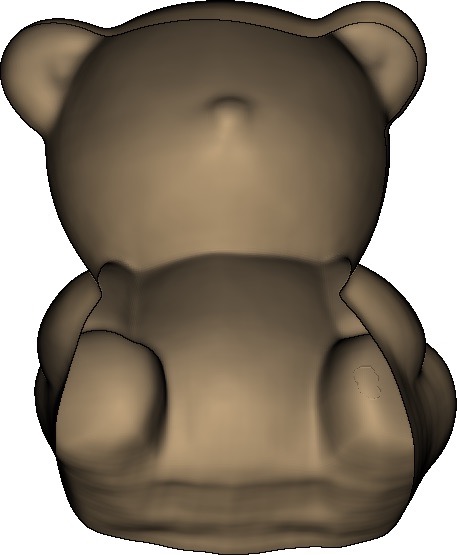} &
			\includegraphics[height=0.2\linewidth]{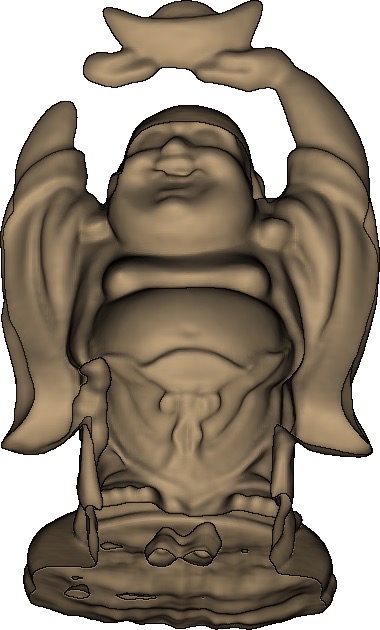} &
			\includegraphics[height=0.2\linewidth]{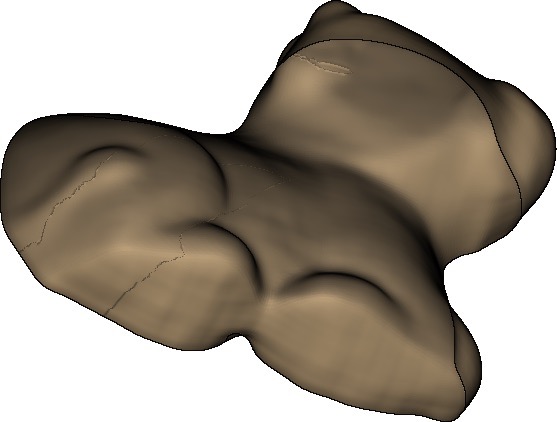} &
			\includegraphics[height=0.2\linewidth]{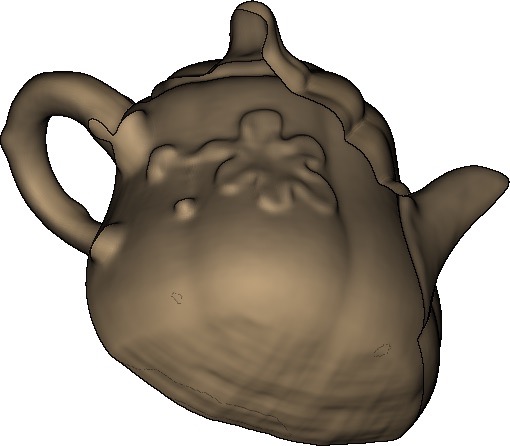} &
			\includegraphics[height=0.2\linewidth]{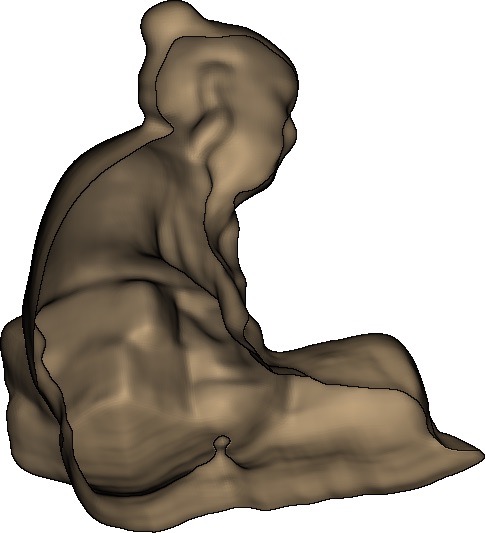} 
	\end{tabular}
	\caption{Visualization of the inner space of our recovered surfaces (cut in half vertically).
		We consider that our evaluation of shape accuracy using visible surface points is fair because the inner space is clean. No post-processing is performed on the meshes after we extract them using marching cubes~\cite{lorensen1987marching}.}
	\label{fig.diligent_inner}
\end{figure}

%% file: sections/figures_tables/diligent_shape_supp.tex
\begin{figure}[t!]
	\tiny
	\newcommand{\figwidtha}{0.16}
	\centering
	\begin{tabular}{c@{}c@{}c@{}c@{}c@{}c}
		\rmvps~\cite{park2016robust} & \bmvps~\cite{li2020multi} & \uanet~\cite{kaya2022uncertainty} & \psnerf~\cite{yang2022psnerf} & \mvas (ours)  & GT \\
		\includegraphics[width=\figwidtha\linewidth]{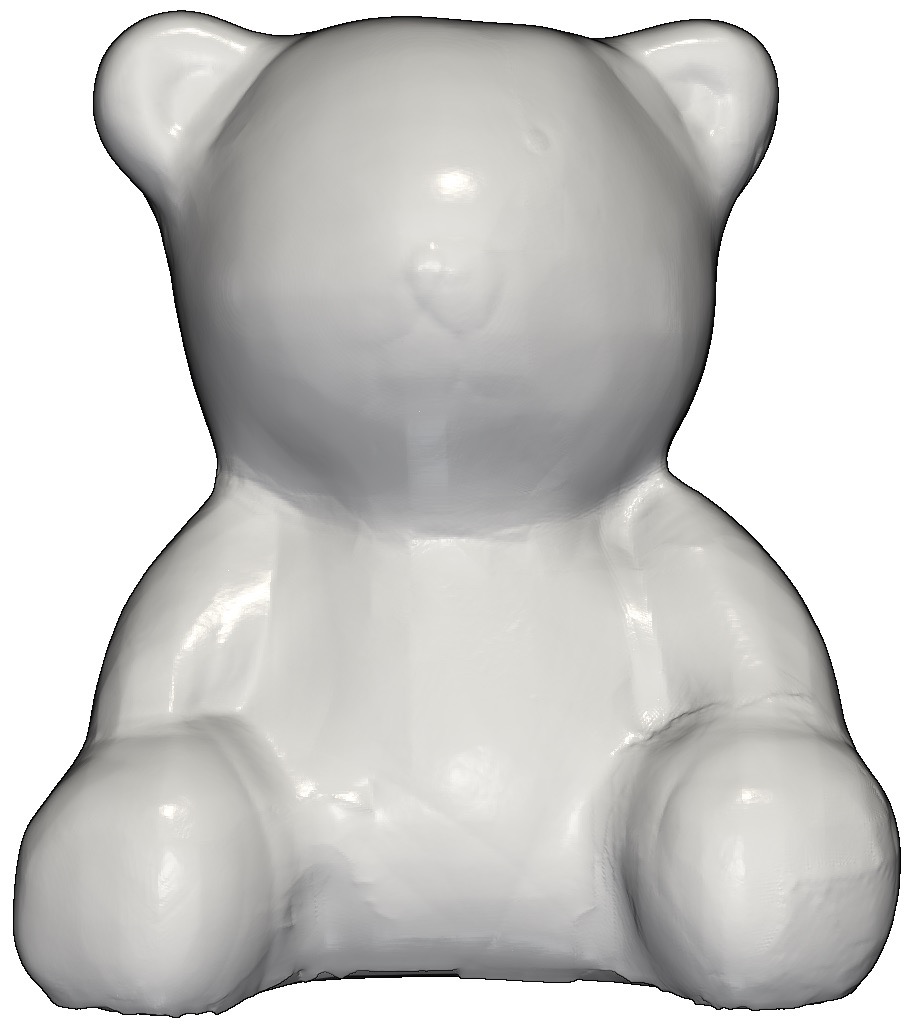} &
		\includegraphics[width=\figwidtha\linewidth]{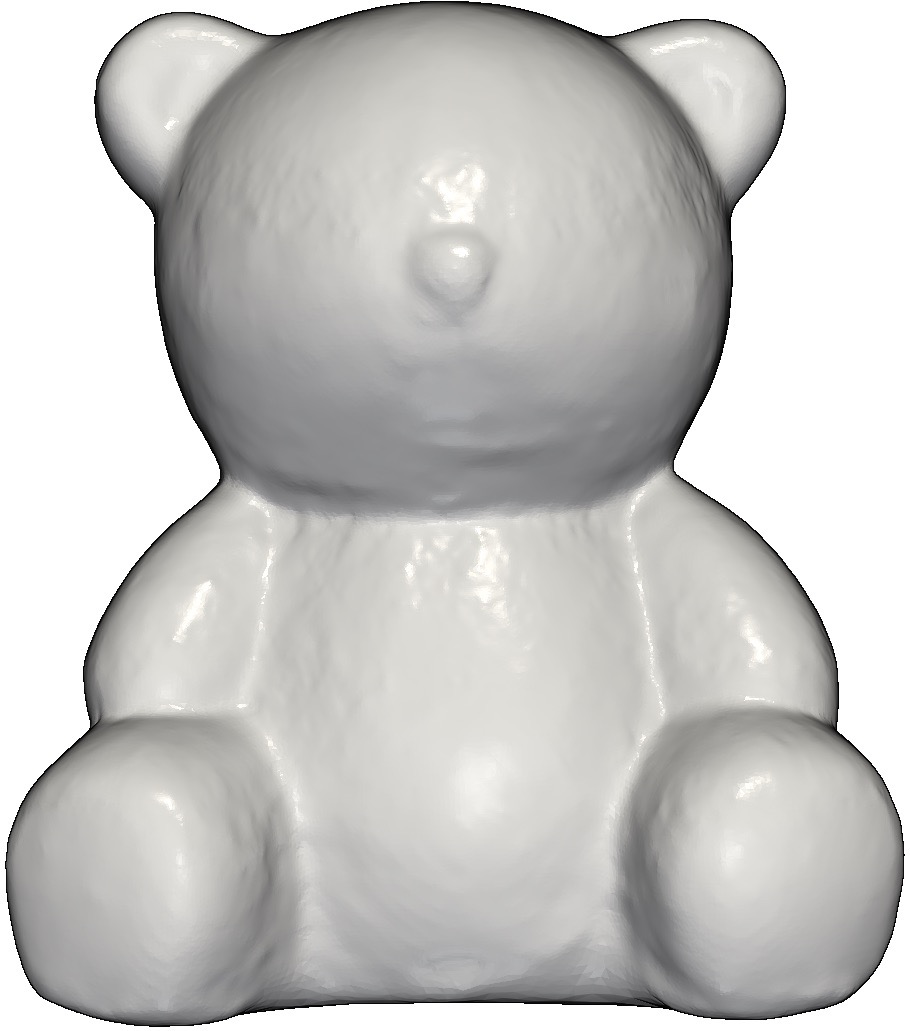} &
		\includegraphics[width=\figwidtha\linewidth]{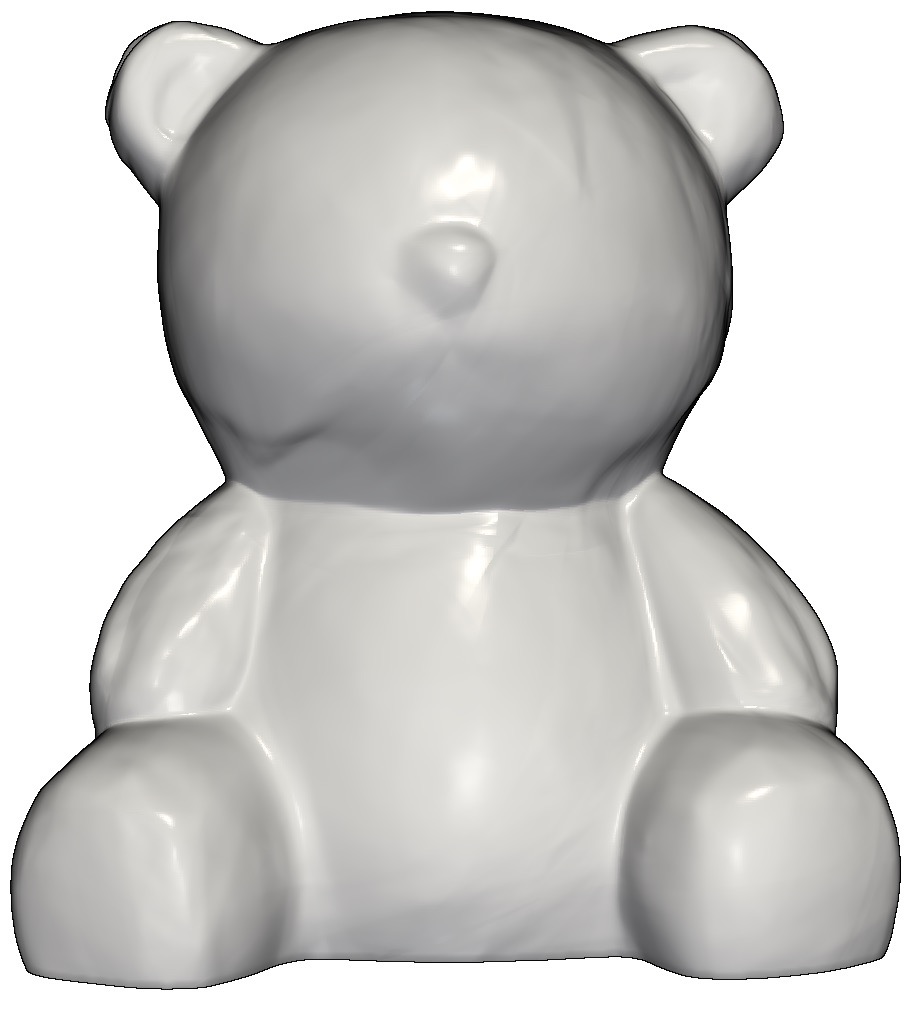} &
		\includegraphics[width=\figwidtha \linewidth]{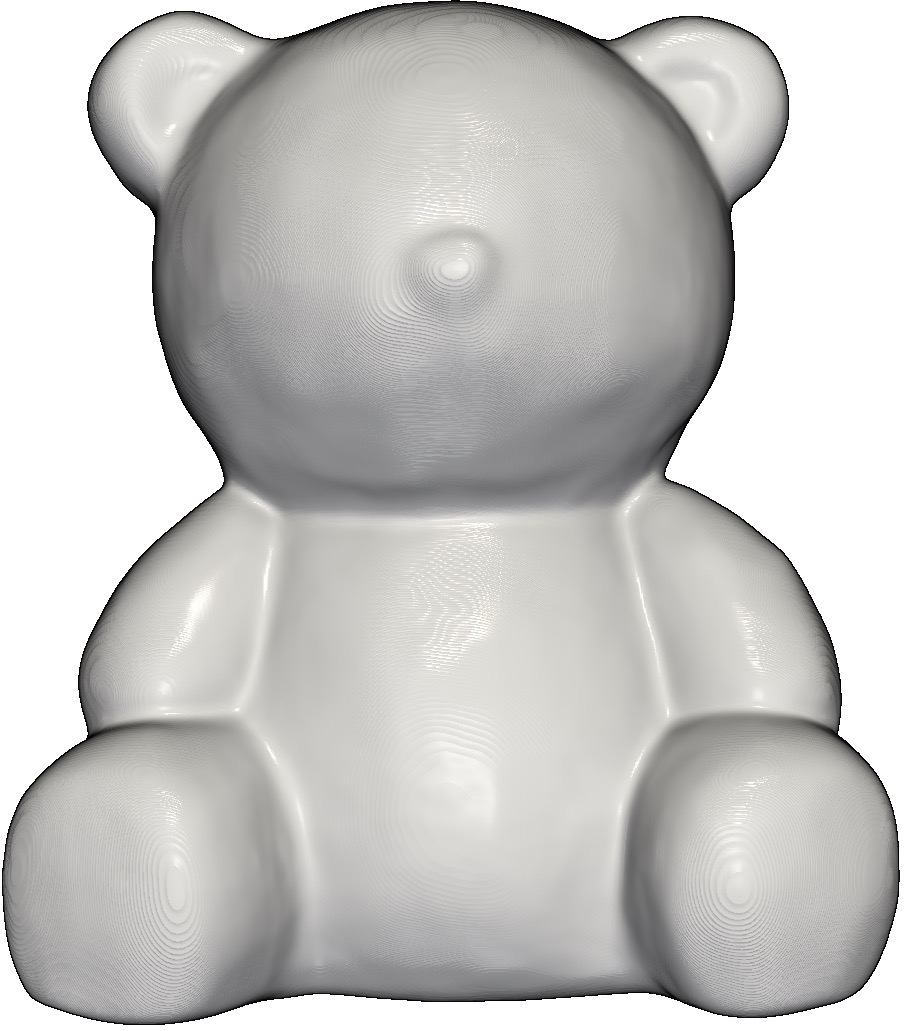} &
		\includegraphics[width=\figwidtha \linewidth]{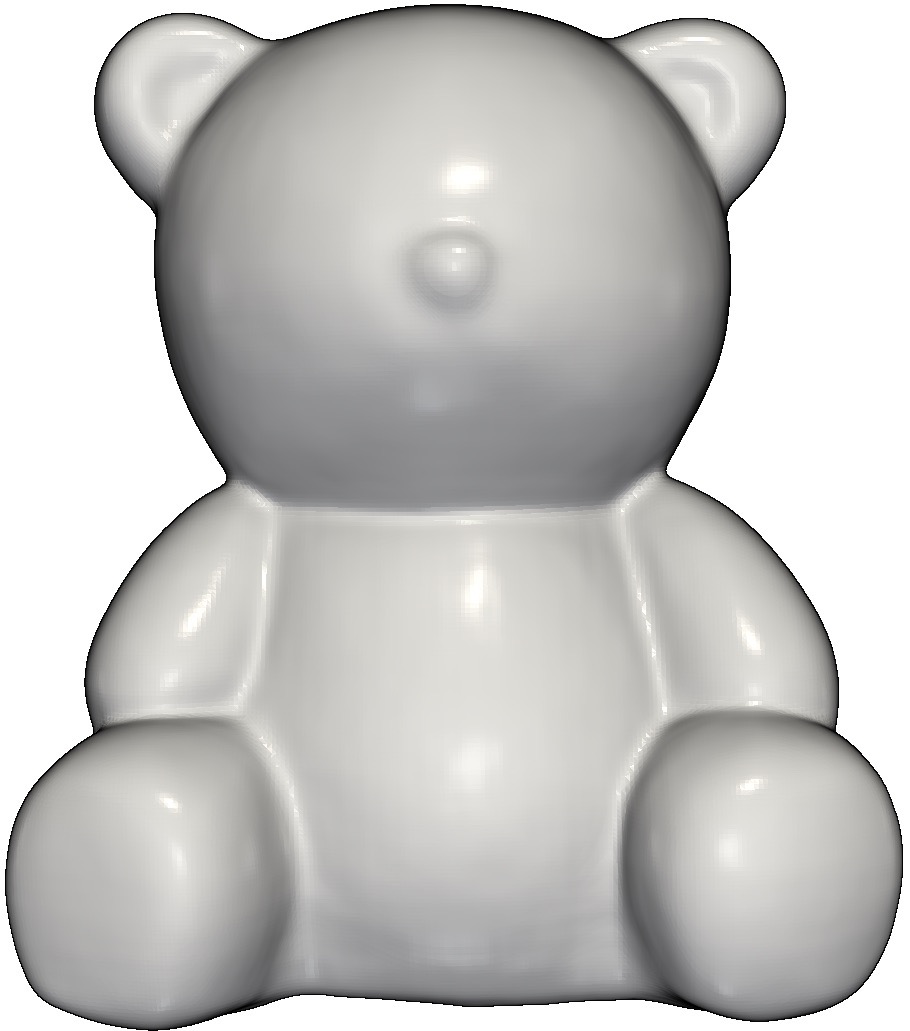} &
		\includegraphics[width=\figwidtha \linewidth]{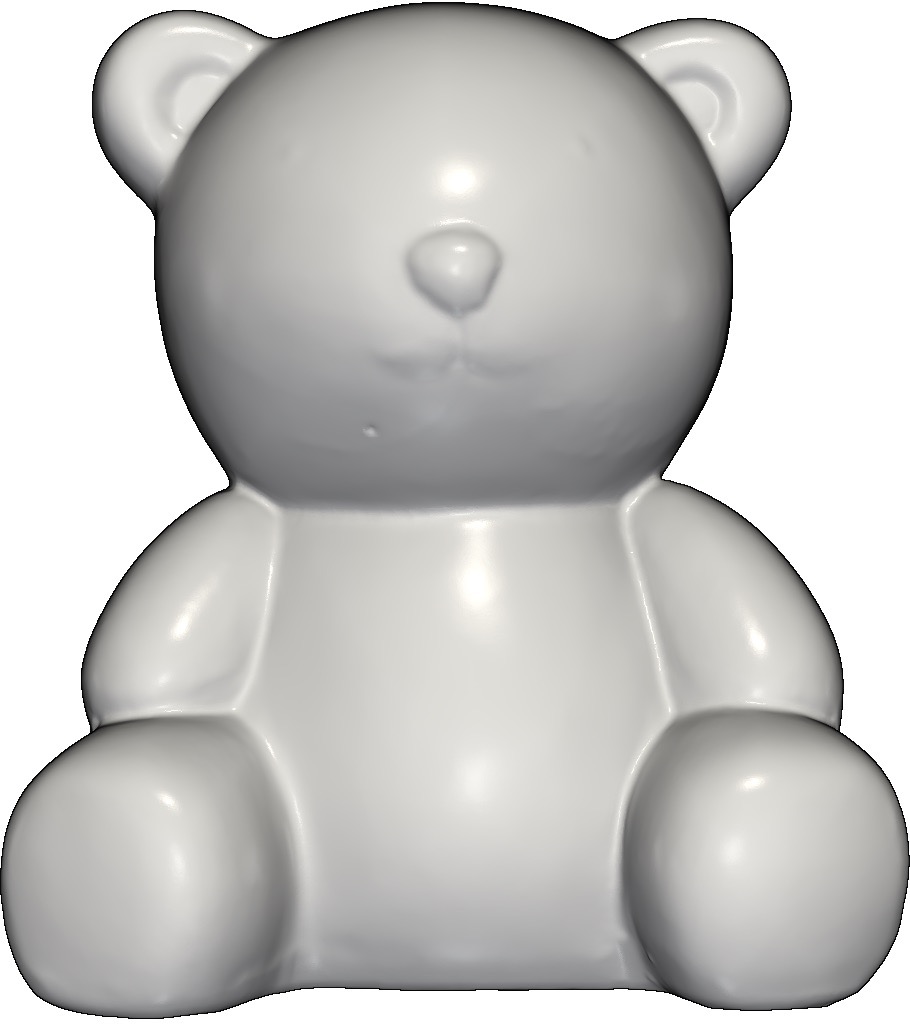}  
		\\
		\includegraphics[width=\figwidtha\linewidth]{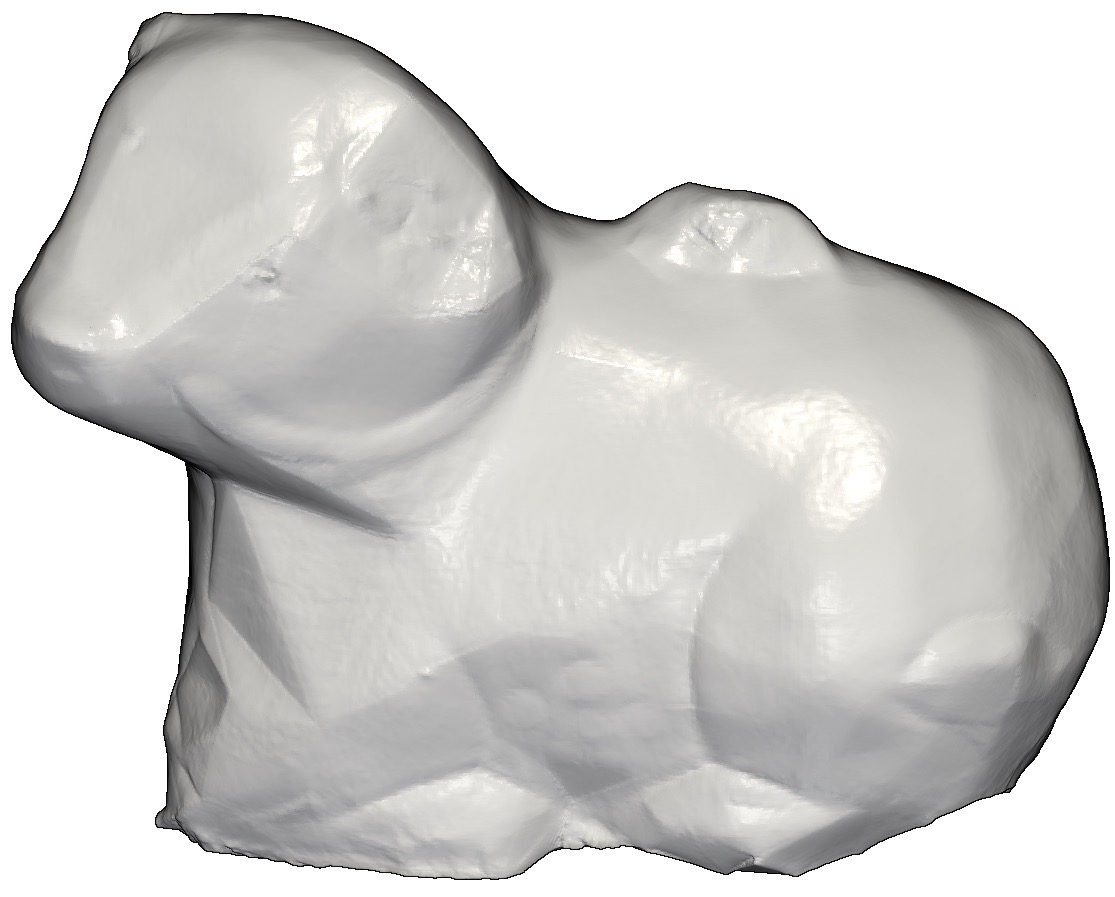} &
		\includegraphics[width=\figwidtha\linewidth]{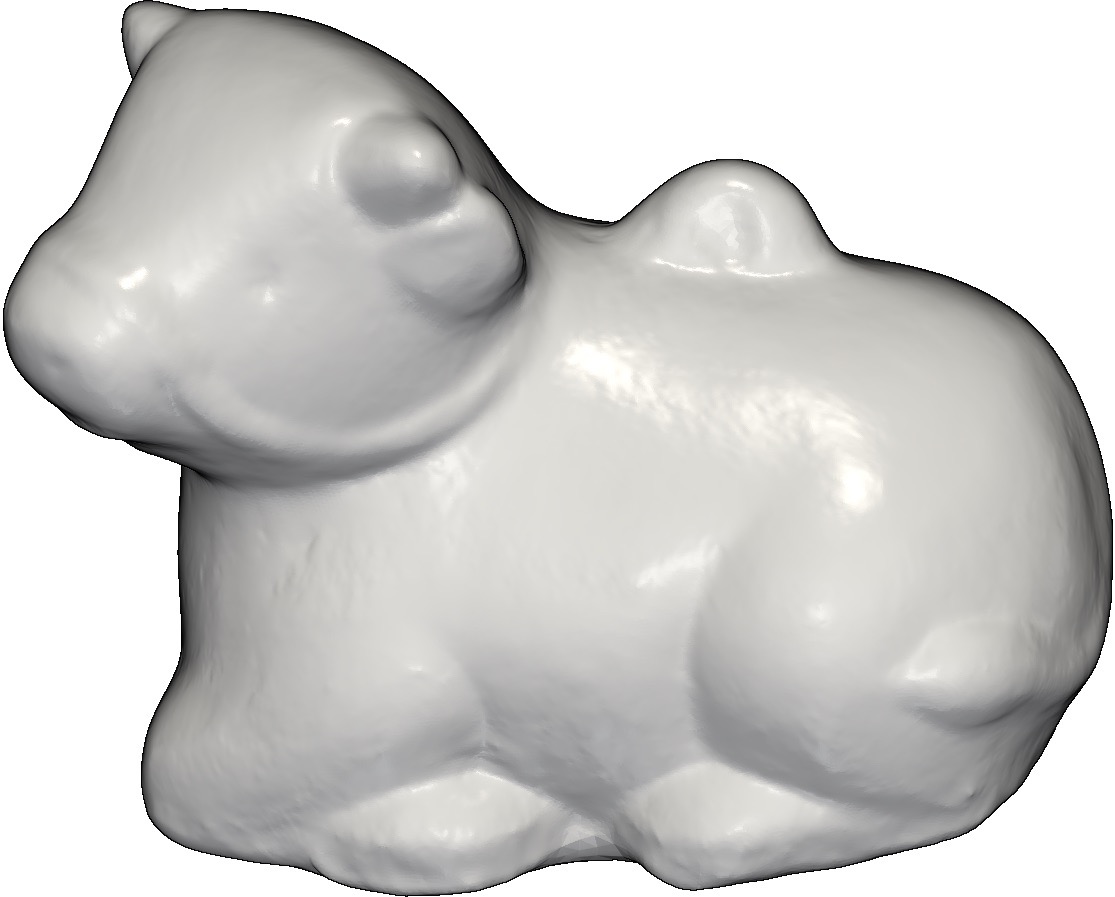} &
		\includegraphics[width=\figwidtha\linewidth]{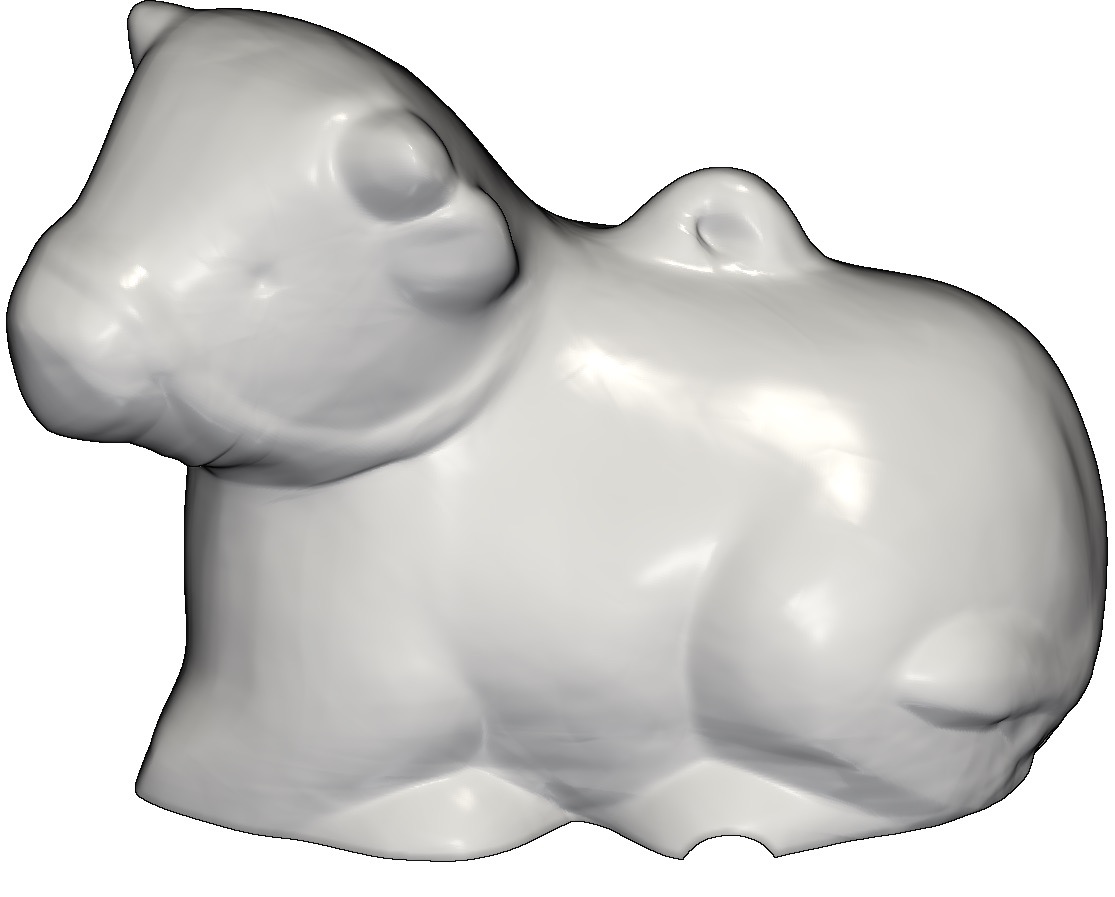} &
		\includegraphics[width=\figwidtha \linewidth]{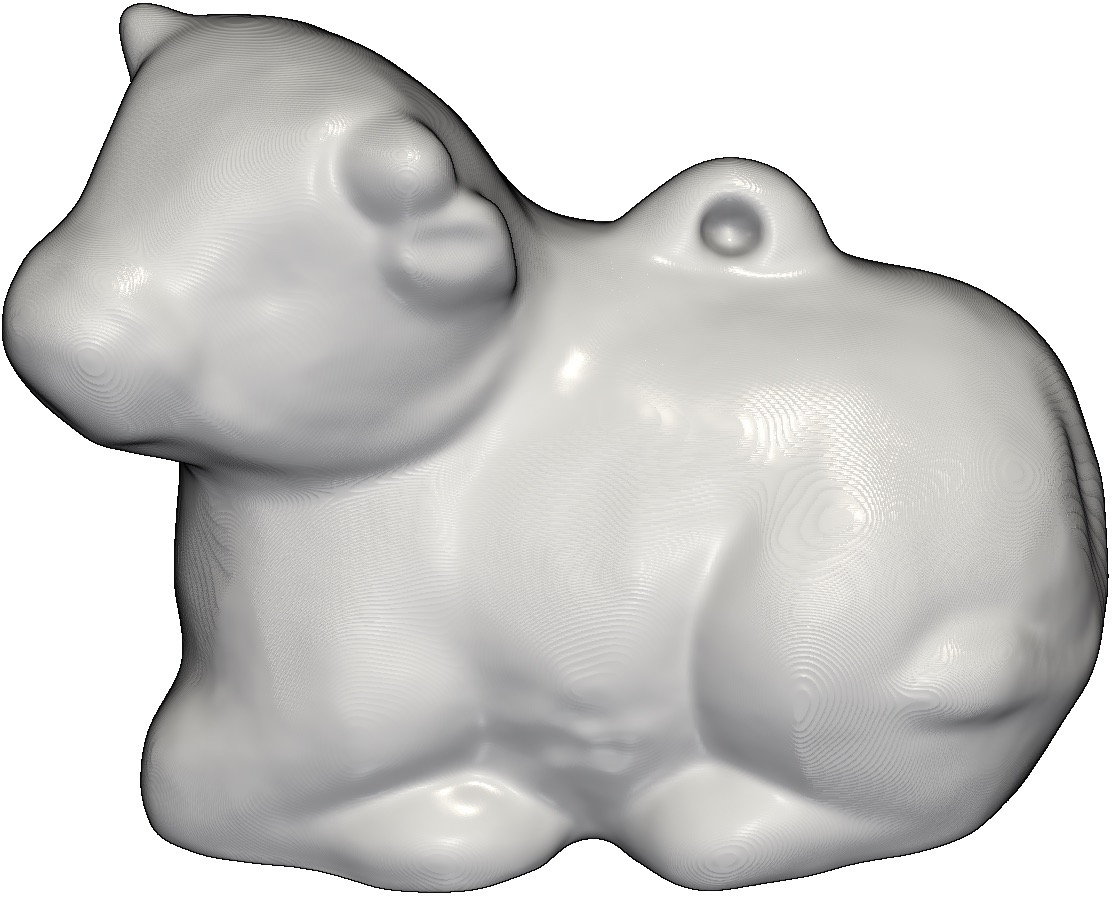} &
		\includegraphics[width=\figwidtha \linewidth]{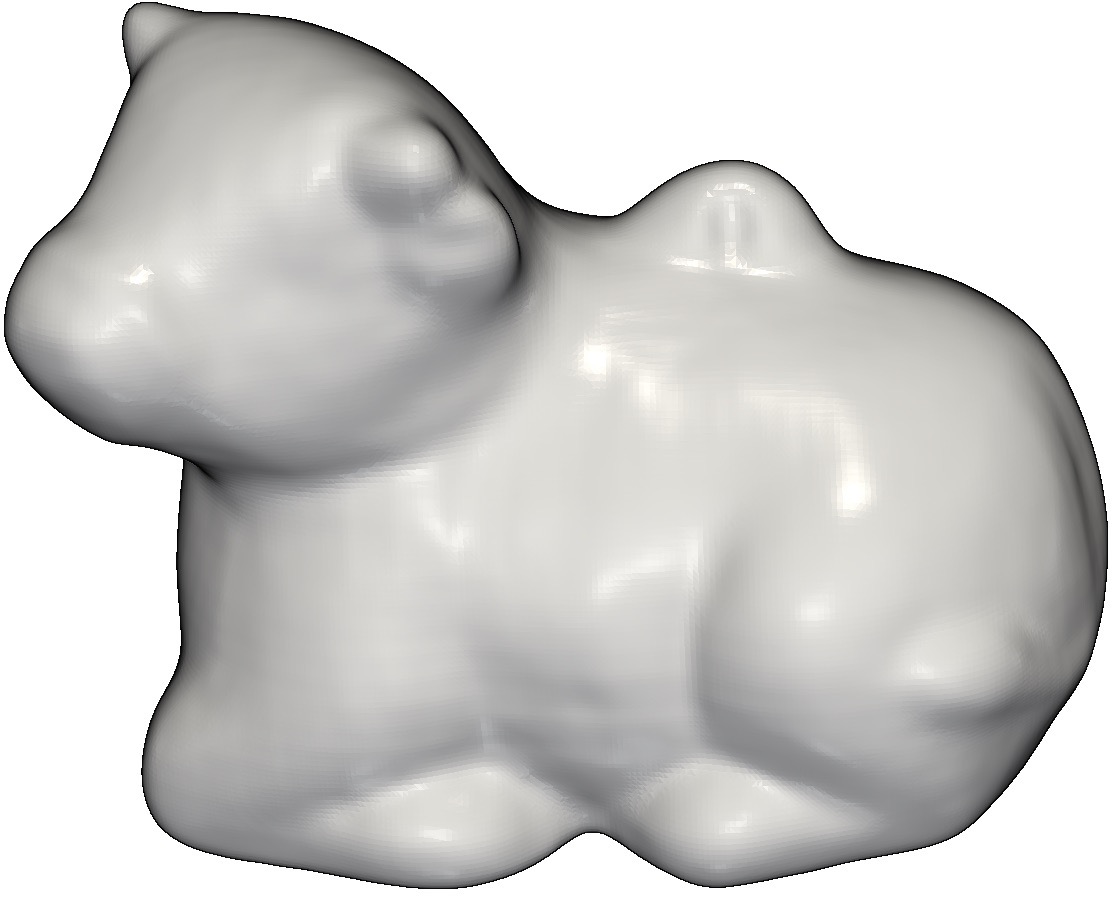} &
		\includegraphics[width=\figwidtha \linewidth]{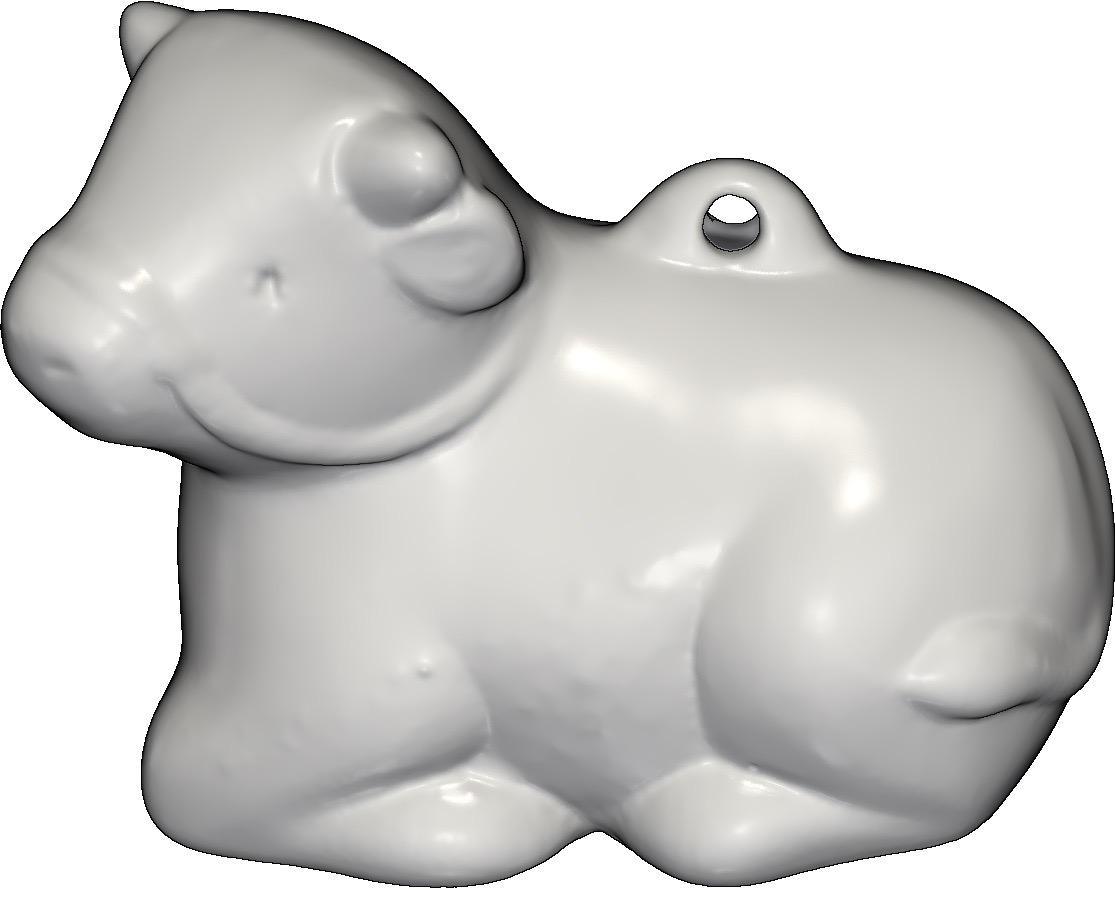} 
		\\
		\includegraphics[width=\figwidtha\linewidth]{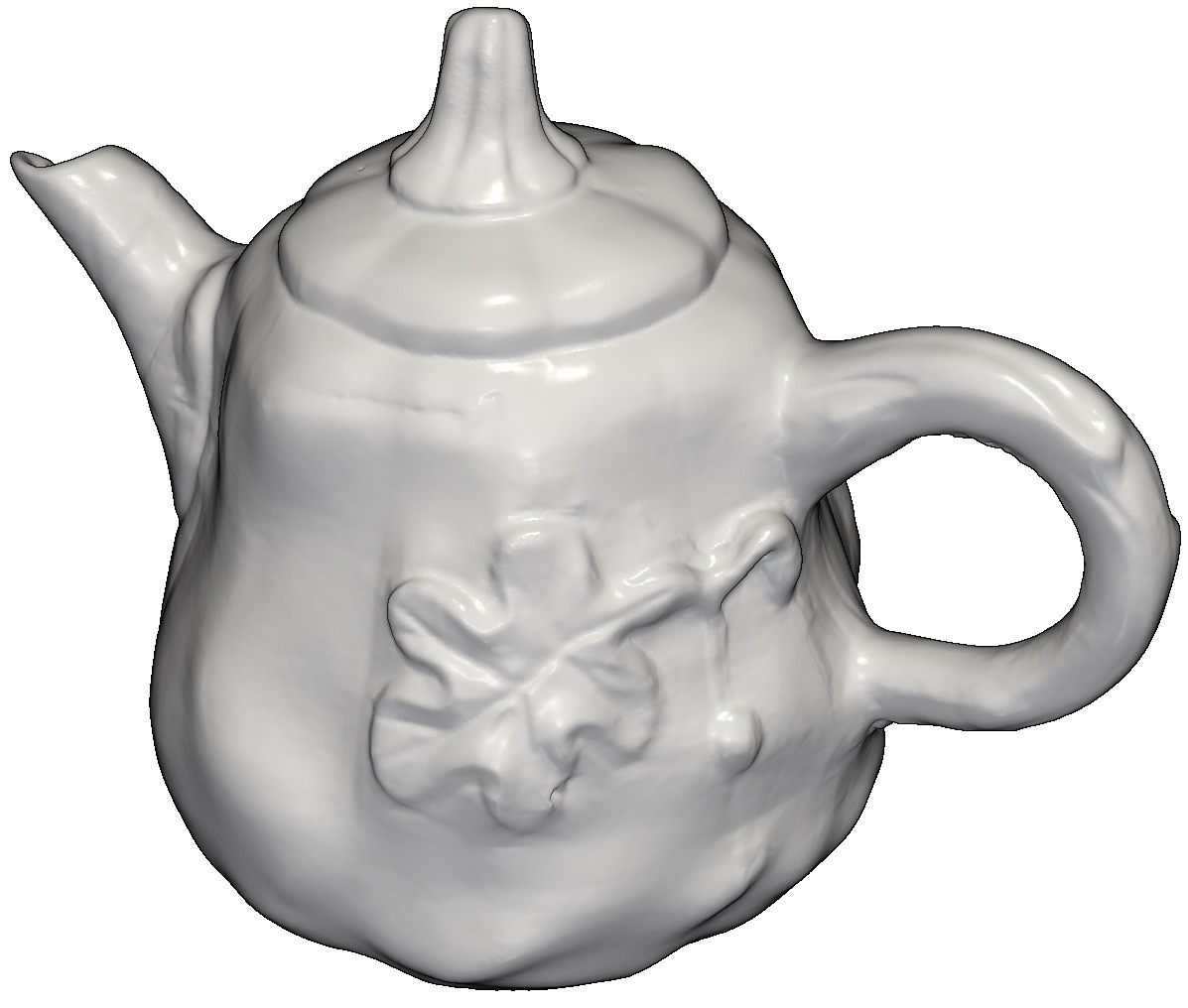} &
		\includegraphics[width=\figwidtha\linewidth]{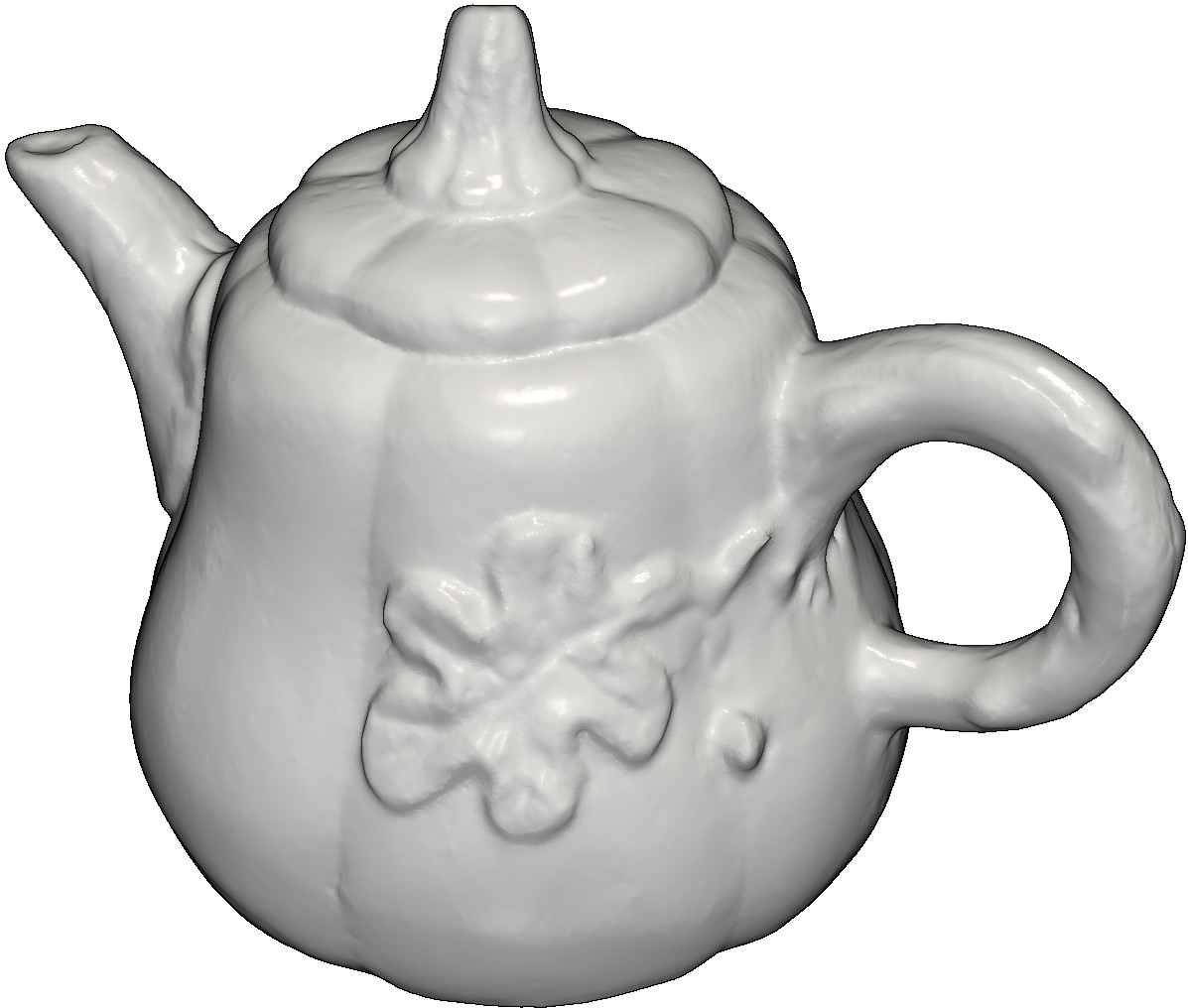} &
		\includegraphics[width=\figwidtha\linewidth]{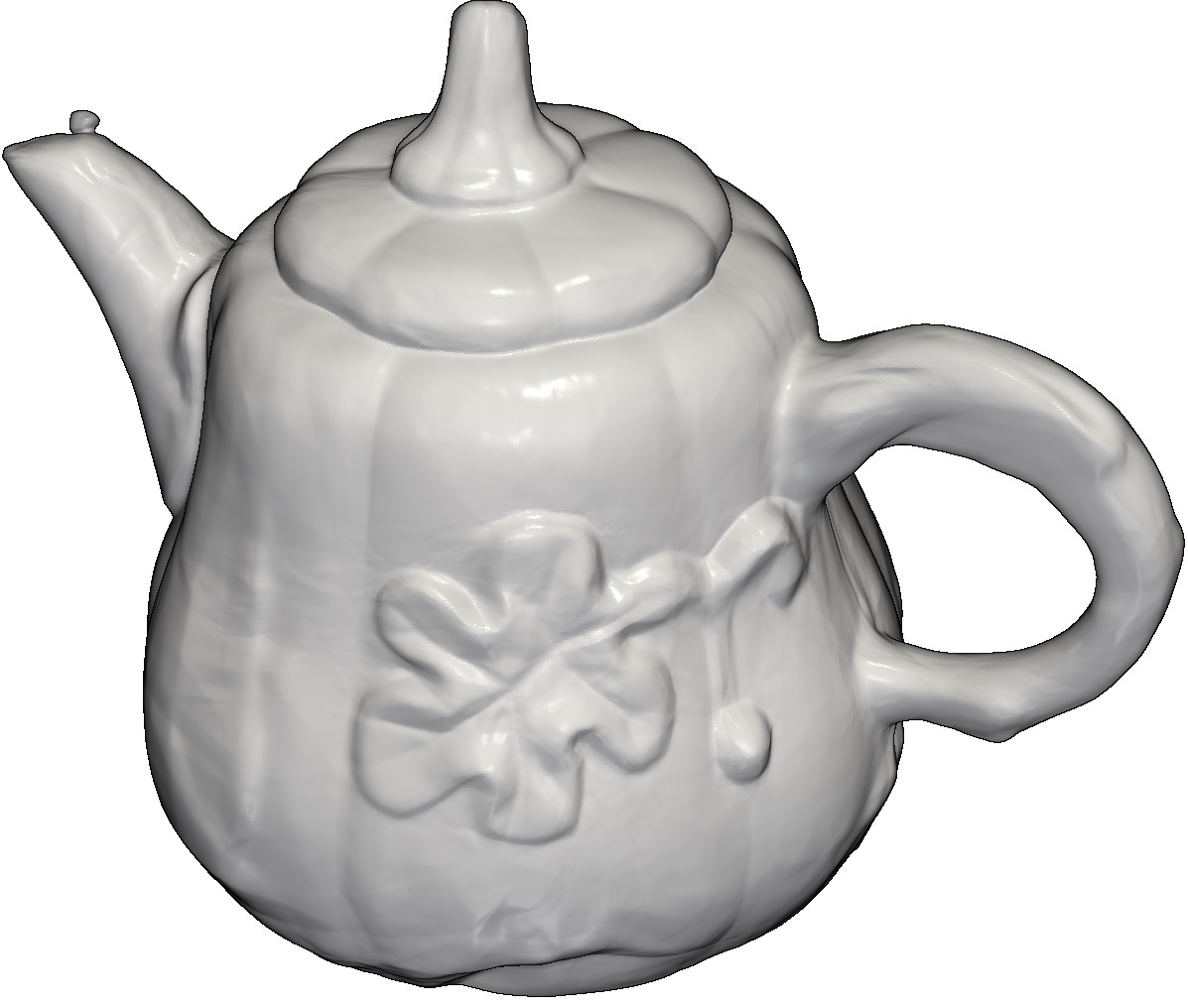} &
		\includegraphics[width=\figwidtha \linewidth]{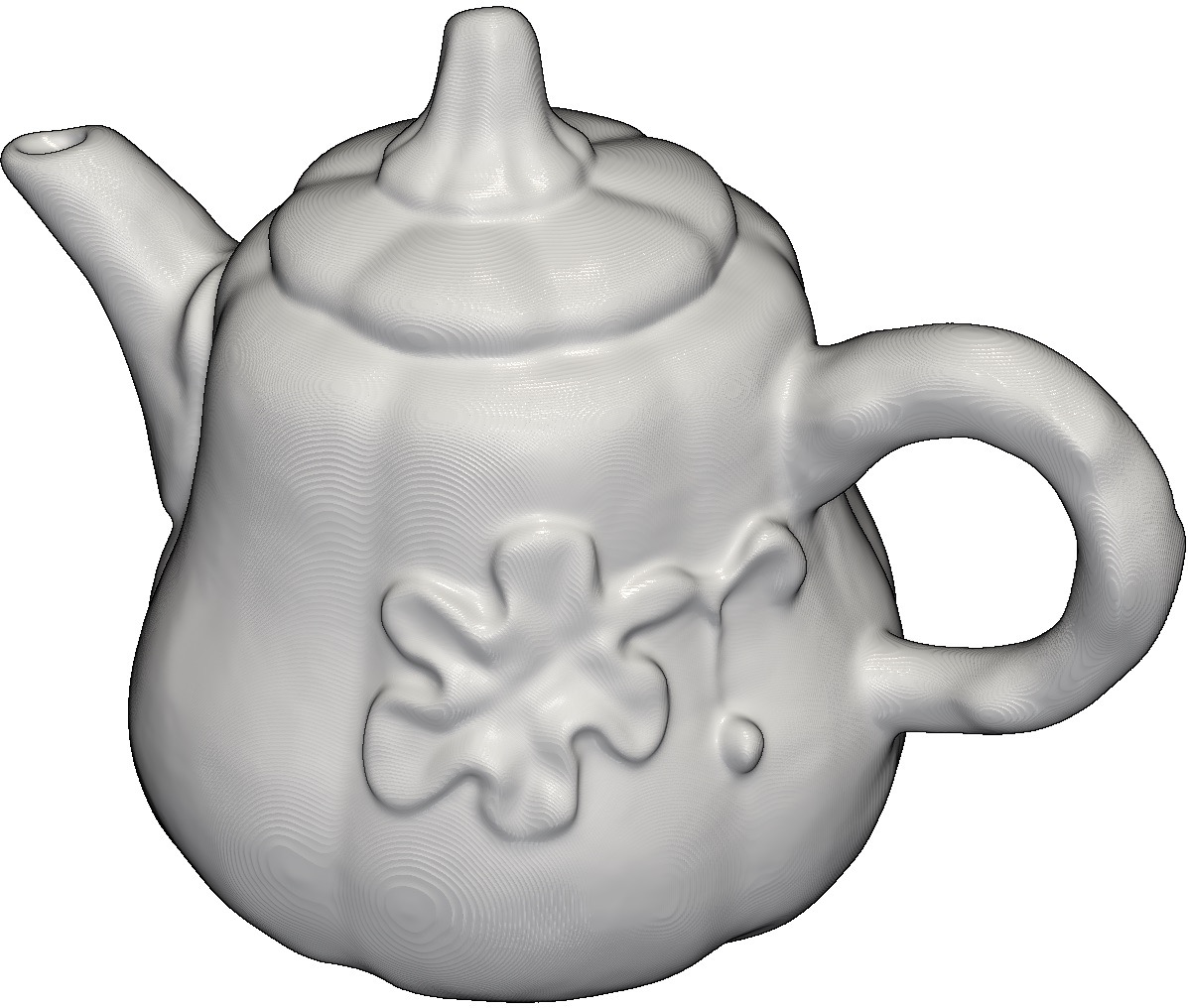} &
		\includegraphics[width=\figwidtha \linewidth]{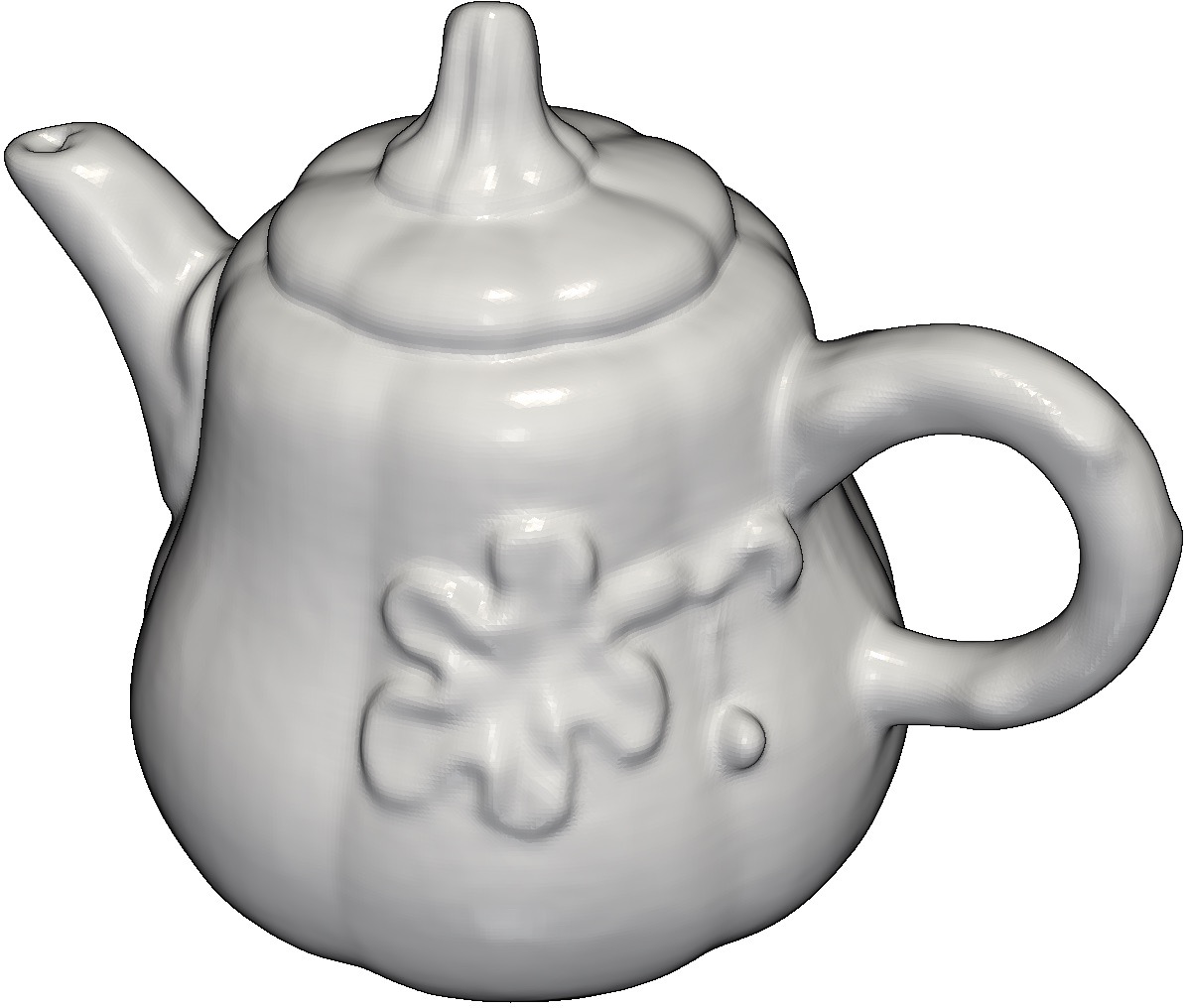} &
		\includegraphics[width=\figwidtha \linewidth]{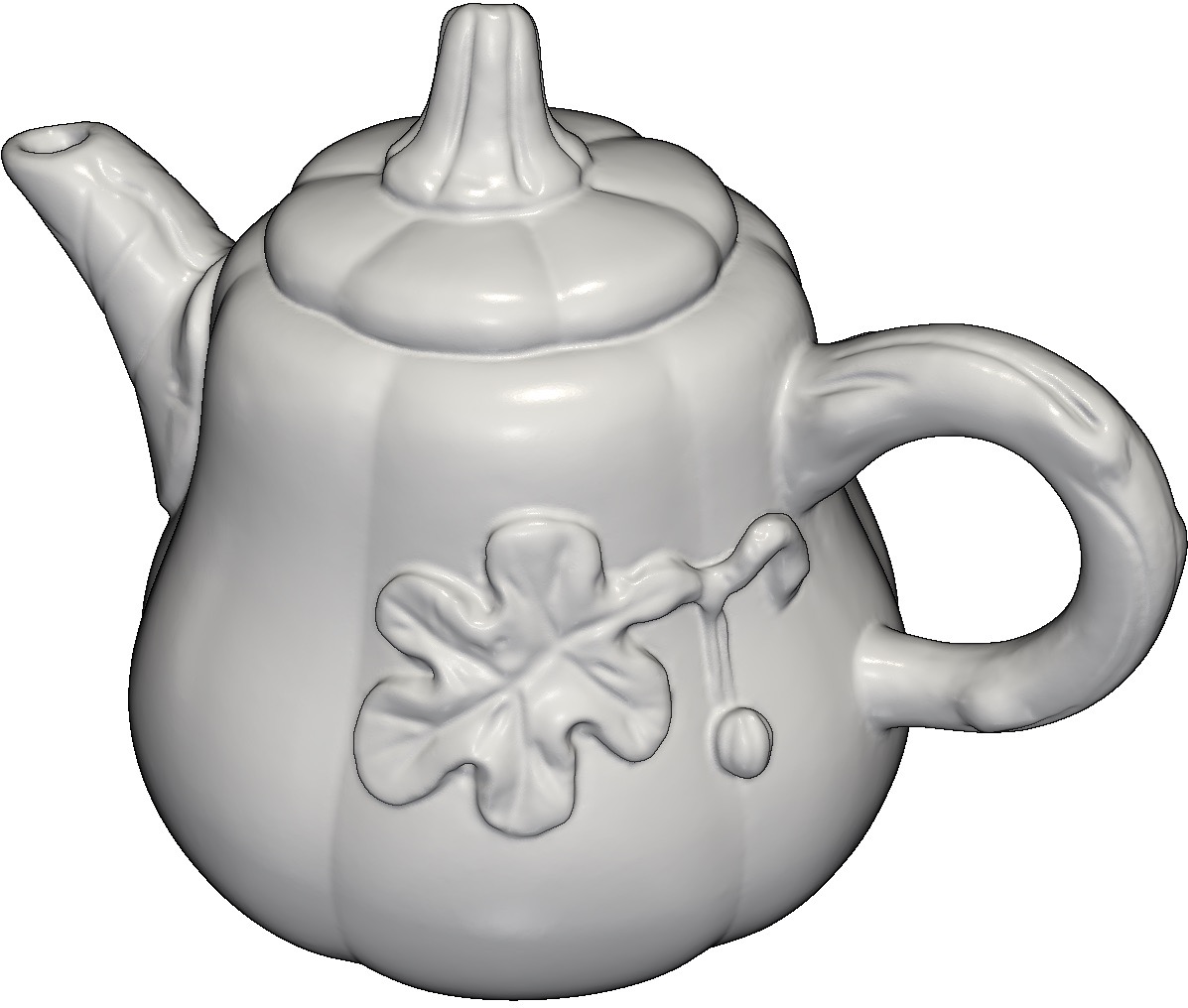} 
	\end{tabular}
	\caption{More visual comparisons of recovered shapes of \diligentmv~\cite{li2020multi} objects ``Bear,''``Cow,'' and ``Pot2.''}
	\label{fig.vis_comp_mvps_supp}
\end{figure}

%% file: sections/figures_tables/diligent_normal_supp.tex
\begin{figure}[h!]
	\scriptsize
	\newcommand{\figwidthNormalVisSupp}{0.16}
	\centering
	\begin{tabular}{@{}c@{}c@{}c@{}c@{}c@{}c@{}}
		\rmvps~\cite{park2016robust} & \bmvps~\cite{li2020multi} & \psnerf~\cite{yang2022psnerf} & \sdps~\cite{chen2019SDPS_Net} & \mvas (ours) & GT \\
		\includegraphics[width=\figwidthNormalVisSupp\linewidth]{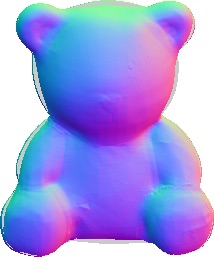}&
		\includegraphics[width=\figwidthNormalVisSupp\linewidth]{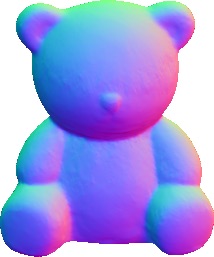}
		&
		\includegraphics[width=\figwidthNormalVisSupp\linewidth]{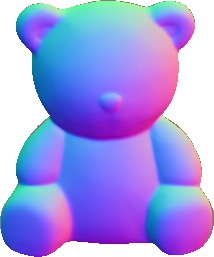}
		&
		\includegraphics[width=\figwidthNormalVisSupp\linewidth]{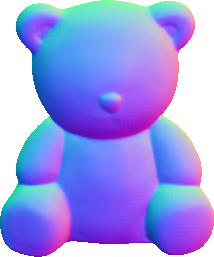}
		&
		\includegraphics[width=\figwidthNormalVisSupp\linewidth]{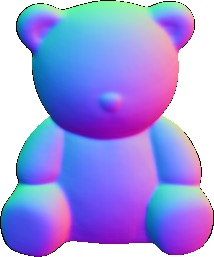}
		& 
		\includegraphics[width=\figwidthNormalVisSupp\linewidth]{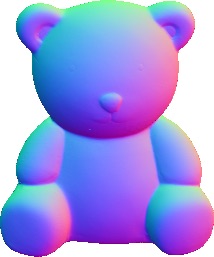}
		\\
		\includegraphics[width=\figwidthNormalVisSupp\linewidth]{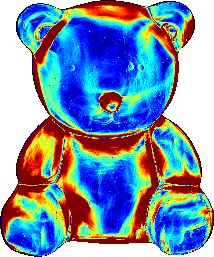}&
		\includegraphics[width=\figwidthNormalVisSupp\linewidth]{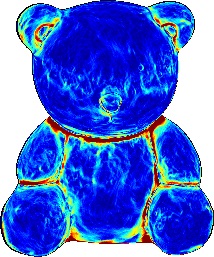}
		&
		\includegraphics[width=\figwidthNormalVisSupp\linewidth]{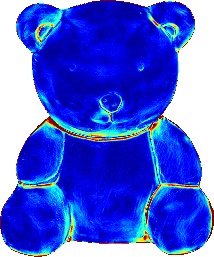}
		&
		\includegraphics[width=\figwidthNormalVisSupp\linewidth]{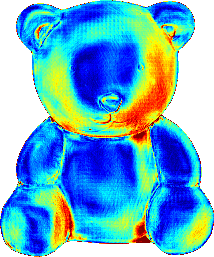}
		&
		\includegraphics[width=\figwidthNormalVisSupp\linewidth]{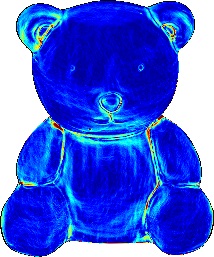}
		& 
		\colorbar{0.18}{$20^\circ$}{30}
		\\
		\includegraphics[width=\figwidthNormalVisSupp\linewidth]{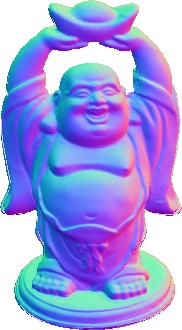}&
		\includegraphics[width=\figwidthNormalVisSupp\linewidth]{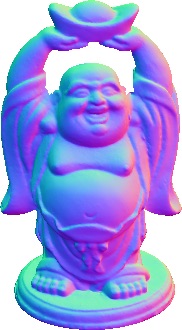}
		&
		\includegraphics[width=\figwidthNormalVisSupp\linewidth]{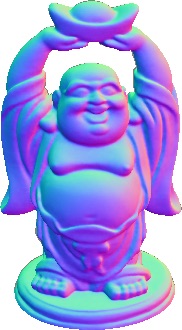}
		&
		\includegraphics[width=\figwidthNormalVisSupp\linewidth]{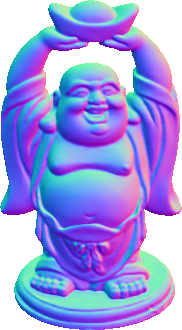}
		&
		\includegraphics[width=\figwidthNormalVisSupp\linewidth]{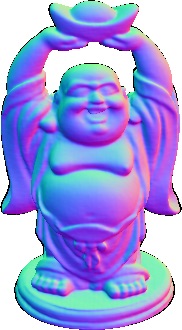}
		& 
		\includegraphics[width=\figwidthNormalVisSupp\linewidth]{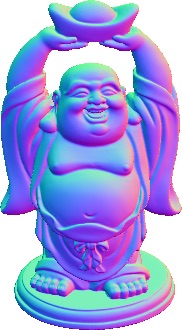}
		\\
		\includegraphics[width=\figwidthNormalVisSupp\linewidth]{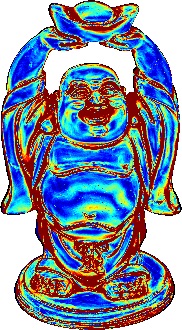}&
		\includegraphics[width=\figwidthNormalVisSupp\linewidth]{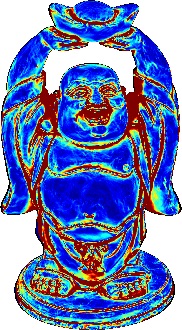}
		&
		\includegraphics[width=\figwidthNormalVisSupp\linewidth]{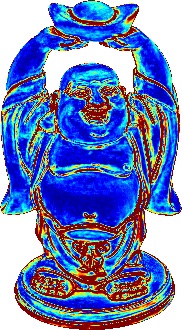}
		&
		\includegraphics[width=\figwidthNormalVisSupp\linewidth]{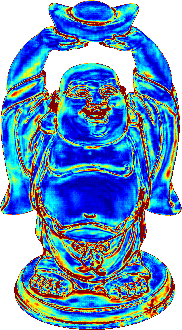}
		&
		\includegraphics[width=\figwidthNormalVisSupp\linewidth]{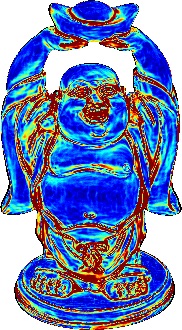}
		& 
		\colorbar{0.28}{$20^\circ$}{53}
		\\
		\includegraphics[width=\figwidthNormalVisSupp\linewidth]{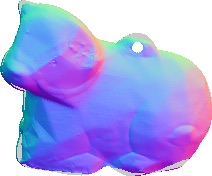}&
		\includegraphics[width=\figwidthNormalVisSupp\linewidth]{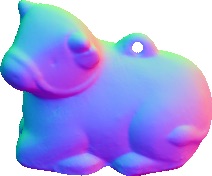}
		&
		\includegraphics[width=\figwidthNormalVisSupp\linewidth]{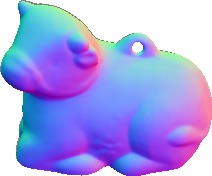}
		&
		\includegraphics[width=\figwidthNormalVisSupp\linewidth]{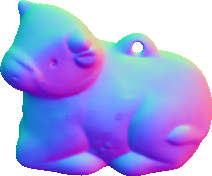}
		&
		\includegraphics[width=\figwidthNormalVisSupp\linewidth]{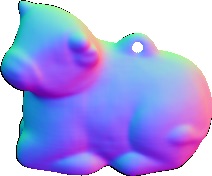}
		& 
		\includegraphics[width=\figwidthNormalVisSupp\linewidth]{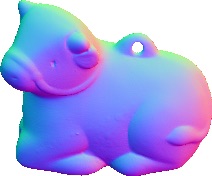}
		\\
		\includegraphics[width=\figwidthNormalVisSupp\linewidth]{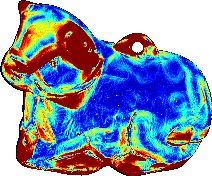}&
		\includegraphics[width=\figwidthNormalVisSupp\linewidth]{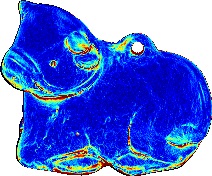}
		&
		\includegraphics[width=\figwidthNormalVisSupp\linewidth]{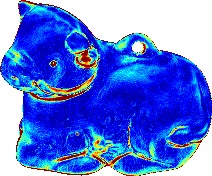}
		&
		\includegraphics[width=\figwidthNormalVisSupp\linewidth]{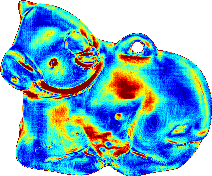}
		&
		\includegraphics[width=\figwidthNormalVisSupp\linewidth]{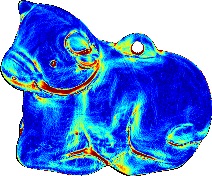}
		& 
		\colorbar{0.12}{$20^\circ$}{16}
	\end{tabular}
	\caption{More visual comparisons of recovered normal maps and angular error maps from the first view of \diligentmv~\cite{li2020multi} objects ``Bear,'' ``Buddha,'' and ``Cow.''}
	\label{fig.comp_mvps_normal_supp}
\end{figure}

%% file: sections/figures_tables/diligent_normal_unseen.tex
\begin{figure*}[h!]
	\small
	\newcommand{\figwidthNormalVisUnseen}{0.1}
	\centering
	\begin{tabular}{@{}c@{}c@{}c@{}c@{}c |c@{}c@{}c@{}c@{}c@{}c}
	\multicolumn{5}{c}{\psnerf~\cite{yang2022psnerf}} &
	\multicolumn{5}{c}{\mvas (ours)}
	\\
	view04 & view08 & view12 & view16 & view20 & view04 & view08 & view12 & view16 & view20 &\\
	\includegraphics[height=0.12\linewidth]{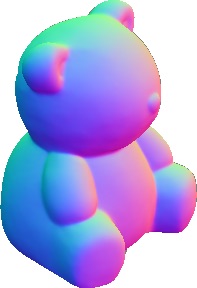}&
	\includegraphics[height=0.12\linewidth]{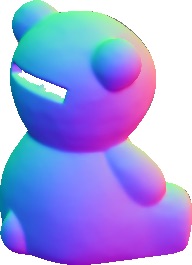}&
	\includegraphics[height=0.12\linewidth]{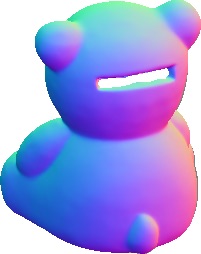}&	
	\includegraphics[height=0.12\linewidth]{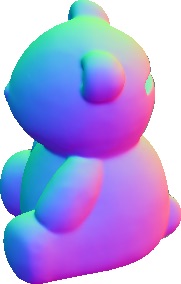}&
	\includegraphics[height=0.12\linewidth]{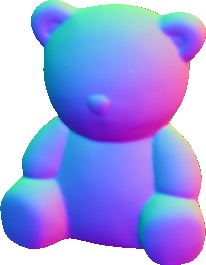}&				
	\includegraphics[height=0.12\linewidth]{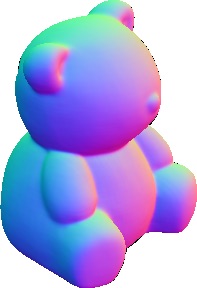}&
	\includegraphics[height=0.12\linewidth]{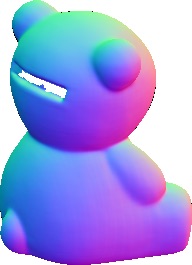}&
	\includegraphics[height=0.12\linewidth]{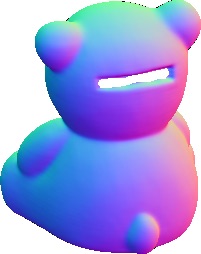}&
	\includegraphics[height=0.12\linewidth]{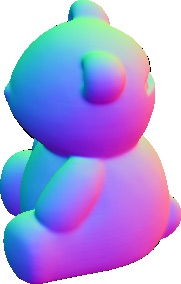}&
	\includegraphics[height=0.12\linewidth]{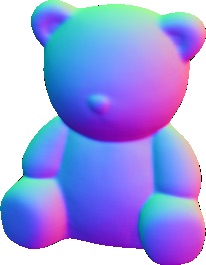}&
	\\
	\includegraphics[height=0.12\linewidth]{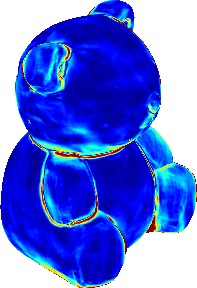}&
	\includegraphics[height=0.12\linewidth]{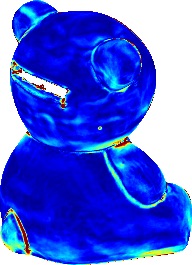}&
 	\includegraphics[height=0.12\linewidth]{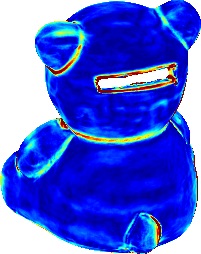}&
	\includegraphics[height=0.12\linewidth]{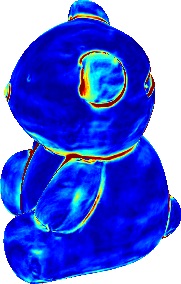}&
	\includegraphics[height=0.12\linewidth]{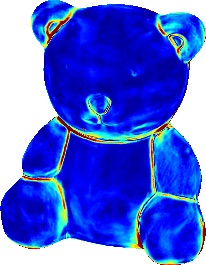}&
	\includegraphics[height=0.12\linewidth]{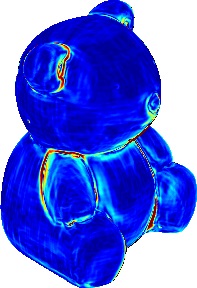}&
	\includegraphics[height=0.12\linewidth]{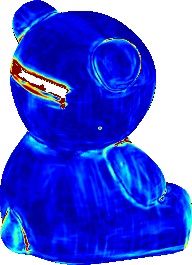}&
	\includegraphics[height=0.12\linewidth]{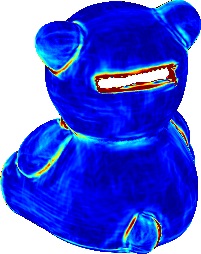}&
	\includegraphics[height=0.12\linewidth]{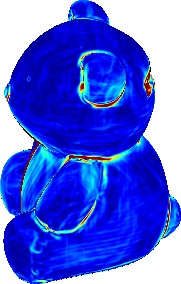}&
	\includegraphics[height=0.12\linewidth]{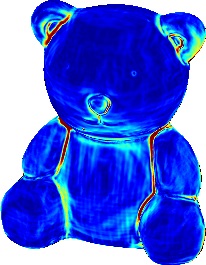}&
	\\
		\includegraphics[height=0.16\linewidth]{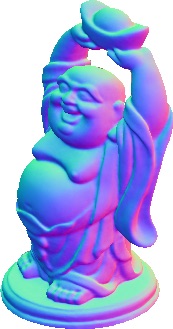}&
	\includegraphics[height=0.16\linewidth]{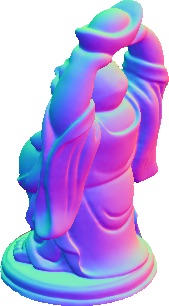}&
	\includegraphics[height=0.16\linewidth]{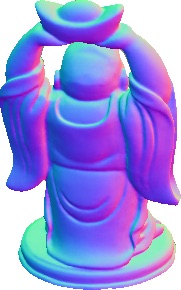}&	
	\includegraphics[height=0.16\linewidth]{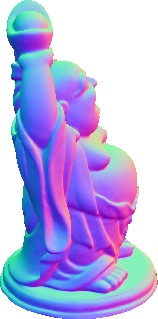}&
	\includegraphics[height=0.16\linewidth]{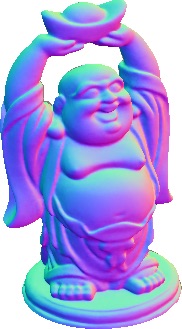}&				
	\includegraphics[height=0.16\linewidth]{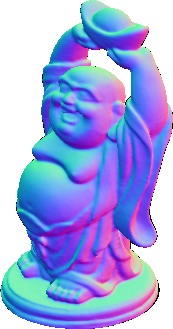}&
	\includegraphics[height=0.16\linewidth]{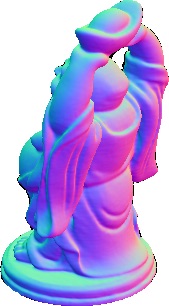}&
	\includegraphics[height=0.16\linewidth]{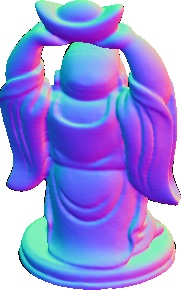}&
	\includegraphics[height=0.16\linewidth]{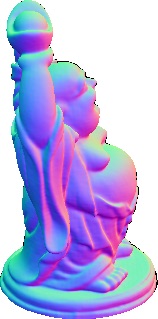}&
	\includegraphics[height=0.16\linewidth]{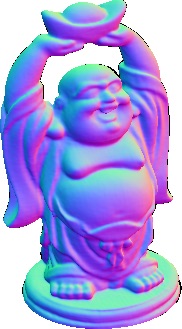}&
	\\
	\includegraphics[height=0.16\linewidth]{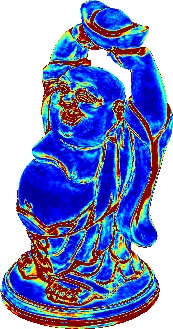}&
	\includegraphics[height=0.16\linewidth]{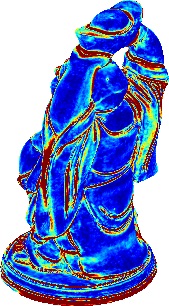}&
	\includegraphics[height=0.16\linewidth]{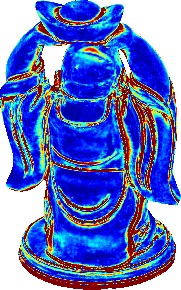}&
	\includegraphics[height=0.16\linewidth]{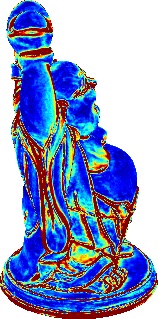}&
	\includegraphics[height=0.16\linewidth]{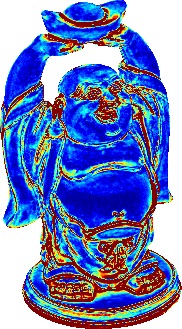}&
	\includegraphics[height=0.16\linewidth]{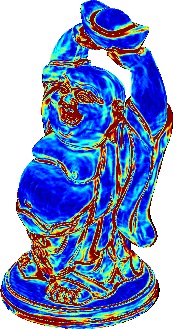}&
	\includegraphics[height=0.16\linewidth]{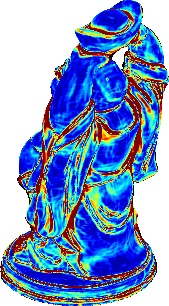}&
	\includegraphics[height=0.16\linewidth]{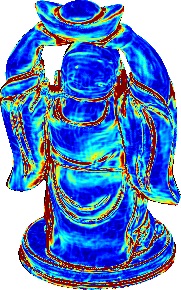}&
	\includegraphics[height=0.16\linewidth]{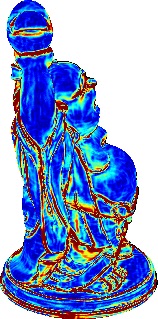}&
	\includegraphics[height=0.16\linewidth]{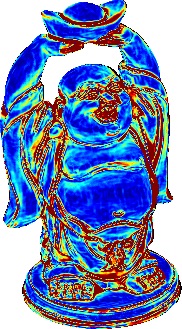}&
	\\
		\includegraphics[height=0.08\linewidth]{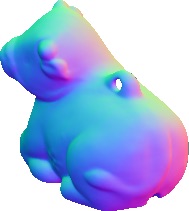}&
	\includegraphics[height=0.08\linewidth]{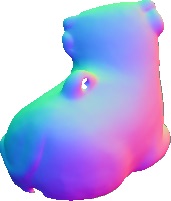}&
	\includegraphics[height=0.08\linewidth]{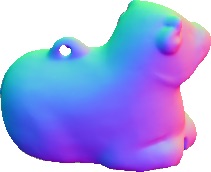}&	
	\includegraphics[height=0.08\linewidth]{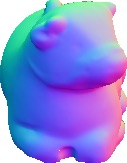}&
	\includegraphics[height=0.08\linewidth]{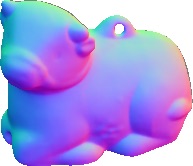}&				
	\includegraphics[height=0.08\linewidth]{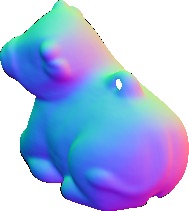}&
	\includegraphics[height=0.08\linewidth]{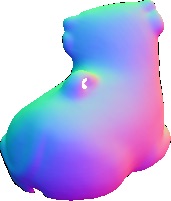}&
	\includegraphics[height=0.08\linewidth]{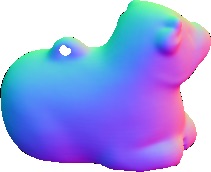}&
	\includegraphics[height=0.08\linewidth]{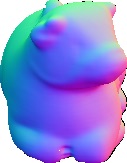}&
	\includegraphics[height=0.08\linewidth]{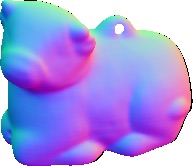}&
	\\
	\includegraphics[height=0.08\linewidth]{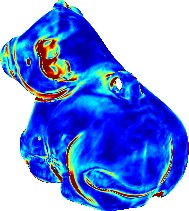}&
	\includegraphics[height=0.08\linewidth]{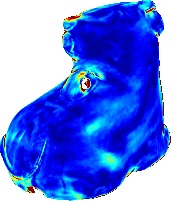}&
	\includegraphics[height=0.08\linewidth]{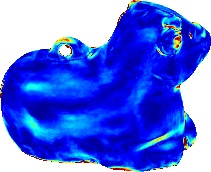}&
	\includegraphics[height=0.08\linewidth]{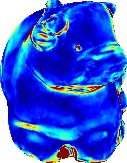}&
	\includegraphics[height=0.08\linewidth]{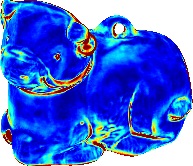}&
	\includegraphics[height=0.08\linewidth]{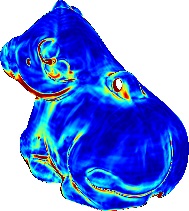}&
	\includegraphics[height=0.08\linewidth]{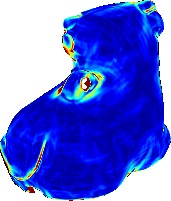}&
	\includegraphics[height=0.08\linewidth]{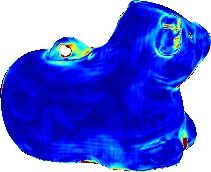}&
	\includegraphics[height=0.08\linewidth]{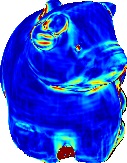}&
	\includegraphics[height=0.08\linewidth]{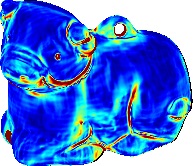}&
	\\
		\includegraphics[height=0.08\linewidth]{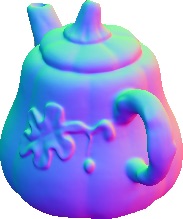}&
	\includegraphics[height=0.08\linewidth]{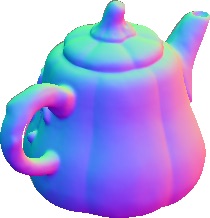}&
	\includegraphics[height=0.08\linewidth]{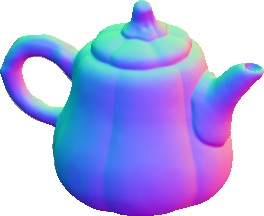}&	
	\includegraphics[height=0.08\linewidth]{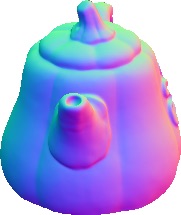}&
	\includegraphics[height=0.08\linewidth]{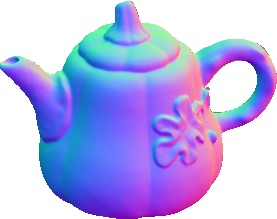}&				
	\includegraphics[height=0.08\linewidth]{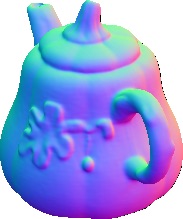}&
	\includegraphics[height=0.08\linewidth]{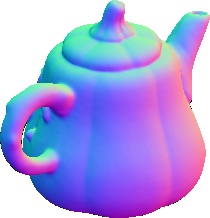}&
	\includegraphics[height=0.08\linewidth]{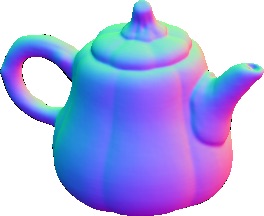}&
	\includegraphics[height=0.08\linewidth]{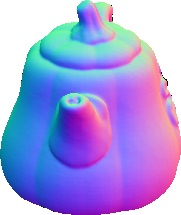}&
	\includegraphics[height=0.08\linewidth]{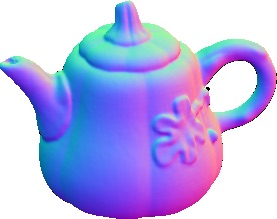}&
	\\
	\includegraphics[height=0.08\linewidth]{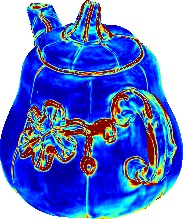}&
	\includegraphics[height=0.08\linewidth]{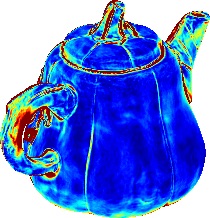}&
	\includegraphics[height=0.08\linewidth]{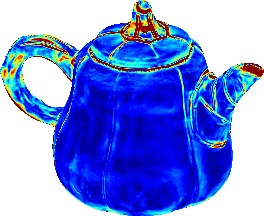}&
	\includegraphics[height=0.08\linewidth]{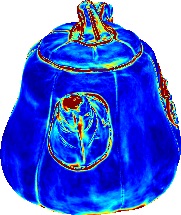}&
	\includegraphics[height=0.08\linewidth]{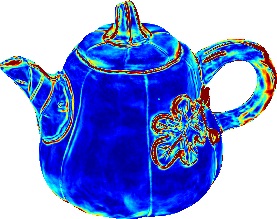}&
	\includegraphics[height=0.08\linewidth]{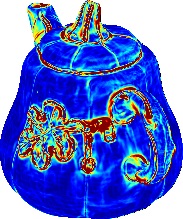}&
	\includegraphics[height=0.08\linewidth]{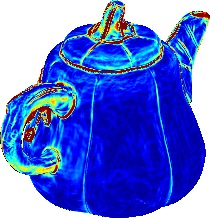}&
	\includegraphics[height=0.08\linewidth]{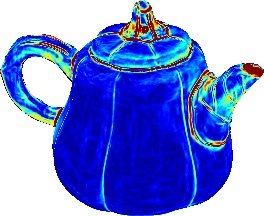}&
	\includegraphics[height=0.08\linewidth]{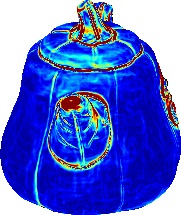}&
	\includegraphics[height=0.08\linewidth]{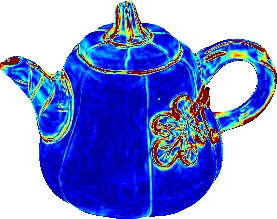}&
	\\
	\includegraphics[height=\figwidthNormalVisUnseen\linewidth]{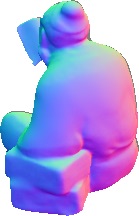}&
	\includegraphics[height=\figwidthNormalVisUnseen\linewidth]{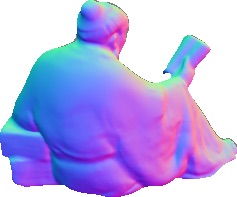}&
	\includegraphics[height=\figwidthNormalVisUnseen\linewidth]{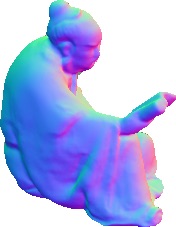}&	
	\includegraphics[height=\figwidthNormalVisUnseen\linewidth]{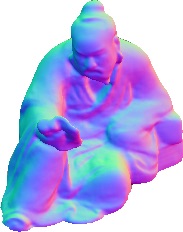}&
	\includegraphics[height=\figwidthNormalVisUnseen\linewidth]{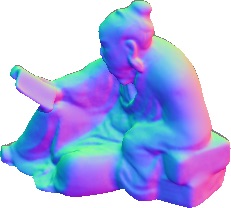}&				
	\includegraphics[height=\figwidthNormalVisUnseen\linewidth]{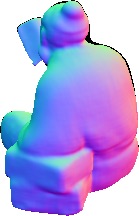}&
	\includegraphics[height=\figwidthNormalVisUnseen\linewidth]{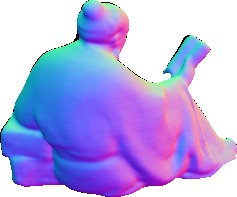}&
	\includegraphics[height=\figwidthNormalVisUnseen\linewidth]{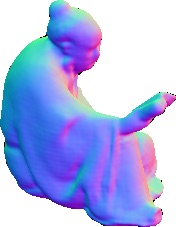}&
	\includegraphics[height=\figwidthNormalVisUnseen\linewidth]{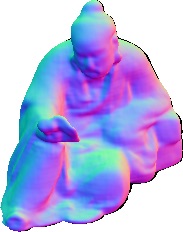}&
	\includegraphics[height=\figwidthNormalVisUnseen\linewidth]{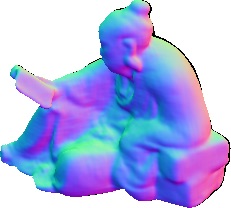}&
	\\
	\includegraphics[height=\figwidthNormalVisUnseen\linewidth]{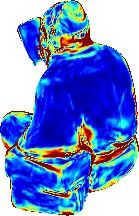}&
	\includegraphics[height=\figwidthNormalVisUnseen\linewidth]{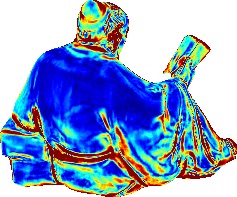}&
	\includegraphics[height=\figwidthNormalVisUnseen\linewidth]{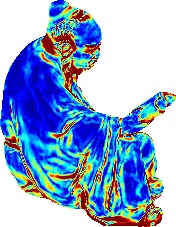}&
	\includegraphics[height=\figwidthNormalVisUnseen\linewidth]{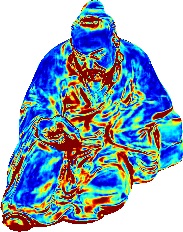}&
	\includegraphics[height=\figwidthNormalVisUnseen\linewidth]{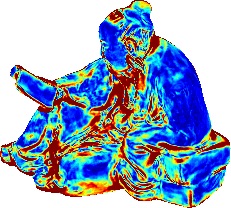}&
	\includegraphics[height=\figwidthNormalVisUnseen\linewidth]{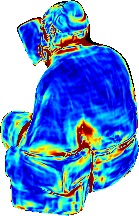}&
	\includegraphics[height=\figwidthNormalVisUnseen\linewidth]{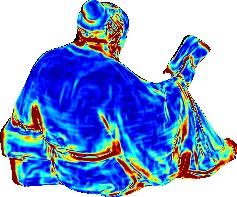}&
	\includegraphics[height=\figwidthNormalVisUnseen\linewidth]{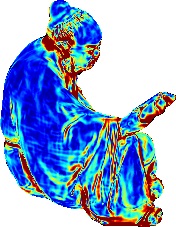}&
	\includegraphics[height=\figwidthNormalVisUnseen\linewidth]{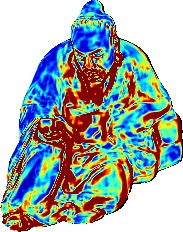}&
	\includegraphics[height=\figwidthNormalVisUnseen\linewidth]{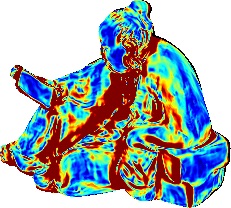}&
	\colorbar{0.1}{$20^\circ$}{32}
	\\
	\end{tabular}
	\caption{Visual comparisons to \psnerf~\cite{yang2022psnerf} of the $5$ unseen views over the training.}
	\label{fig.comp_mvps_normal_supp_unseen}
\end{figure*}

%% file: sections/figures_tables/diligent_num_views_quan.tex
\begin{table}
	\centering
	\small
	\caption{Effect of the number of views used for shape and normal recovery. MAE is averaged over \{5, 10, 12, 14, 15\} unseen views, respectively.}
	\begin{tabular}{@{}lccccc}
		\toprule
	Metrics & 15   & 10  & 8  & 6  & 5\\
	\midrule
	CD ($\downarrow$) &0.357& 0.372& 0.449& 0.424  & 0.422\\
	F-score ($\uparrow$) &0.754 & 0.739 & 0.648& 0.702 & 0.715\\
	MAE ($\downarrow$) &9.90 & 10.80& 12.23&13.35 & 14.25\\
	\bottomrule
	\end{tabular}
\label{tab.num_view}
\end{table}

%% file: sections/figures_tables/diligent_num_views.tex
\begin{figure*}
	\centering
	\begin{tabular}{c@{}cc@{}cc@{}cc@{}cc@{}c@{}c}
		\multicolumn{2}{c}{15 views}& 
		\multicolumn{2}{c}{10 views} & 
		\multicolumn{2}{c}{8 views}& 
		\multicolumn{2}{c}{6 views} & 
		\multicolumn{2}{c}{5 views}\\
		\multicolumn{2}{c}{\includegraphics[width=0.17\linewidth]{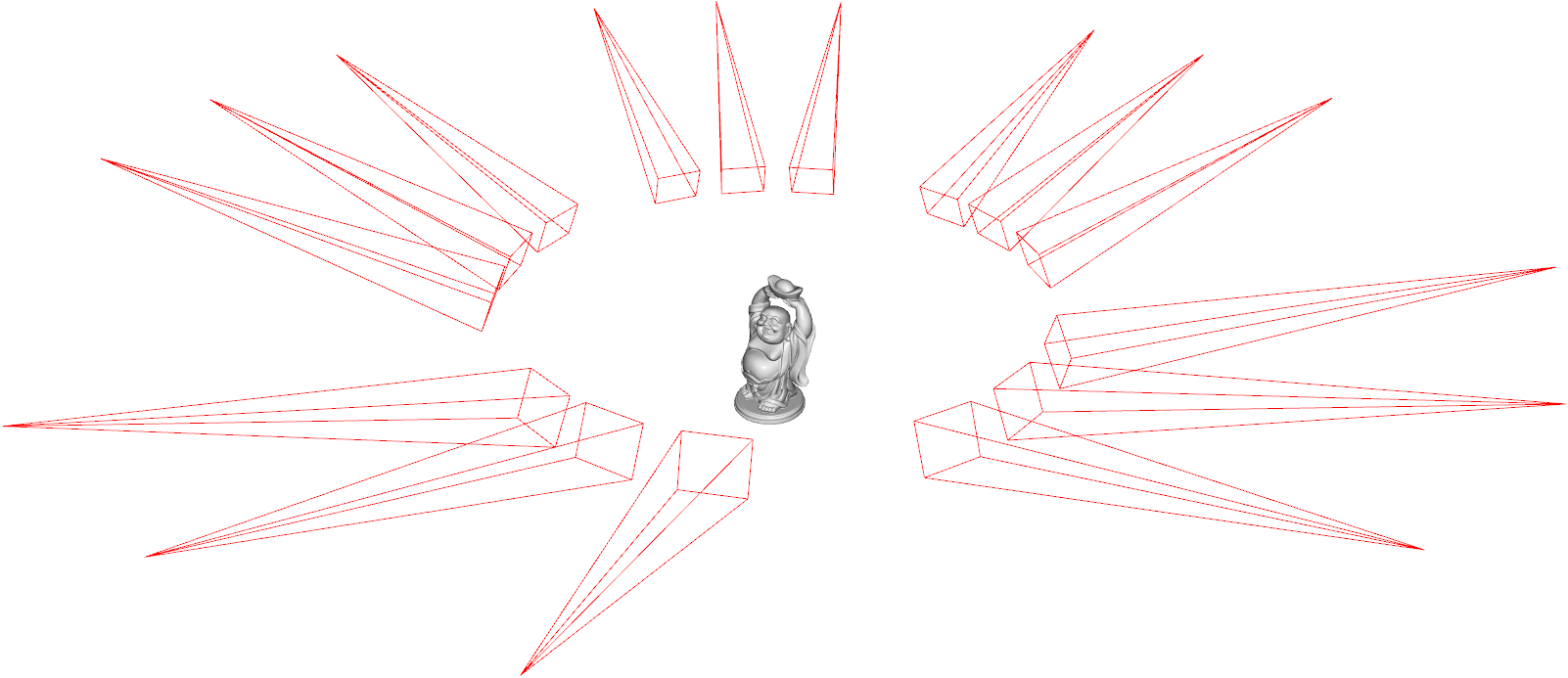}}&
		\multicolumn{2}{c}{\includegraphics[width=0.17\linewidth]{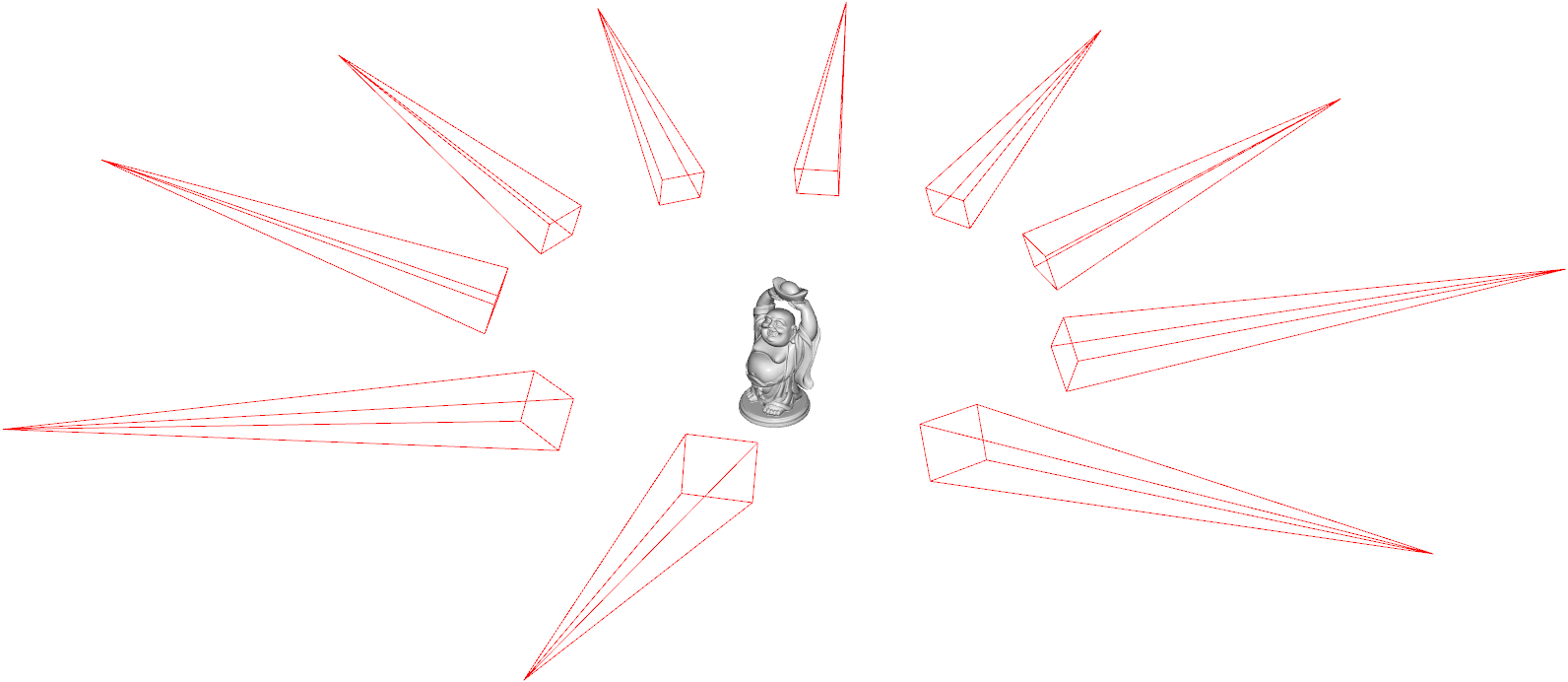}}&
		\multicolumn{2}{c}{\includegraphics[width=0.17\linewidth]{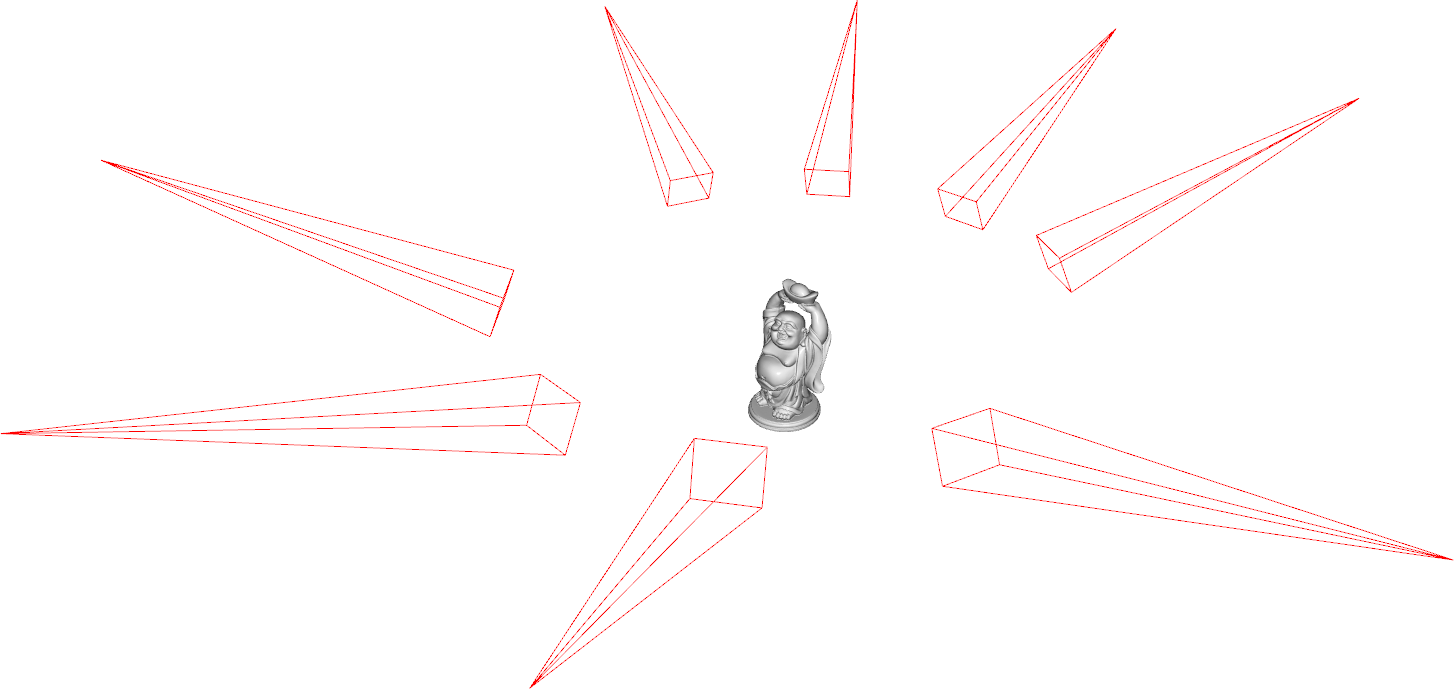}}&		
		\multicolumn{2}{c}{\includegraphics[width=0.17\linewidth]{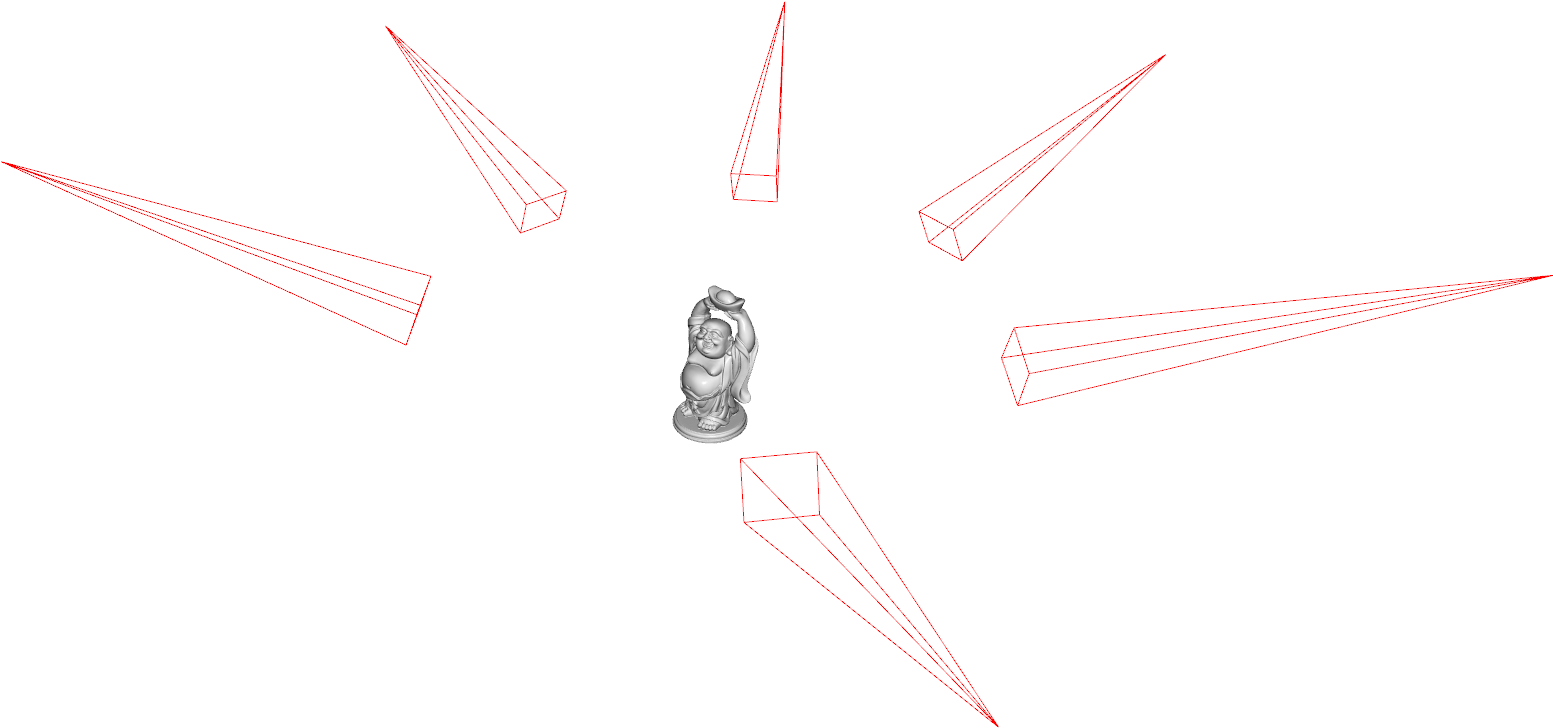}}&
		\multicolumn{2}{c}{\includegraphics[width=0.17\linewidth]{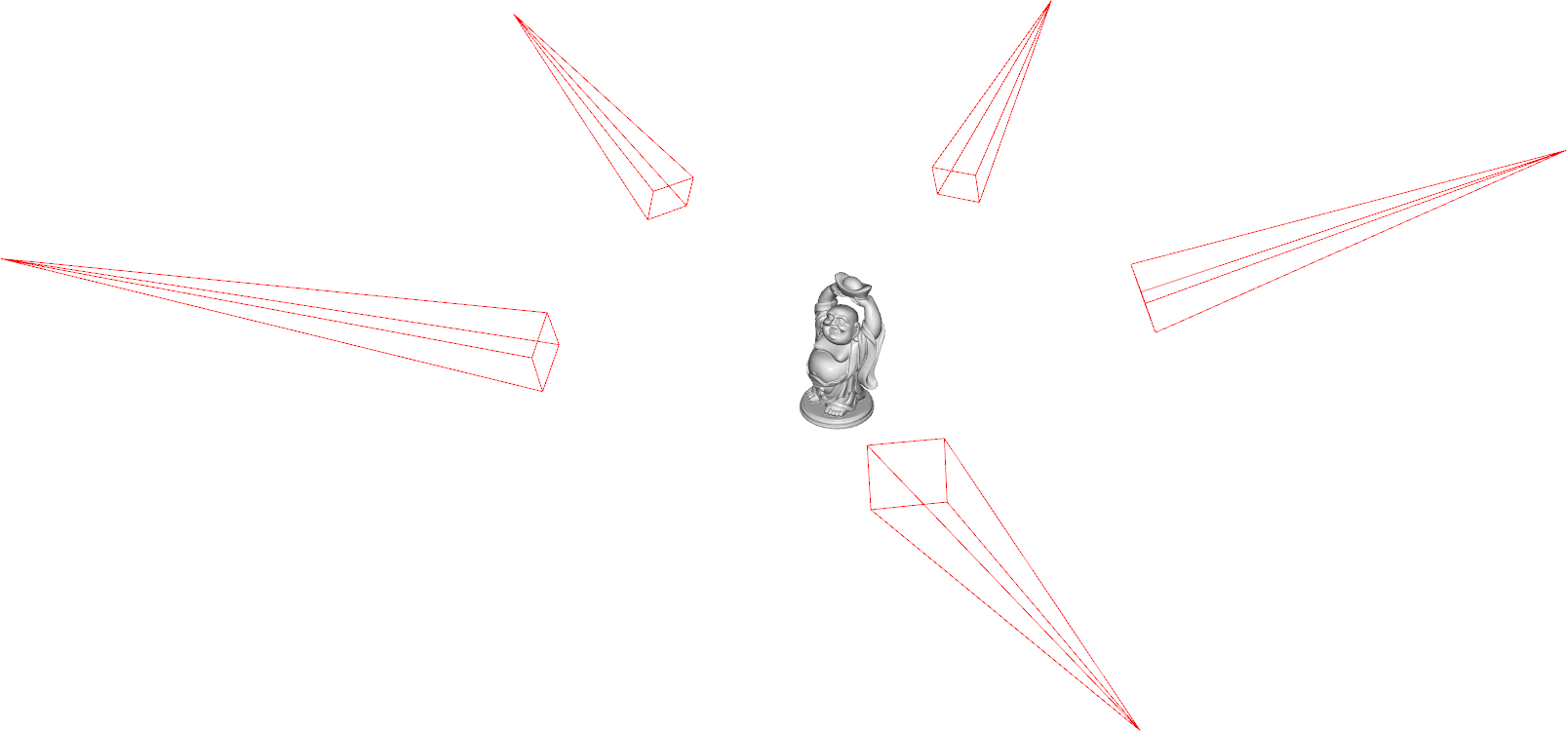}}&
		\\
		\includegraphics[height=0.14\linewidth]{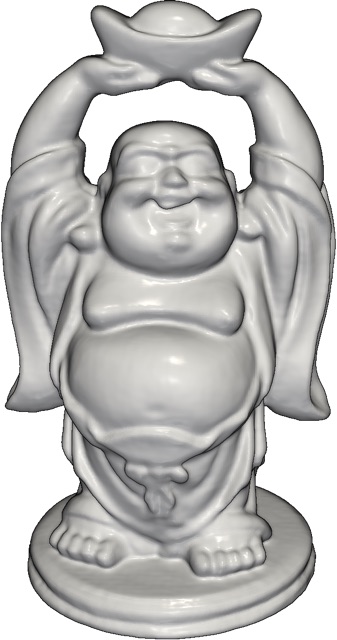}&
		\includegraphics[height=0.14\linewidth]{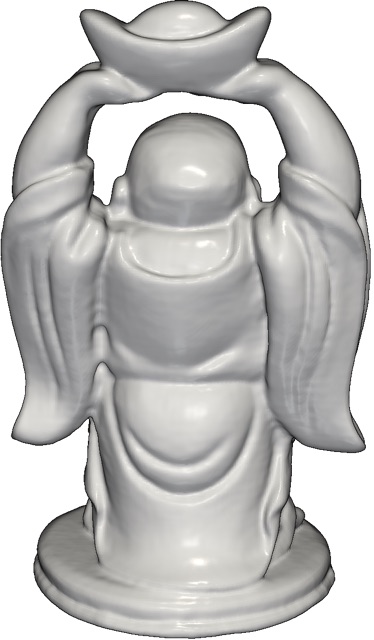}&
		\includegraphics[height=0.14\linewidth]{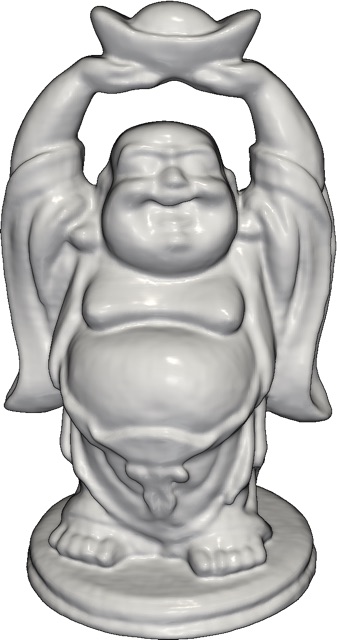}&
		\includegraphics[height=0.14\linewidth]{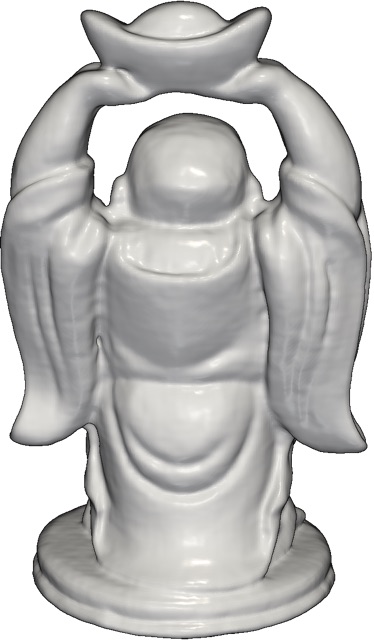}&
		\includegraphics[height=0.14\linewidth]{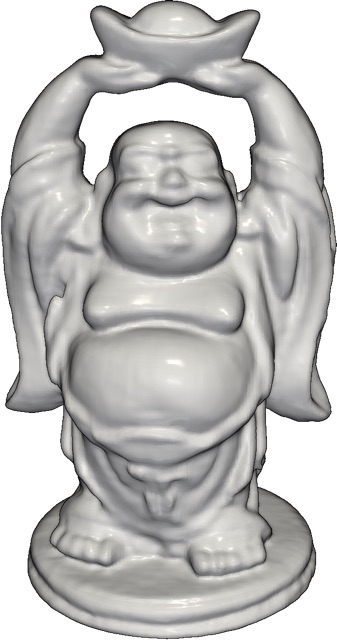}&
		\includegraphics[height=0.14\linewidth]{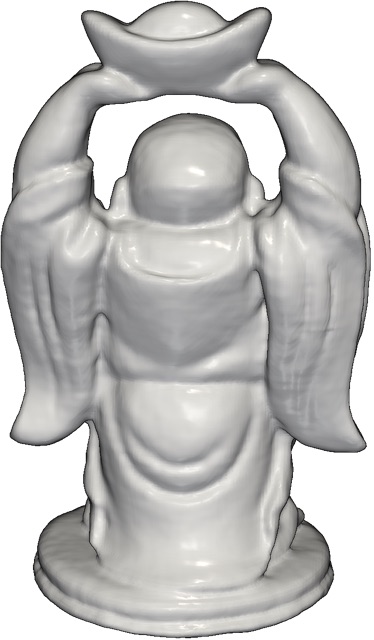}&
		\includegraphics[height=0.14\linewidth]{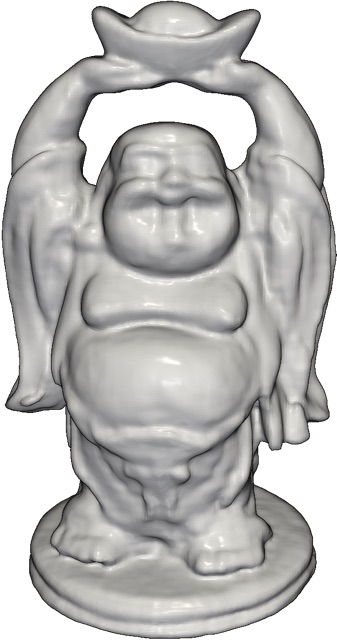}&
		\includegraphics[height=0.14\linewidth]{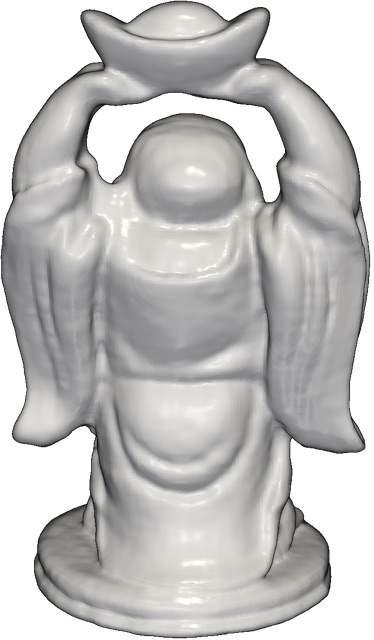}&
		\includegraphics[height=0.14\linewidth]{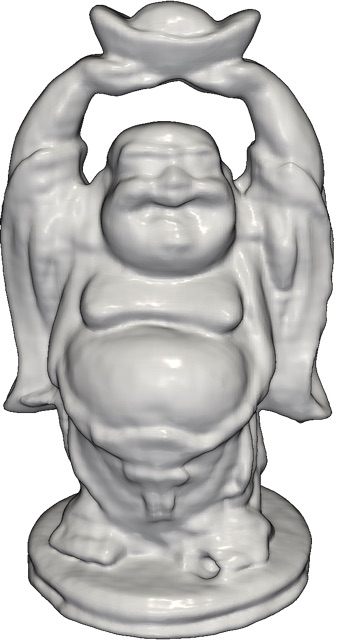}&
		\includegraphics[height=0.14\linewidth]{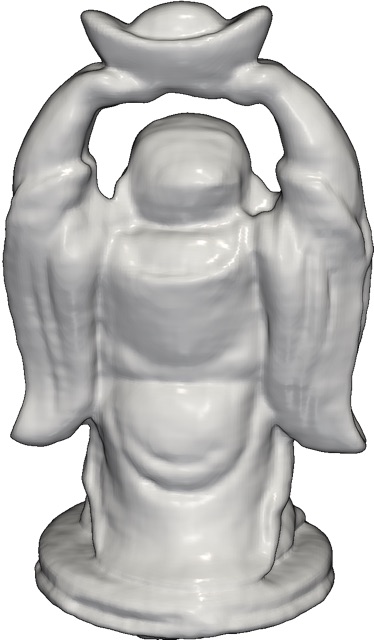}&
		\\
		\includegraphics[height=0.14\linewidth]{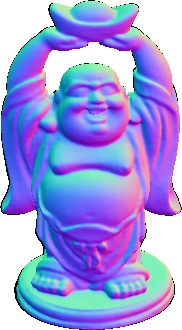}&
		\includegraphics[height=0.14\linewidth]{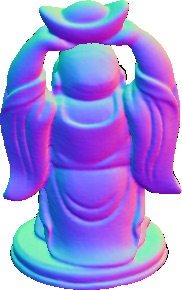}&		\includegraphics[height=0.14\linewidth]{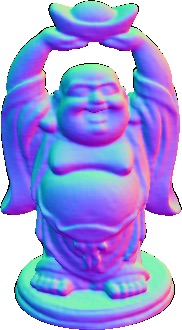}&
		\includegraphics[height=0.14\linewidth]{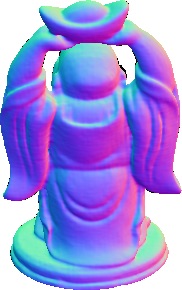}&
		\includegraphics[height=0.14\linewidth]{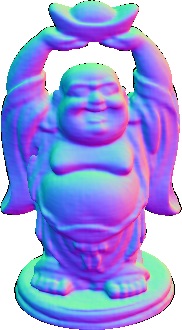}&
		\includegraphics[height=0.14\linewidth]{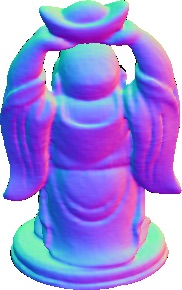}&
		\includegraphics[height=0.14\linewidth]{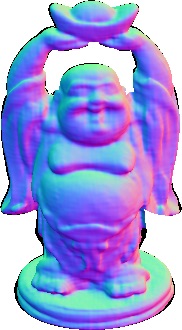}&
		\includegraphics[height=0.14\linewidth]{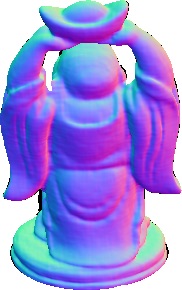}&
		\includegraphics[height=0.14\linewidth]{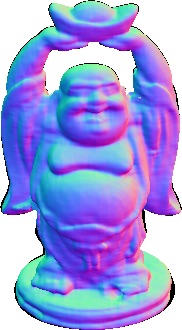}&
		\includegraphics[height=0.14\linewidth]{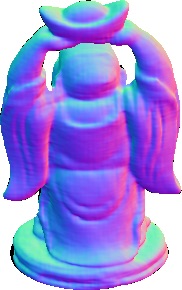}&
		\\
		\includegraphics[height=0.14\linewidth]{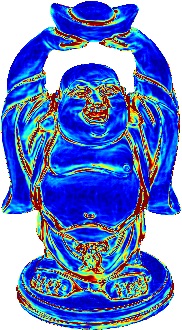}&
		\includegraphics[height=0.14\linewidth]{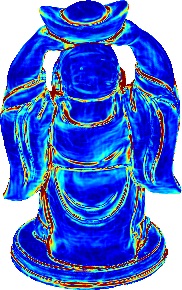}&	
		\includegraphics[height=0.14\linewidth]{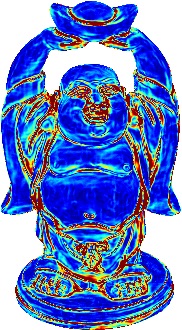}&
		\includegraphics[height=0.14\linewidth]{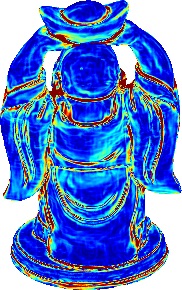}&	
		\includegraphics[height=0.14\linewidth]{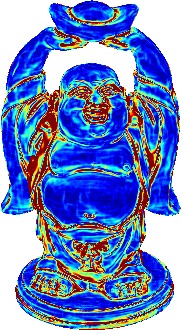}&
		\includegraphics[height=0.14\linewidth]{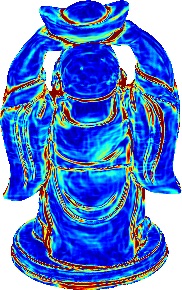}&	
		\includegraphics[height=0.14\linewidth]{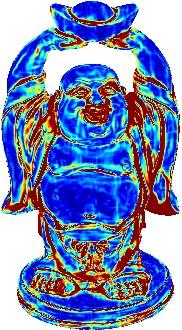}&
		\includegraphics[height=0.14\linewidth]{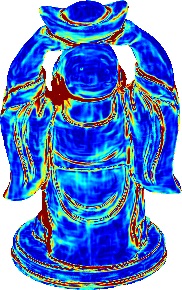}&	
		\includegraphics[height=0.14\linewidth]{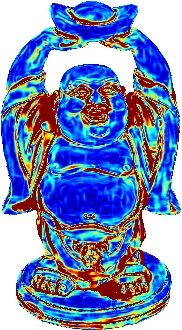}&
		\includegraphics[height=0.14\linewidth]{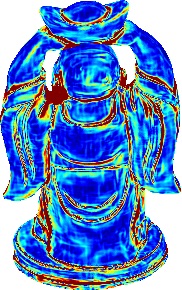}&
		\colorbar{0.14}{$30^\circ$}{50} \\
		10.88 & 8.42 & 11.63 & 9.65 & 12.76 & 9.99 & 16.88 & 10.90 & 16.29 & 11.90 &
	\end{tabular}
\caption{Surface and normal recovery results using different number of viewpoints. From top to bottom: input viewpoints, front and back views of recovered shapes, front and back normal maps, front and back angular error maps, and MAEs in corresponding views.
It can be seen that MVAS is robust to sparse view inputs. 
Most surface details are still distinguishable using as few as $5$-view azimuth maps.}
\label{fig.diligent_num_views}
\end{figure*}

%% file: sections/paragraph_implementation_details.tex
\subsection{Training details}
We initialize the MLP parameters such that the initial zero level set approximates a sphere with a radius $0.6$~\cite{sal2020cvpr}.
We set $\lambda_1=100$ and $\lambda_2=0.1$ for the loss function. 
ADAM optimizer is used with an initial learning rate $1\times 10^{-4}$.
We optimize the MLP parameters for $50$ epochs with a batchsize $4096$ pixels.
The learning rate and $\alpha$ in silhouette loss are divided by $2$ every $10$ epochs.

As most pixels from the input images are outside silhouette, randomly sampling from all pixels can be inefficient for training.
To improve the efficiency, we dilate the silhouette (\ie, the boundary of the mask) for $30$ times and sample pixels from the expanded regions as input.
For \diligentmv~\cite{li2020multi} objects, we use their provided masks.
For \pandora~\cite{dave2022pandora} and our captured images, we use an automatic image background removal tool~\cite{removebg} to generate the masks.
The input image dimensions are $612 \times 512$ for \diligentmv~\cite{li2020multi}, $1224 \times 1024$ for \pandora~\cite{dave2022pandora},  and $1566 \times 1045$ for our objects.

The training took about $3$ hours per \diligentmv object~\cite{li2020multi}, about $7$ hours per PANDORA object~\cite{dave2022pandora}, and about $10$ hours for our captured objects using one GTX 2080Ti graphics card.
As a comparison, \psnerf took about $22$ hours to train one  \diligentmv object~\cite{yang2022psnerf}.
It took us about 30 hours to reproduce \pandora results per object~\cite{dave2022pandora}.

%% file: main.bbl
\begin{thebibliography}{10}\itemsep=-1pt

\bibitem{removebg}
Remove {BG}.
\newblock \url{https://www.remove.bg}.
\newblock Accessed: 2022-11-10.

\bibitem{sonyPolar}
Sony polarization image sensor.
\newblock
  \url{https://www.sony-semicon.com/en/products/is/industry/polarization.html}.
\newblock Accessed: 2023-03-23.

\bibitem{alldrin2007toward}
Neil~G Alldrin and David~J Kriegman.
\newblock Toward reconstructing surfaces with arbitrary isotropic reflectance:
  {A} stratified photometric stereo approach.
\newblock In {\em {Proc. of International Conference on Computer Vision
  (ICCV)}}, pages 1--8, 2007.

\bibitem{sal2020cvpr}
Matan Atzmon and Yaron Lipman.
\newblock Sal: {S}ign agnostic learning of shapes from raw data.
\newblock In {\em {Proc. of Computer Vision and Pattern Recognition (CVPR)}},
  pages 2565--2574, 2020.

\bibitem{chandraker2012differential}
Manmohan Chandraker, Jiamin Bai, and Ravi Ramamoorthi.
\newblock On differential photometric reconstruction for unknown, isotropic
  {BRDF}s.
\newblock {\em {IEEE Transactions on Pattern Analysis and Machine Intelligence
  (PAMI)}}, 35(12):2941--2955, 2012.

\bibitem{chang2007multiview}
Ju~Yong Chang, Kyoung~Mu Lee, and Sang~Uk Lee.
\newblock Multiview normal field integration using level set methods.
\newblock In {\em {Proc. of Computer Vision and Pattern Recognition (CVPR)}},
  pages 1--8. IEEE, 2007.

\bibitem{chen2019SDPS_Net}
Guanying Chen, Kai Han, Boxin Shi, Yasuyuki Matsushita, and Kwan-Yee~K. Wong.
\newblock {SDPS}-{N}et: {S}elf-calibrating deep photometric stereo networks.
\newblock In {\em {Proc. of Computer Vision and Pattern Recognition (CVPR)}},
  2019.

\bibitem{Cui_2017_CVPR}
Zhaopeng Cui, Jinwei Gu, Boxin Shi, Ping Tan, and Jan Kautz.
\newblock Polarimetric multi-view stereo.
\newblock In {\em {Proc. of Computer Vision and Pattern Recognition (CVPR)}},
  2017.

\bibitem{dave2022pandora}
Akshat Dave, Yongyi Zhao, and Ashok Veeraraghavan.
\newblock {PANDORA}: {P}olarization-aided neural decomposition of radiance.
\newblock {\em {Proc. of European Conference on Computer Vision (ECCV)}}, 2022.

\bibitem{ding2021polarimetric}
Yuqi Ding, Yu Ji, Mingyuan Zhou, Sing~Bing Kang, and Jinwei Ye.
\newblock Polarimetric {H}elmholtz stereopsis.
\newblock In {\em {Proc. of International Conference on Computer Vision
  (ICCV)}}, 2021.

\bibitem{drbohlav2001unambiguous}
Ondrej Drbohlav and Radim Sara.
\newblock Unambiguous determination of shape from photometric stereo with
  unknown light sources.
\newblock In {\em {Proc. of International Conference on Computer Vision
  (ICCV)}}, volume~1, pages 581--586, 2001.

\bibitem{fukao2021polarimetric}
Yoshiki Fukao, Ryo Kawahara, Shohei Nobuhara, and Ko Nishino.
\newblock Polarimetric normal stereo.
\newblock In {\em Proceedings of the IEEE/CVF Conference on Computer Vision and
  Pattern Recognition}, pages 682--690, 2021.

\bibitem{furukawa2015multi}
Yasutaka Furukawa, Carlos Hern{\'a}ndez, et~al.
\newblock Multi-view stereo: {A} tutorial.
\newblock {\em Foundations and Trends in Computer Graphics and Vision},
  9(1-2):1--148, 2015.

\bibitem{furukawa2009accurate}
Yasutaka Furukawa and Jean Ponce.
\newblock Accurate, dense, and robust multiview stereopsis.
\newblock {\em {IEEE Transactions on Pattern Analysis and Machine Intelligence
  (PAMI)}}, 32(8):1362--1376, 2009.

\bibitem{multiview2007goesele}
Michael Goesele, Noah Snavely, Brian Curless, Hugues Hoppe, and Steven~M Seitz.
\newblock Multi-view stereo for community photo collections.
\newblock In {\em {Proc. of International Conference on Computer Vision
  (ICCV)}}, 2007.

\bibitem{igr2020icml}
Amos Gropp, Lior Yariv, Niv Haim, Matan Atzmon, and Yaron Lipman.
\newblock Implicit geometric regularization for learning shapes.
\newblock In {\em Proceedings of Machine Learning and Systems 2020}, pages
  3569--3579. 2020.

\bibitem{sphere1996hart}
John~C Hart.
\newblock Sphere tracing: {A} geometric method for the antialiased ray tracing
  of implicit surfaces.
\newblock {\em The Visual Computer}, 12(10):527--545, 1996.

\bibitem{hernandez2008multiview}
Carlos Hernandez, George Vogiatzis, and Roberto Cipolla.
\newblock Multiview photometric stereo.
\newblock {\em {IEEE Transactions on Pattern Analysis and Machine Intelligence
  (PAMI)}}, 30(3):548--554, 2008.

\bibitem{facerelighting2022}
Andrew Hou, Michel Sarkis, Ning Bi, Yiying Tong, and Xiaoming Liu.
\newblock Face relighting with geometrically consistent shadows.
\newblock In {\em {Proc. of Computer Vision and Pattern Recognition (CVPR)}},
  pages 4217--4226, 2022.

\bibitem{kadambi2015polarized}
Achuta Kadambi, Vage Taamazyan, Boxin Shi, and Ramesh Raskar.
\newblock Polarized 3{D}: High-quality depth sensing with polarization cues.
\newblock In {\em {Proc. of International Conference on Computer Vision
  (ICCV)}}, pages 3370--3378, 2015.

\bibitem{kadambi2017depth}
Achuta Kadambi, Vage Taamazyan, Boxin Shi, and Ramesh Raskar.
\newblock Depth sensing using geometrically constrained polarization normals.
\newblock {\em {International Journal of Computer Vision (IJCV)}},
  125(1):34--51, 2017.

\bibitem{kaya2022uncertainty}
Berk Kaya, Suryansh Kumar, Carlos Oliveira, Vittorio Ferrari, and Luc Van~Gool.
\newblock Uncertainty-aware deep multi-view photometric stereo.
\newblock In {\em {Proc. of Computer Vision and Pattern Recognition (CVPR)}},
  pages 12601--12611, 2022.

\bibitem{knapitsch2017tanks}
Arno Knapitsch, Jaesik Park, Qian-Yi Zhou, and Vladlen Koltun.
\newblock Tanks and temples: {B}enchmarking large-scale scene reconstruction.
\newblock {\em ACM Transactions on Graphics (TOG)}, 36(4):1--13, 2017.

\bibitem{visualhull1994}
Aldo Laurentini.
\newblock The visual hull concept for silhouette-based image understanding.
\newblock {\em {IEEE Transactions on Pattern Analysis and Machine Intelligence
  (PAMI)}}, 1994.

\bibitem{li2020multi}
Min Li, Zhenglong Zhou, Zhe Wu, Boxin Shi, Changyu Diao, and Ping Tan.
\newblock Multi-view photometric stereo: {A} robust solution and benchmark
  dataset for spatially varying isotropic materials.
\newblock {\em IEEE Transactions on Image Processing}, 29:4159--4173, 2020.

\bibitem{logothetis2019differential}
Fotios Logothetis, Roberto Mecca, and Roberto Cipolla.
\newblock A differential volumetric approach to multi-view photometric stereo.
\newblock In {\em {Proc. of International Conference on Computer Vision
  (ICCV)}}, pages 1052--1061, 2019.

\bibitem{lorensen1987marching}
William~E Lorensen and Harvey~E Cline.
\newblock Marching cubes: {A} high resolution 3{D} surface construction
  algorithm.
\newblock {\em ACM siggraph computer graphics}, 21(4):163--169, 1987.

\bibitem{mildenhall2020nerf}
Ben Mildenhall, Pratul~P. Srinivasan, Matthew Tancik, Jonathan~T. Barron, Ravi
  Ramamoorthi, and Ren Ng.
\newblock Nerf: {R}epresenting scenes as neural radiance fields for view
  synthesis.
\newblock In {\em {Proc. of European Conference on Computer Vision (ECCV)}},
  2020.

\bibitem{minami2022symmetric}
Kazuma Minami, Hiroaki Santo, Fumio Okura, and Yasuyuki Matsushita.
\newblock Symmetric-light photometric stereo.
\newblock In {\em {Proc. of IEEE/CVF Winter Conference on Applications of
  Computer Vision (WACV)}}, pages 2706--2714, 2022.

\bibitem{miyazaki2003polarizationtwoview}
Daisuke Miyazaki, Masataka Kagesawa, and Katsushi Ikeuchi.
\newblock Polarization-based transparent surface modeling from two views.
\newblock In {\em {Proc. of International Conference on Computer Vision
  (ICCV)}}, volume~3, pages 1381--1381, 2003.

\bibitem{miyazaki2003polarization}
Daisuke Miyazaki, Robby~T Tan, Kenji Hara, and Katsushi Ikeuchi.
\newblock Polarization-based inverse rendering from a single view.
\newblock In {\em {Proc. of International Conference on Computer Vision
  (ICCV)}}, volume~3, pages 982--982, 2003.

\bibitem{nehab2005efficiently}
Diego Nehab, Szymon Rusinkiewicz, James Davis, and Ravi Ramamoorthi.
\newblock Efficiently combining positions and normals for precise 3{D}
  geometry.
\newblock {\em ACM Transactions on Graphics (TOG)}, 24(3):536--543, 2005.

\bibitem{osher2004level}
Stanley Osher, Ronald Fedkiw, and K Piechor.
\newblock Level set methods and dynamic implicit surfaces.
\newblock {\em Applied Mechanics Reviews}, 57(3):B15--B15, 2004.

\bibitem{park2016robust}
Jaesik Park, Sudipta~N Sinha, Yasuyuki Matsushita, Yu-Wing Tai, and In~So
  Kweon.
\newblock Robust multiview photometric stereo using planar mesh
  parameterization.
\newblock {\em {IEEE Transactions on Pattern Analysis and Machine Intelligence
  (PAMI)}}, 39(8):1591--1604, 2016.

\bibitem{rahmann2001reconstruction}
Stefan Rahmann and Nikos Canterakis.
\newblock Reconstruction of specular surfaces using polarization imaging.
\newblock In {\em {Proc. of Computer Vision and Pattern Recognition (CVPR)}},
  2001.

\bibitem{schoenberger2016sfm}
Johannes~Lutz Sch\"{o}nberger and Jan-Michael Frahm.
\newblock Structure-from-motion revisited.
\newblock In {\em {Proc. of Computer Vision and Pattern Recognition (CVPR)}},
  2016.

\bibitem{schoenberger2016mvs}
Johannes~Lutz Sch\"{o}nberger, Enliang Zheng, Marc Pollefeys, and Jan-Michael
  Frahm.
\newblock Pixelwise view selection for unstructured multi-view stereo.
\newblock In {\em {Proc. of European Conference on Computer Vision (ECCV)}},
  2016.

\bibitem{shi2019}
Boxin Shi, Zhipeng Mo, Zhe Wu, Dinglong Duan, Sai-Kit Yeung, and Ping Tan.
\newblock A benchmark dataset and evaluation for non-{L}ambertian and
  uncalibrated photometric stereo.
\newblock {\em {IEEE Transactions on Pattern Analysis and Machine Intelligence
  (PAMI)}}, 2019.

\bibitem{siren2020sitzmann}
Vincent Sitzmann, Julien Martel, Alexander Bergman, David Lindell, and Gordon
  Wetzstein.
\newblock Implicit neural representations with periodic activation functions.
\newblock {\em {Advances in Neural Information Processing Systems (NeurIPS)}},
  33:7462--7473, 2020.

\bibitem{smith2016linear}
William~AP Smith, Ravi Ramamoorthi, and Silvia Tozza.
\newblock Linear depth estimation from an uncalibrated, monocular polarisation
  image.
\newblock In {\em {Proc. of European Conference on Computer Vision (ECCV)}},
  pages 109--125. Springer, 2016.

\bibitem{smith2018height}
William~AP Smith, Ravi Ramamoorthi, and Silvia Tozza.
\newblock Height-from-polarisation with unknown lighting or albedo.
\newblock {\em {IEEE Transactions on Pattern Analysis and Machine Intelligence
  (PAMI)}}, 41(12):2875--2888, 2018.

\bibitem{stolz2012shape}
Christophe Stolz, Mathias Ferraton, and Fabrice Meriaudeau.
\newblock Shape from polarization: {A} method for solving zenithal angle
  ambiguity.
\newblock {\em Optics Letters}, 37(20):4218--4220, 2012.

\bibitem{multiview2007vogiatzis}
George Vogiatzis, Carlos~Hern{\'a}ndez Esteban, Philip~HS Torr, and Roberto
  Cipolla.
\newblock Multiview stereo via volumetric graph-cuts and occlusion robust
  photo-consistency.
\newblock {\em {IEEE Transactions on Pattern Analysis and Machine Intelligence
  (PAMI)}}, 29(12):2241--2246, 2007.

\bibitem{multiview2005vogiatzis}
George Vogiatzis, Philip~HS Torr, and Roberto Cipolla.
\newblock Multi-view stereo via volumetric graph-cuts.
\newblock In {\em {Proc. of Computer Vision and Pattern Recognition (CVPR)}},
  2005.

\bibitem{vu2011high}
Hoang-Hiep Vu, Patrick Labatut, Jean-Philippe Pons, and Renaud Keriven.
\newblock High accuracy and visibility-consistent dense multiview stereo.
\newblock {\em {IEEE Transactions on Pattern Analysis and Machine Intelligence
  (PAMI)}}, 34(5):889--901, 2011.

\bibitem{woodham1980photometric}
Robert~J Woodham.
\newblock Photometric method for determining surface orientation from multiple
  images.
\newblock {\em Optical Engineering}, 19(1):139--144, 1980.

\bibitem{yang2022psnerf}
Wenqi Yang, Guanying Chen, Chaofeng Chen, Zhenfang Chen, and Kwan-Yee~K. Wong.
\newblock {PS-NeRF}: {N}eural inverse rendering for multi-view photometric
  stereo.
\newblock In {\em {Proc. of European Conference on Computer Vision (ECCV)}},
  2022.

\bibitem{volsdf2021yariv}
Lior Yariv, Jiatao Gu, Yoni Kasten, and Yaron Lipman.
\newblock Volume rendering of neural implicit surfaces.
\newblock {\em {Advances in Neural Information Processing Systems (NeurIPS)}},
  34:4805--4815, 2021.

\bibitem{idr2020multiview}
Lior Yariv, Yoni Kasten, Dror Moran, Meirav Galun, Matan Atzmon, Basri Ronen,
  and Yaron Lipman.
\newblock Multiview neural surface reconstruction by disentangling geometry and
  appearance.
\newblock {\em {Advances in Neural Information Processing Systems (NeurIPS)}},
  33, 2020.

\bibitem{nerfactor2021}
Xiuming Zhang, Pratul~P Srinivasan, Boyang Deng, Paul Debevec, William~T
  Freeman, and Jonathan~T Barron.
\newblock Nerfactor: Neural factorization of shape and reflectance under an
  unknown illumination.
\newblock {\em ACM Transactions on Graphics (Proc. of ACM SIGGRAPH)},
  40(6):1--18, 2021.

\bibitem{zhao2020polarimetric}
Jinyu Zhao, Yusuke Monno, and Masatoshi Okutomi.
\newblock Polarimetric multi-view inverse rendering.
\newblock In {\em {Proc. of European Conference on Computer Vision (ECCV)}},
  pages 85--102. Springer, 2020.

\bibitem{Zhou2010}
Z. Zhou and P. Tan.
\newblock Ring-light photometric stereo.
\newblock In {\em {Proc. of European Conference on Computer Vision (ECCV)}},
  pages 265--279, 2010.

\bibitem{zhu2019depth}
Dizhong Zhu and William~AP Smith.
\newblock Depth from a polarisation+ {RGB} stereo pair.
\newblock In {\em {Proc. of Computer Vision and Pattern Recognition (CVPR)}},
  pages 7586--7595, 2019.

\bibitem{zickler2002helmholtz}
Todd~E Zickler, Peter~N Belhumeur, and David~J Kriegman.
\newblock Helmholtz stereopsis: {E}xploiting reciprocity for surface
  reconstruction.
\newblock {\em {International Journal of Computer Vision (IJCV)}},
  49(2):215--227, 2002.

\end{thebibliography}
